\documentclass{article} % For LaTeX2e
\usepackage{iclr2022_conference,times}

\usepackage{amsmath,amsfonts,bm}

\def\eqref#1{equation~\ref{#1}}
\def\1{\bm{1}}

\DeclareMathAlphabet{\mathsfit}{\encodingdefault}{\sfdefault}{m}{sl}
\SetMathAlphabet{\mathsfit}{bold}{\encodingdefault}{\sfdefault}{bx}{n}

\usepackage{hyperref}
\usepackage{url}

\usepackage{tikz}
\usepackage{comment}
\usepackage{amsmath,amssymb} % define this before the line numbering.
\usepackage{color}
\usepackage[accsupp]{axessibility}  % Improves PDF readability for those with disabilities.

\usepackage[width=122mm,left=12mm,paperwidth=146mm,height=193mm,top=12mm,paperheight=217mm]{geometry}

\usepackage{booktabs}
\usepackage{tabularx}
\usepackage{multirow}

\usepackage{graphicx}
\usepackage{amsmath}
\usepackage{amssymb}
\usepackage{booktabs}
\usepackage{comment}
\usepackage{xcolor}  % colors
\usepackage{caption}

\usepackage[normalem]{ulem}
\usepackage{algorithm}
\usepackage{algorithmic}
\def\correspondingauthor{\footnote{Corresponding author.}}
\usepackage[export]{adjustbox}

\usepackage[capitalize]{cleveref}
\crefname{section}{Sec.}{Secs.}
\crefname{table}{Table}{Tables}
\crefname{figure}{Fig.}{Figs.}
\crefname{table}{Tab.}{Tabs.}

\title{Benchmarking Deepart Detection} % Replace with your title

\author{%
  Yabin Wang\textsuperscript{\rm 1,2}, Zhiwu Huang\textsuperscript{\rm 2,3}\correspondingauthor, Xiaopeng Hong\textsuperscript{\rm 4, 5, 1}\correspondingauthor\\
  \textsuperscript{\rm 1}Xi'an Jiaotong University, P. R. China, \textsuperscript{\rm 2}Singapore Management University, Singapore\\
  \textsuperscript{\rm 3}University of Southampton, United Kingdom\\
  \textsuperscript{\rm 4}Harbin Institute of Technology, P. R. China, \textsuperscript{\rm 5}Pengcheng Laboratory, P. R. China\\  
  \texttt{iamwangyabin@stu.xjtu.edu.cn, zhiwu.huang@soton.ac.uk} \\ \texttt{hongxiaopeng@ieee.org} \\
}

\iclrfinalcopy % Uncomment for camera-ready version, but NOT for submission.

\begin{document}

% \onecolumn[{%
% \renewcommand\twocolumn[1][]{#1}%
\maketitle

\begin{center}
\centering
{\scriptsize
\def\teaserwid{0.082\linewidth}
\begin{tabular}{c@{\hspace{.5mm}}*{11}{c@{\hspace{.5mm}}}}
\rotatebox[origin=c]{90}{LAION-5B}&
\raisebox{-0.45\height}{\includegraphics[width=\teaserwid]{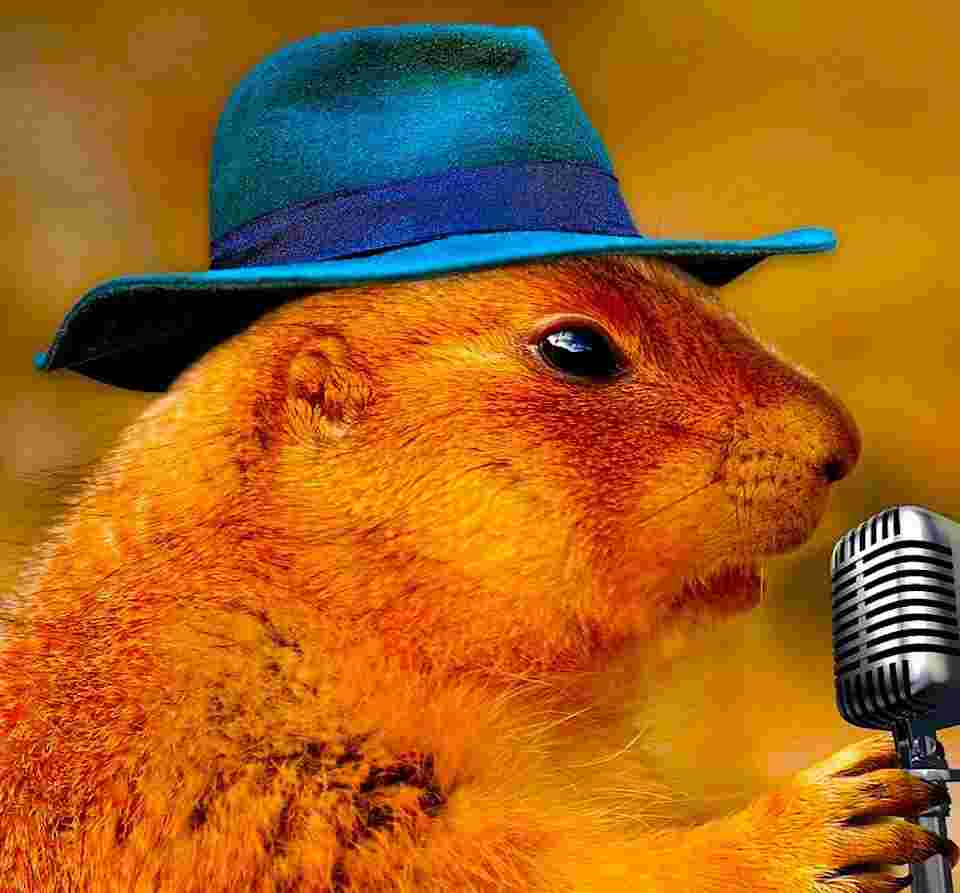}}&  \raisebox{-0.45\height}{\includegraphics[width=\teaserwid]{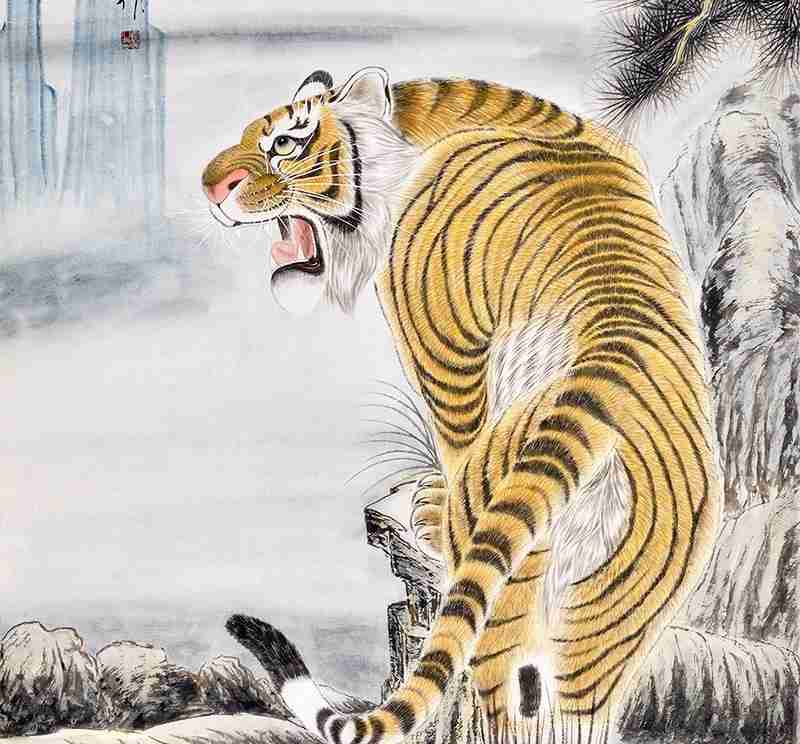}}&
\raisebox{-0.45\height}{\includegraphics[width=\teaserwid]{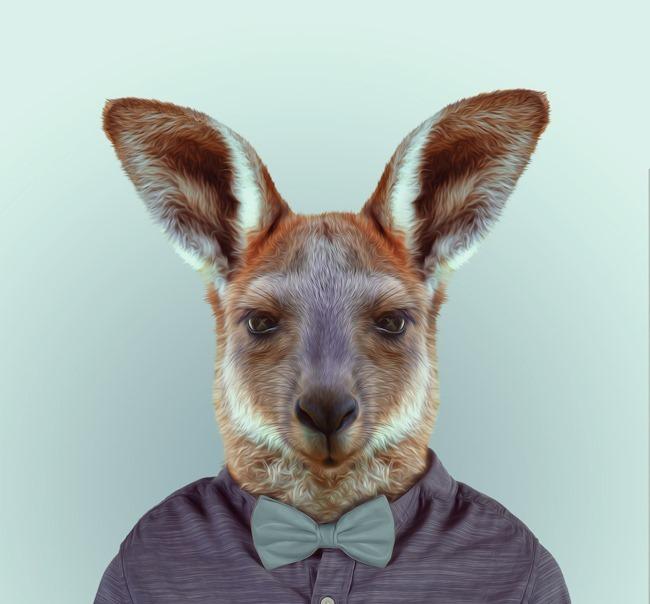}}&
\raisebox{-0.45\height}{\includegraphics[width=\teaserwid]{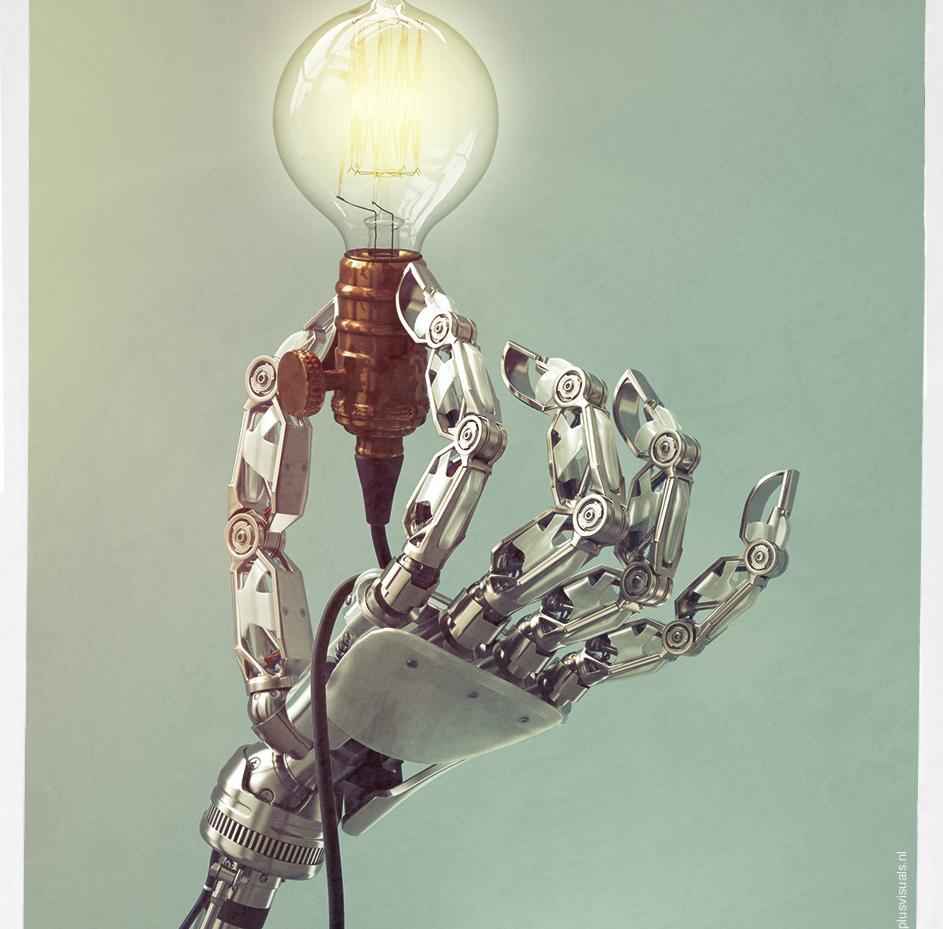}}&
\raisebox{-0.45\height}{\includegraphics[width=\teaserwid]{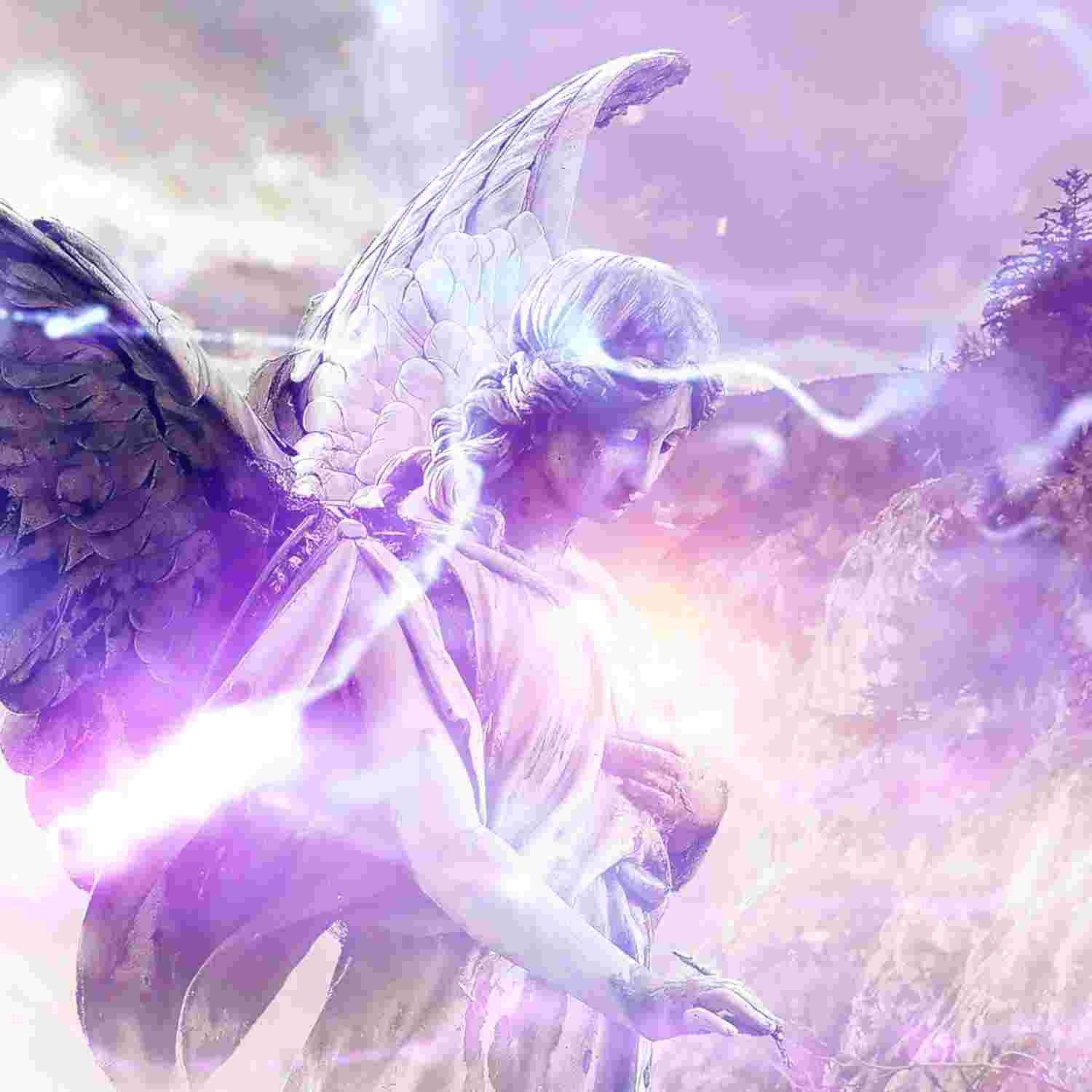}}&
\raisebox{-0.45\height}{\includegraphics[width=\teaserwid]{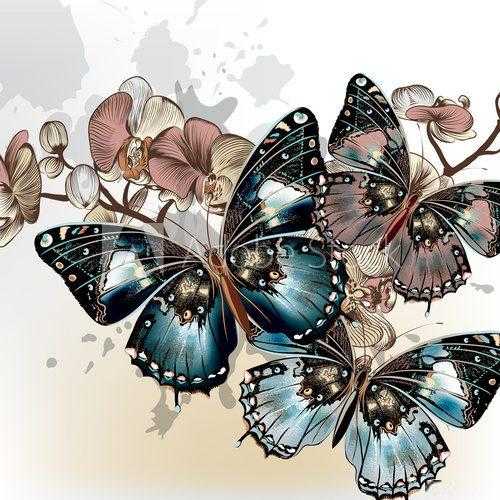}}&
\raisebox{-0.45\height}{\includegraphics[width=\teaserwid,height=\teaserwid]{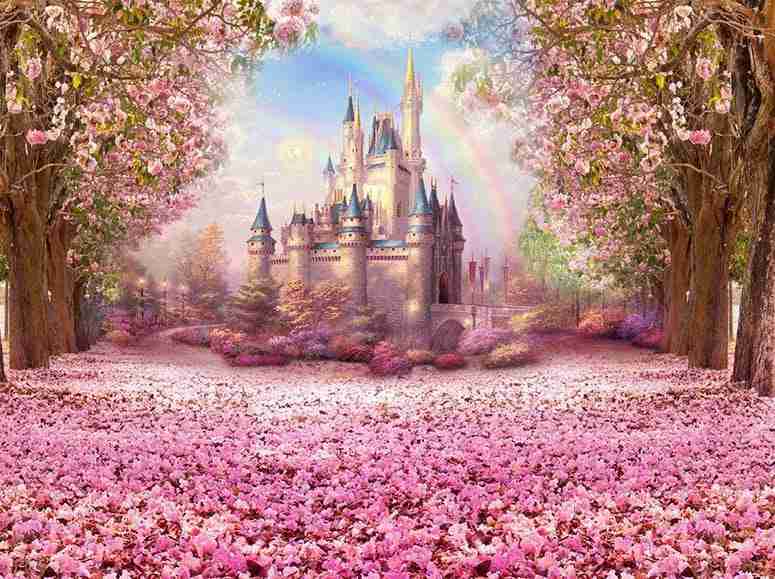}}&
\raisebox{-0.45\height}{\includegraphics[width=\teaserwid,height=\teaserwid]{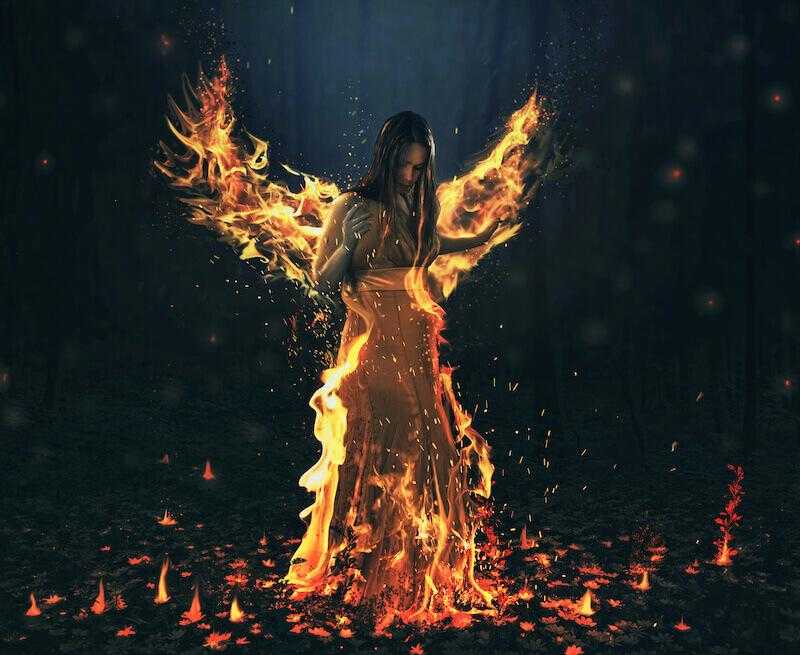}}&
\raisebox{-0.45\height}{\includegraphics[width=\teaserwid,height=\teaserwid]{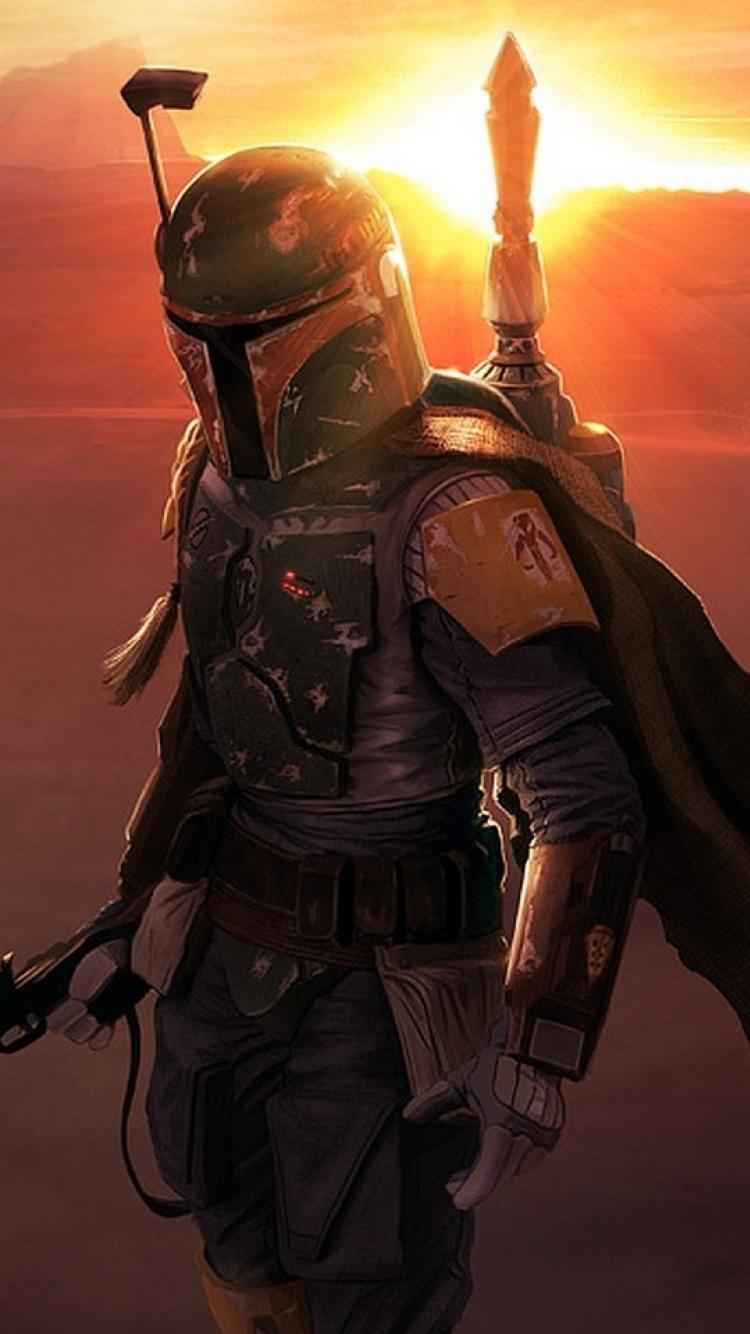}}&
\raisebox{-0.45\height}{\includegraphics[width=\teaserwid,height=\teaserwid]{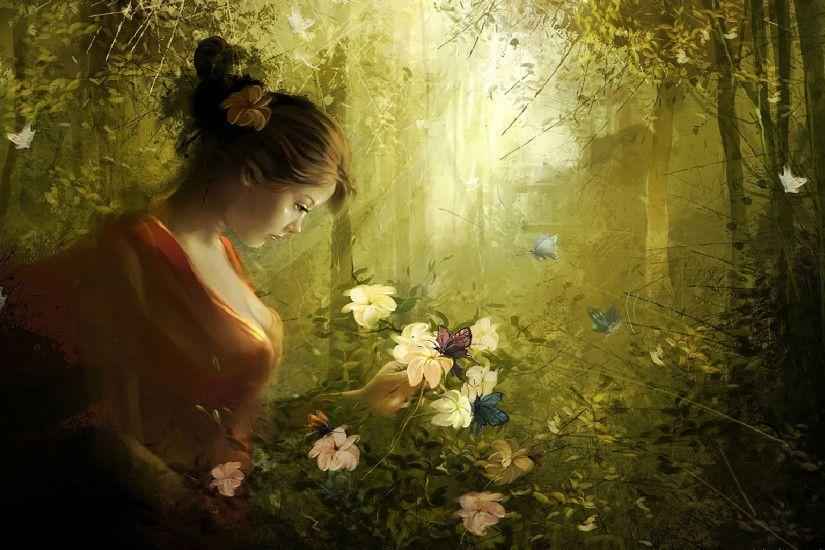}}&
\raisebox{-0.45\height}{\includegraphics[width=\teaserwid,height=\teaserwid]{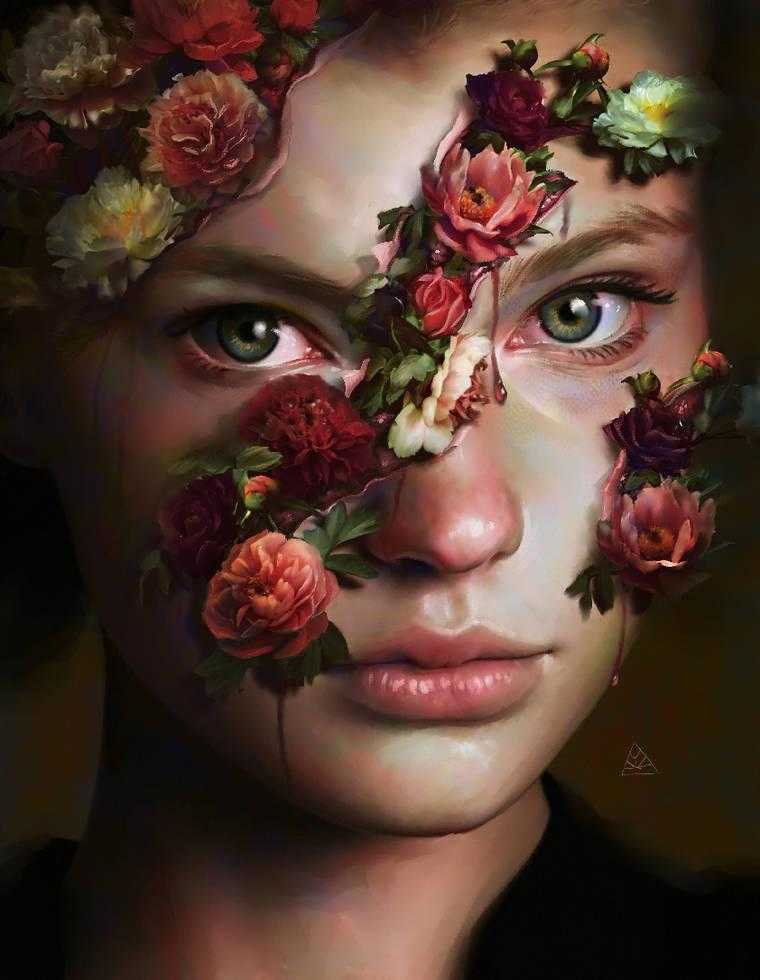}}
\vspace{.05in}
\\
\rotatebox[origin=c]{90}{StableDiff}&
\raisebox{-0.45\height}{\includegraphics[width=\teaserwid]{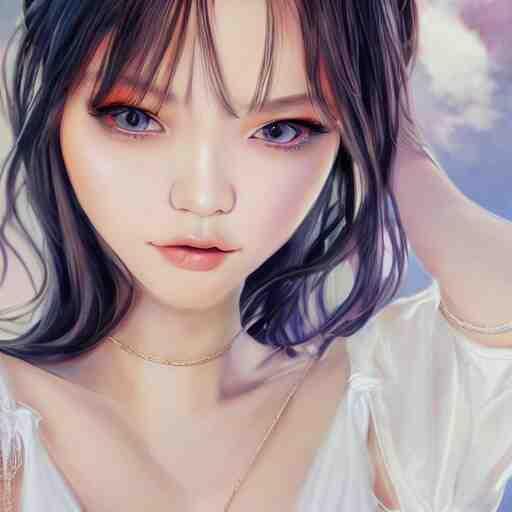}}& \raisebox{-0.45\height}{\includegraphics[width=\teaserwid]{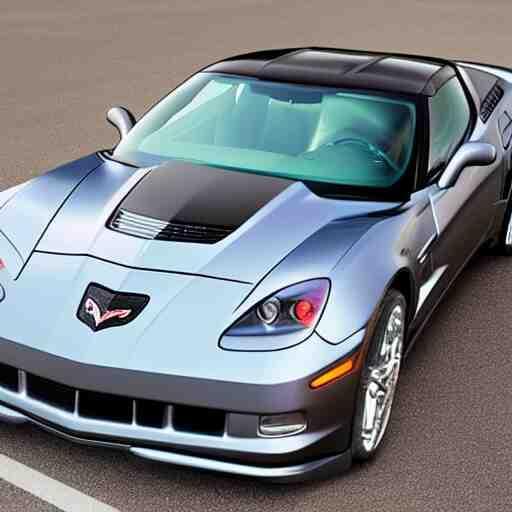}}&
\raisebox{-0.45\height}{\includegraphics[width=\teaserwid]{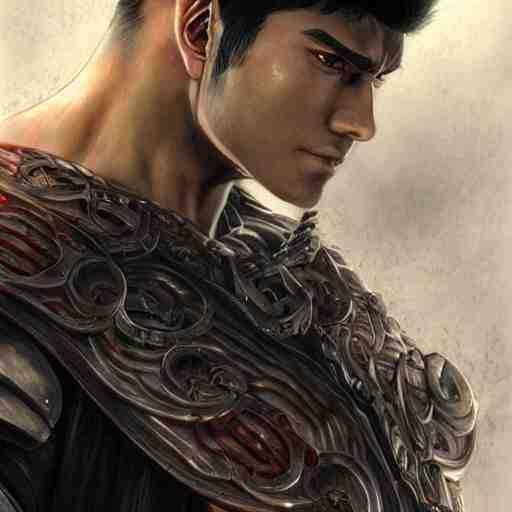}}&
\raisebox{-0.45\height}{\includegraphics[width=\teaserwid]{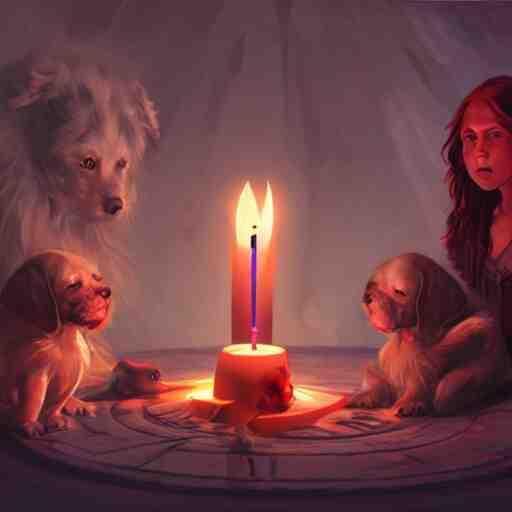}}&
\raisebox{-0.45\height}{\includegraphics[width=\teaserwid]{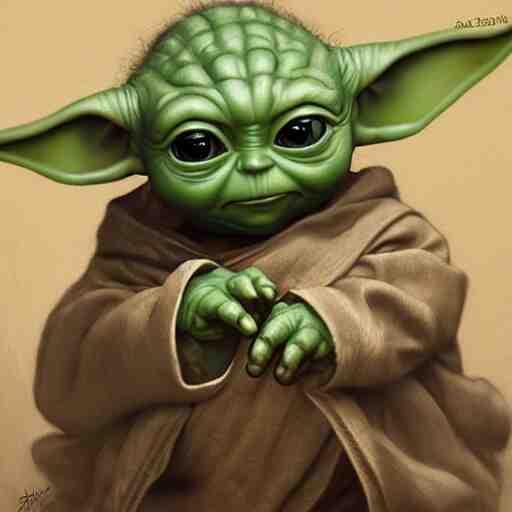}}&
\raisebox{-0.45\height}{\includegraphics[width=\teaserwid]{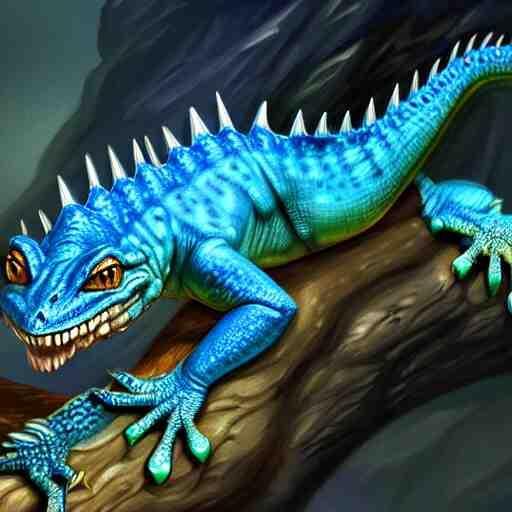}}&
\raisebox{-0.45\height}{\includegraphics[width=\teaserwid,height=\teaserwid]{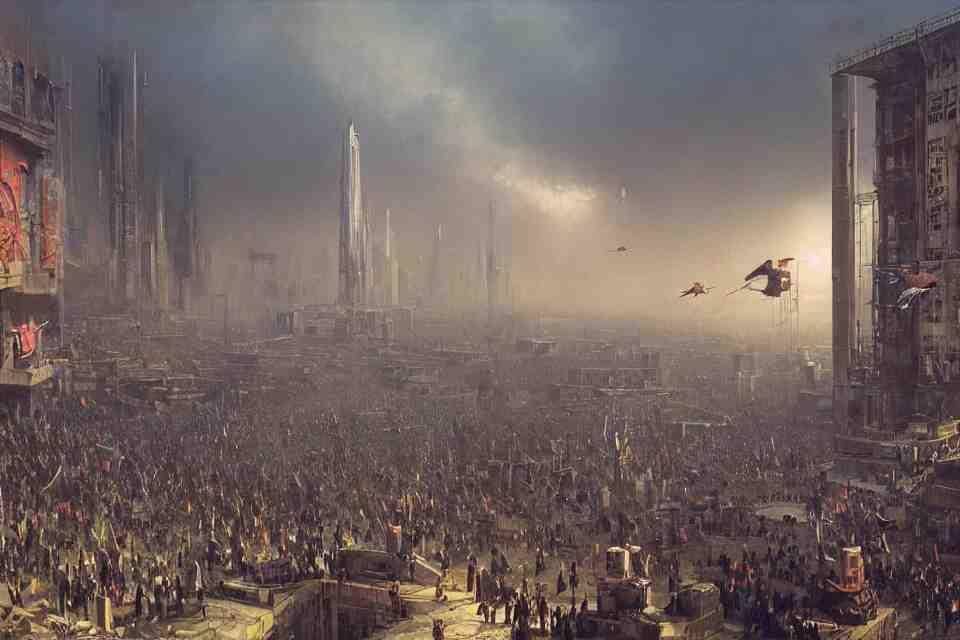}}&
\raisebox{-0.45\height}{\includegraphics[width=\teaserwid,height=\teaserwid]{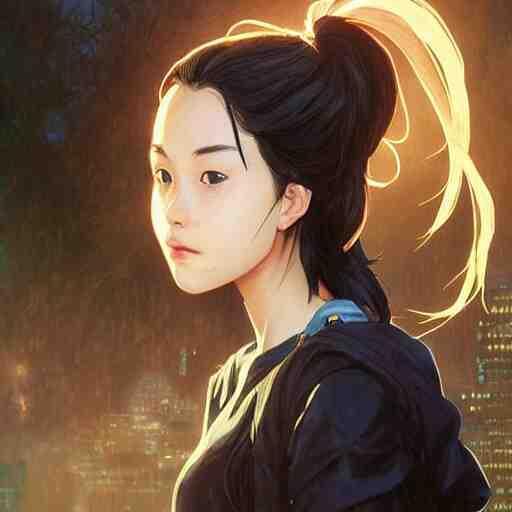}}&
\raisebox{-0.45\height}{\includegraphics[width=\teaserwid,height=\teaserwid]{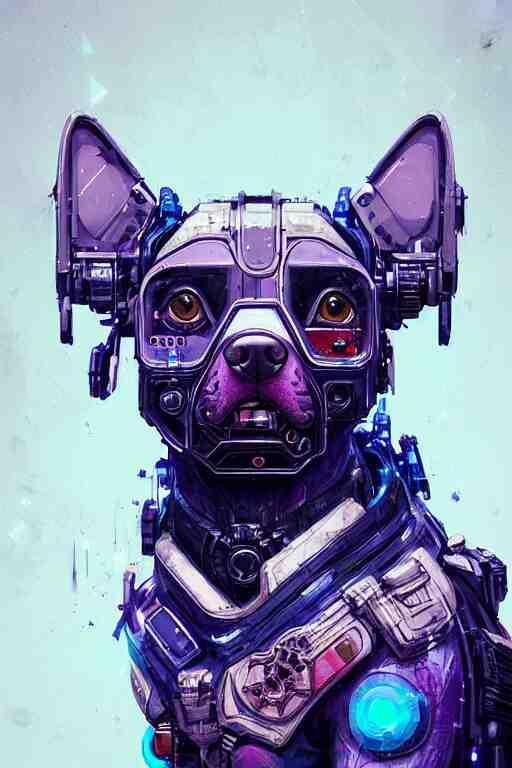}}&
\raisebox{-0.45\height}{\includegraphics[width=\teaserwid,height=\teaserwid]{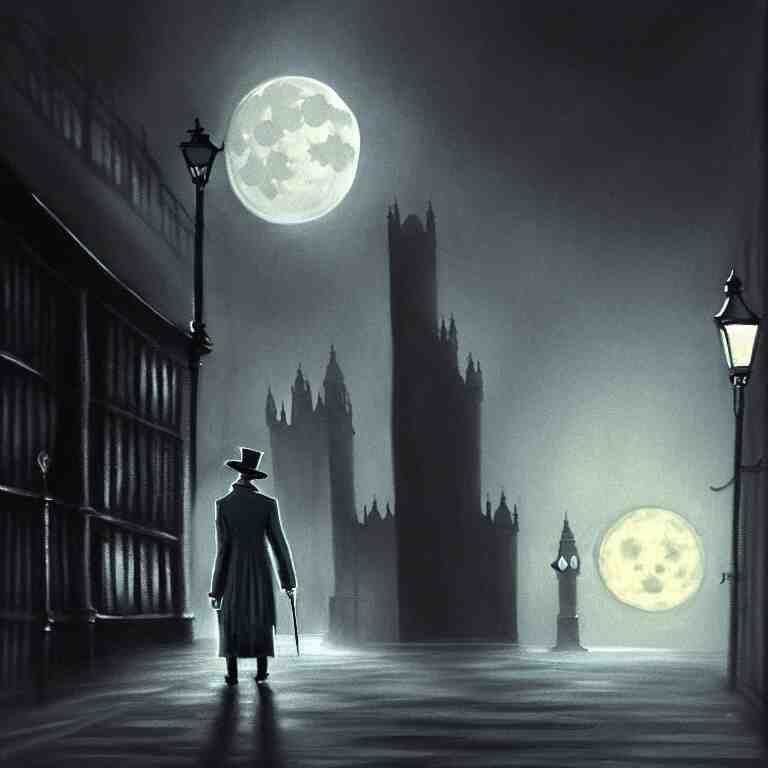}} &
\raisebox{-0.45\height}{\includegraphics[width=\teaserwid,height=\teaserwid]{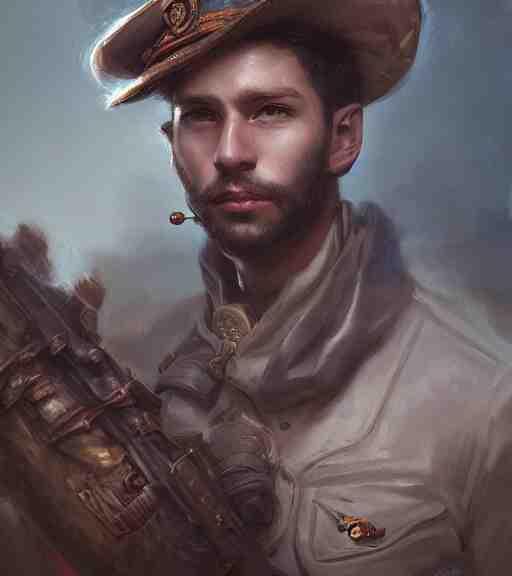}}
\vspace{.05in}
\\
\rotatebox[origin=c]{90}{DALL-E 2}&
\raisebox{-0.45\height}{\includegraphics[width=\teaserwid]{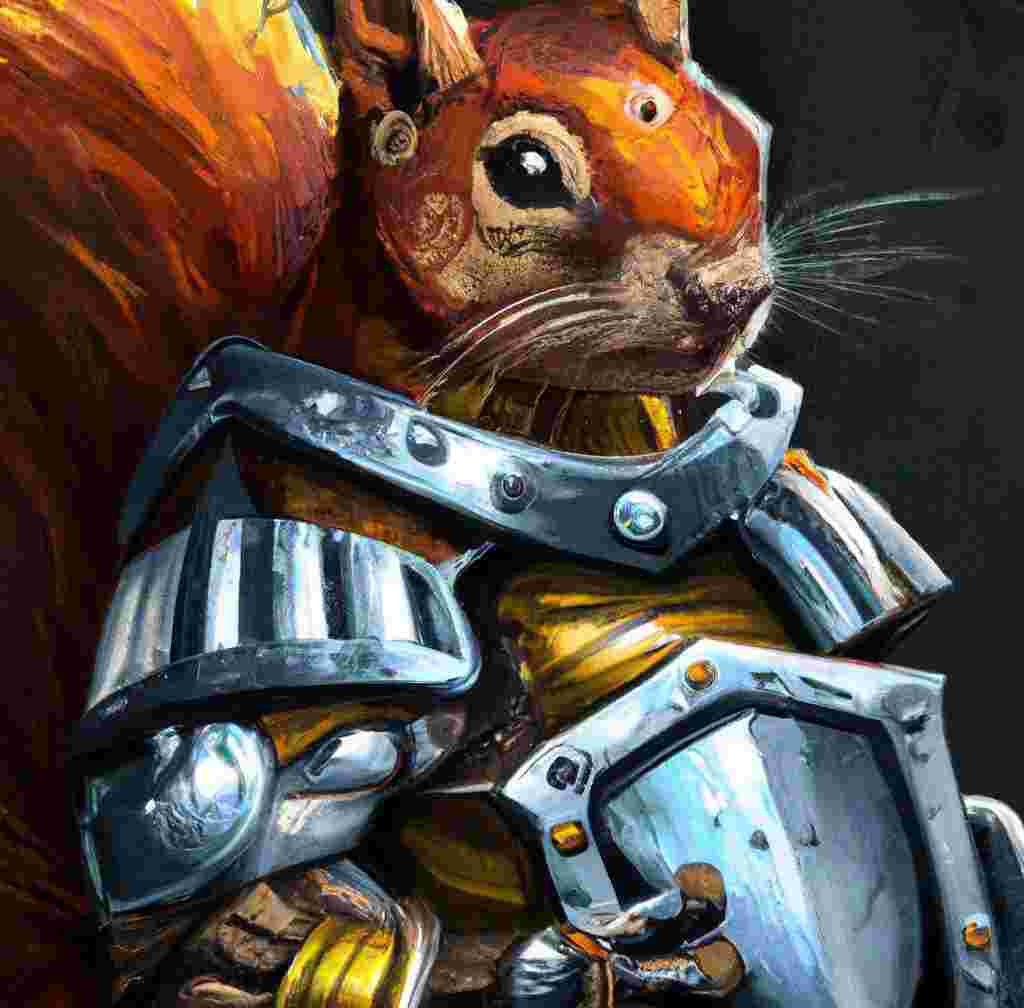}}&  
\raisebox{-0.45\height}{\includegraphics[width=\teaserwid]{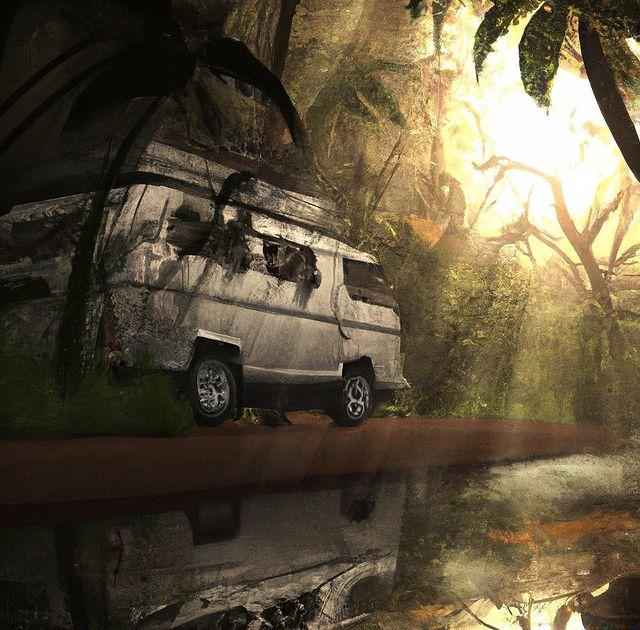}}&
\raisebox{-0.45\height}{\includegraphics[width=\teaserwid]{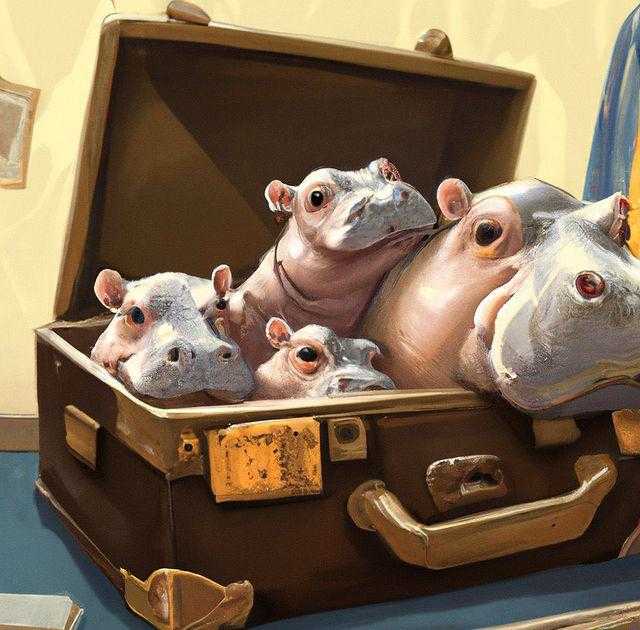}}&
\raisebox{-0.45\height}{\includegraphics[width=\teaserwid]{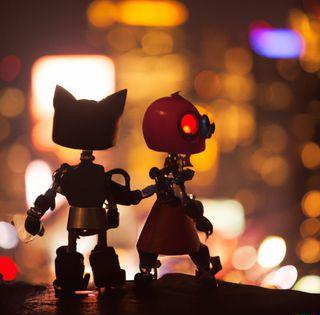}}&
\raisebox{-0.45\height}{\includegraphics[width=\teaserwid]{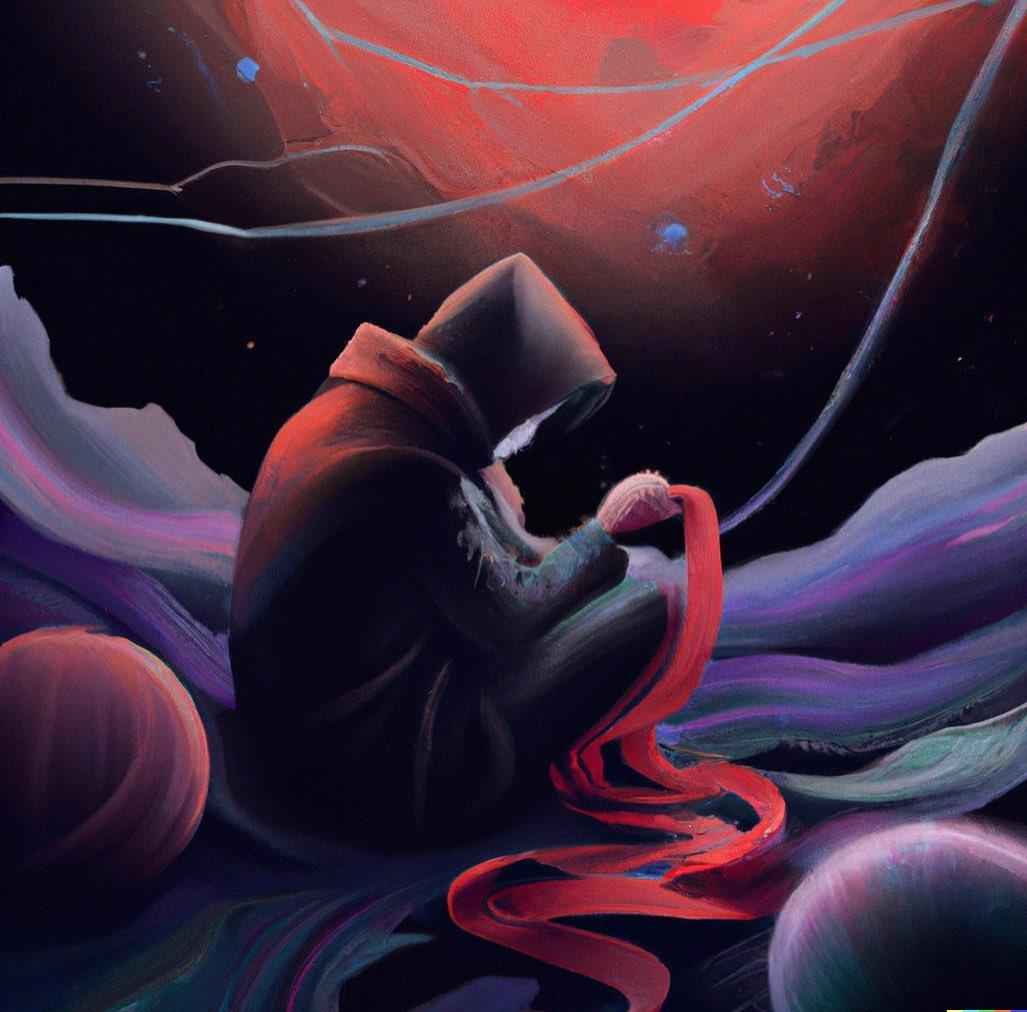}}&
\raisebox{-0.45\height}{\includegraphics[width=\teaserwid]{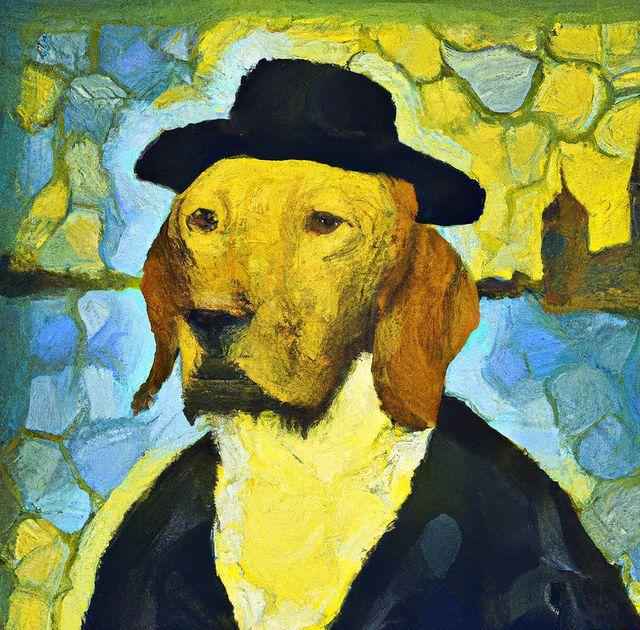}}&
\raisebox{-0.45\height}{\includegraphics[width=\teaserwid,height=\teaserwid]{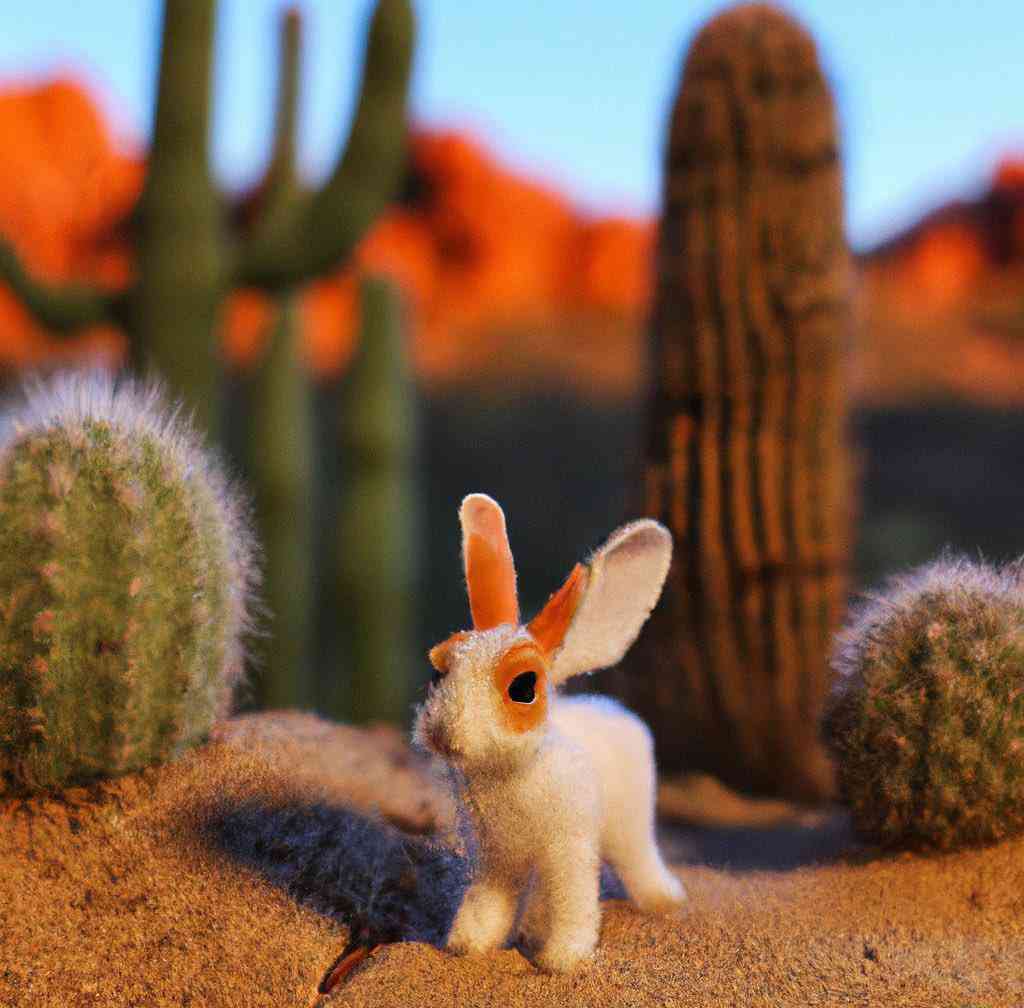}}&
\raisebox{-0.45\height}{\includegraphics[width=\teaserwid,height=\teaserwid]{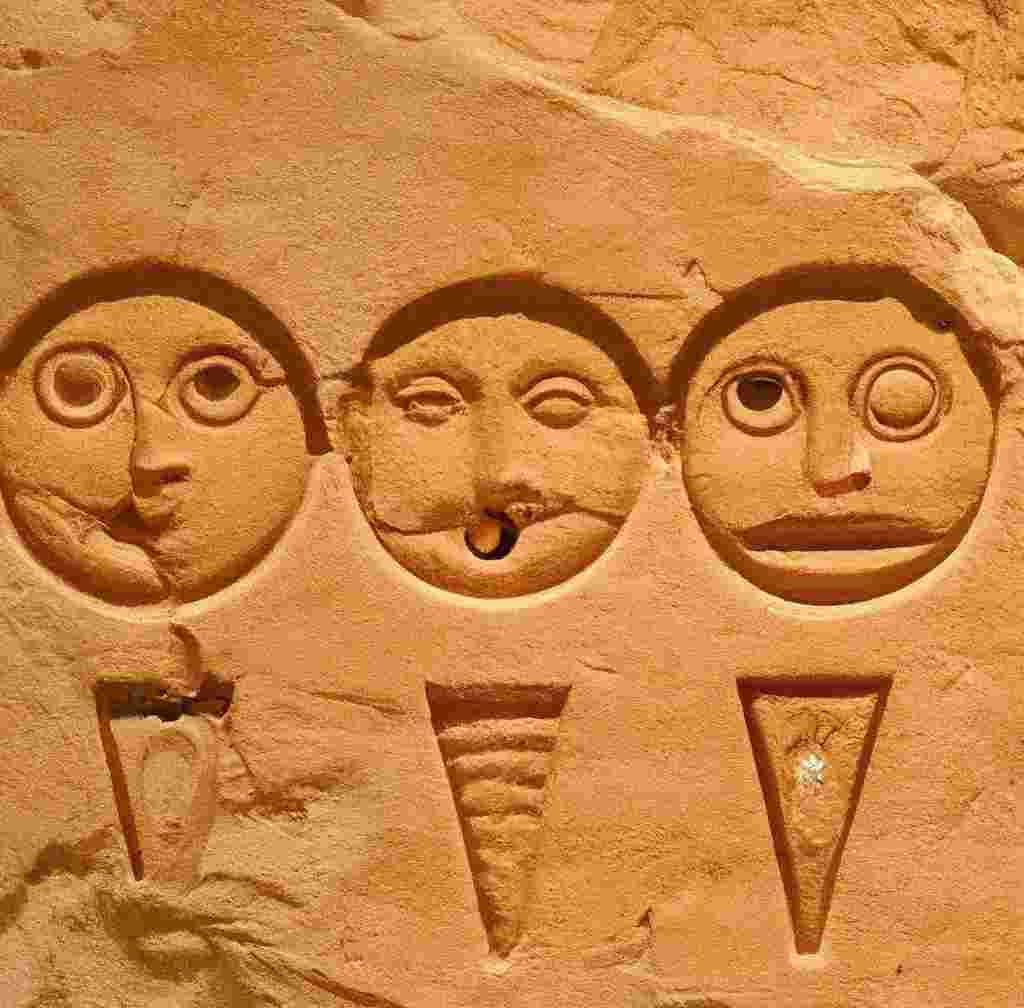}}&
\raisebox{-0.45\height}{\includegraphics[width=\teaserwid,height=\teaserwid]{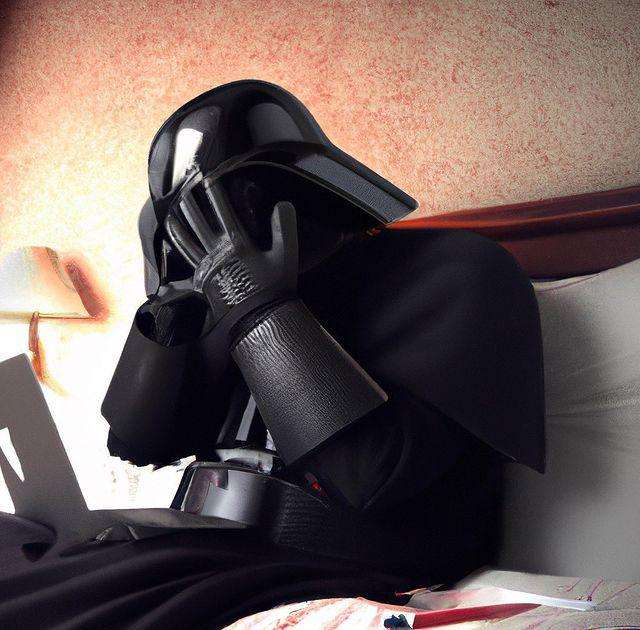}}&
\raisebox{-0.45\height}{\includegraphics[width=\teaserwid,height=\teaserwid]{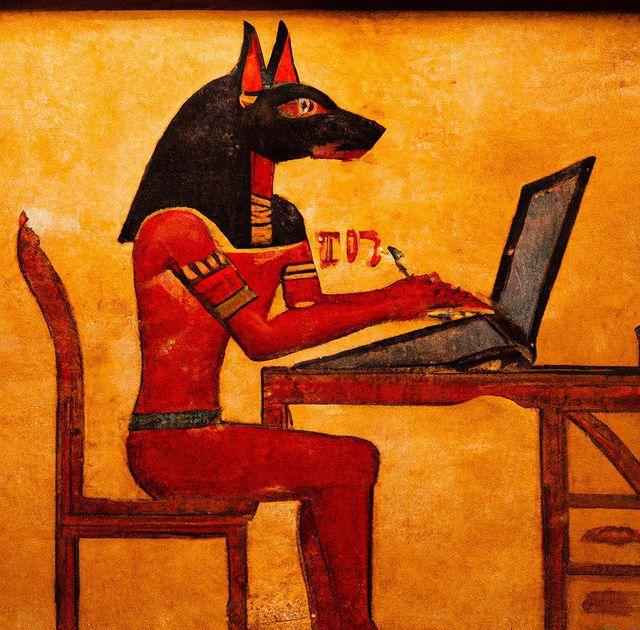}} &
\raisebox{-0.45\height}{\includegraphics[width=\teaserwid,height=\teaserwid]{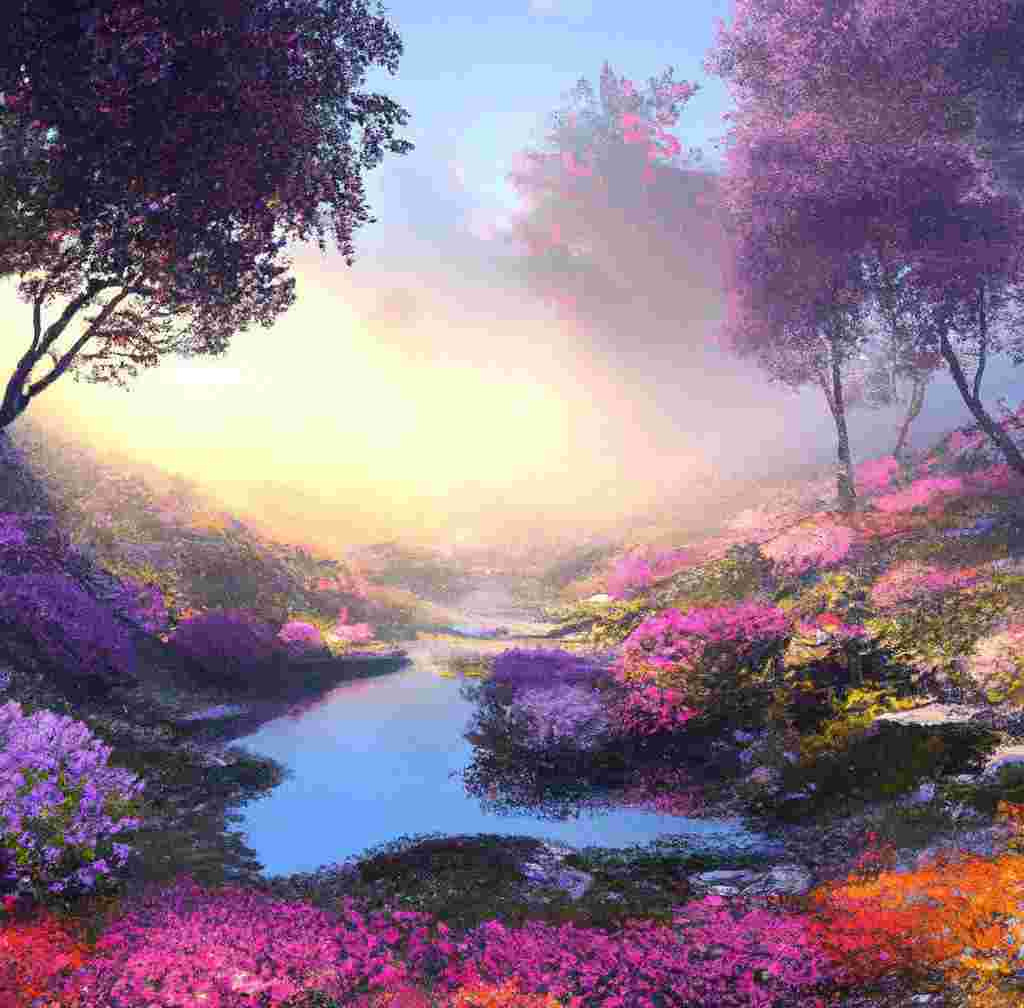}}
\vspace{.05in}
\\
\rotatebox[origin=c]{90}{Imagen}&
\raisebox{-0.45\height}{\includegraphics[width=\teaserwid]{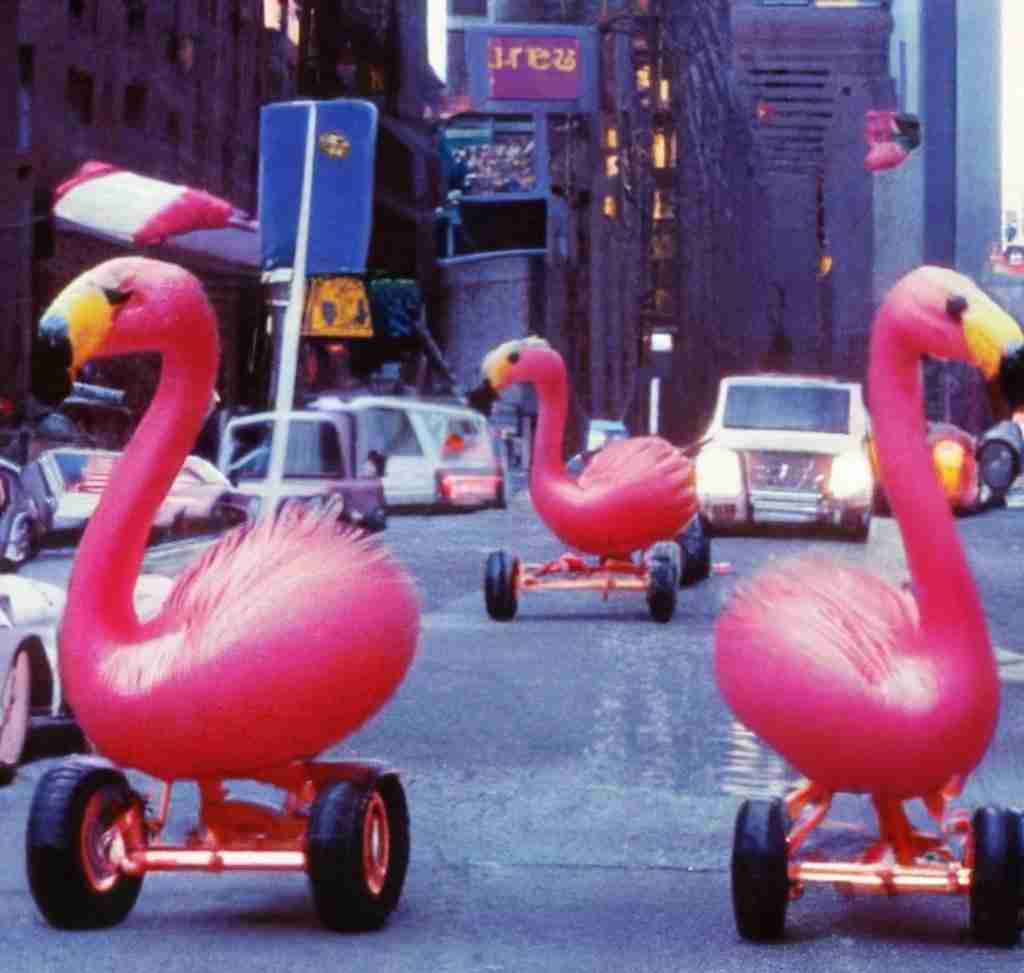}}&  
\raisebox{-0.45\height}{\includegraphics[width=\teaserwid]{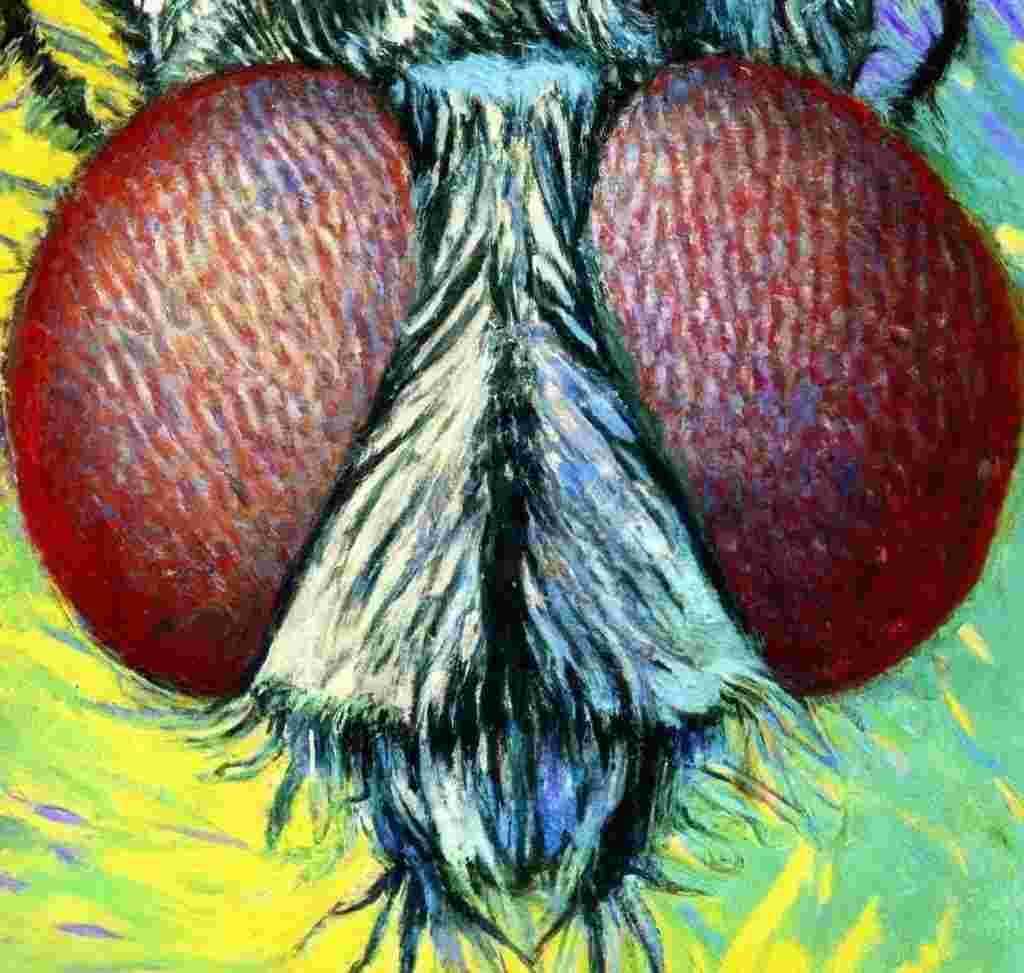}}&
\raisebox{-0.45\height}{\includegraphics[width=\teaserwid]{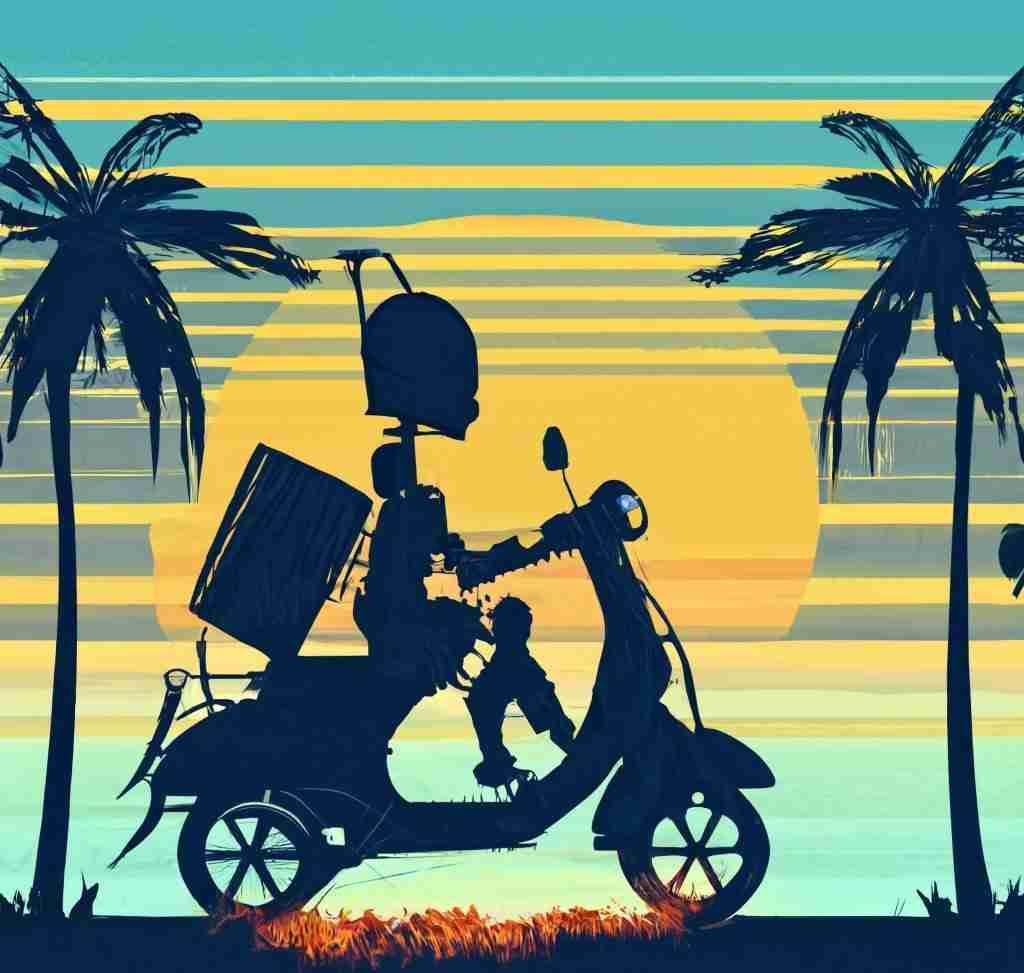}}&
\raisebox{-0.45\height}{\includegraphics[width=\teaserwid]{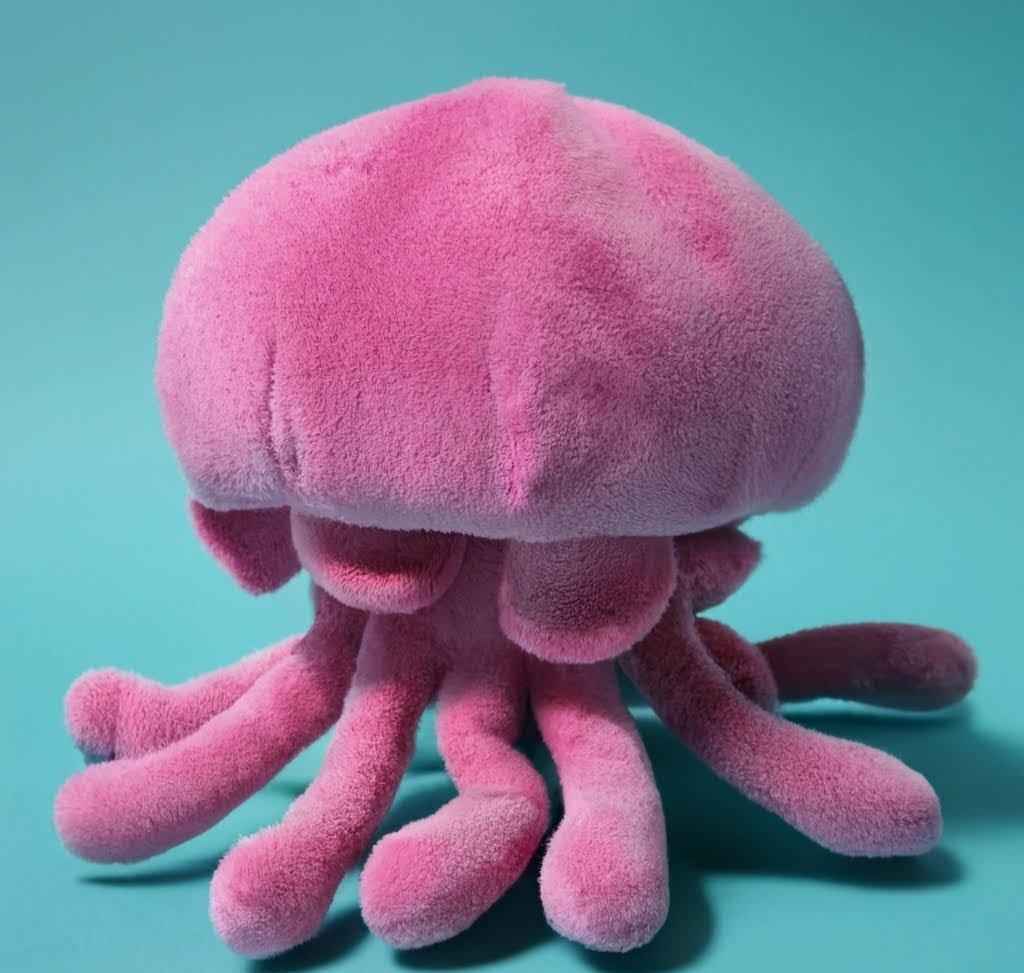}}&
\raisebox{-0.45\height}{\includegraphics[width=\teaserwid]{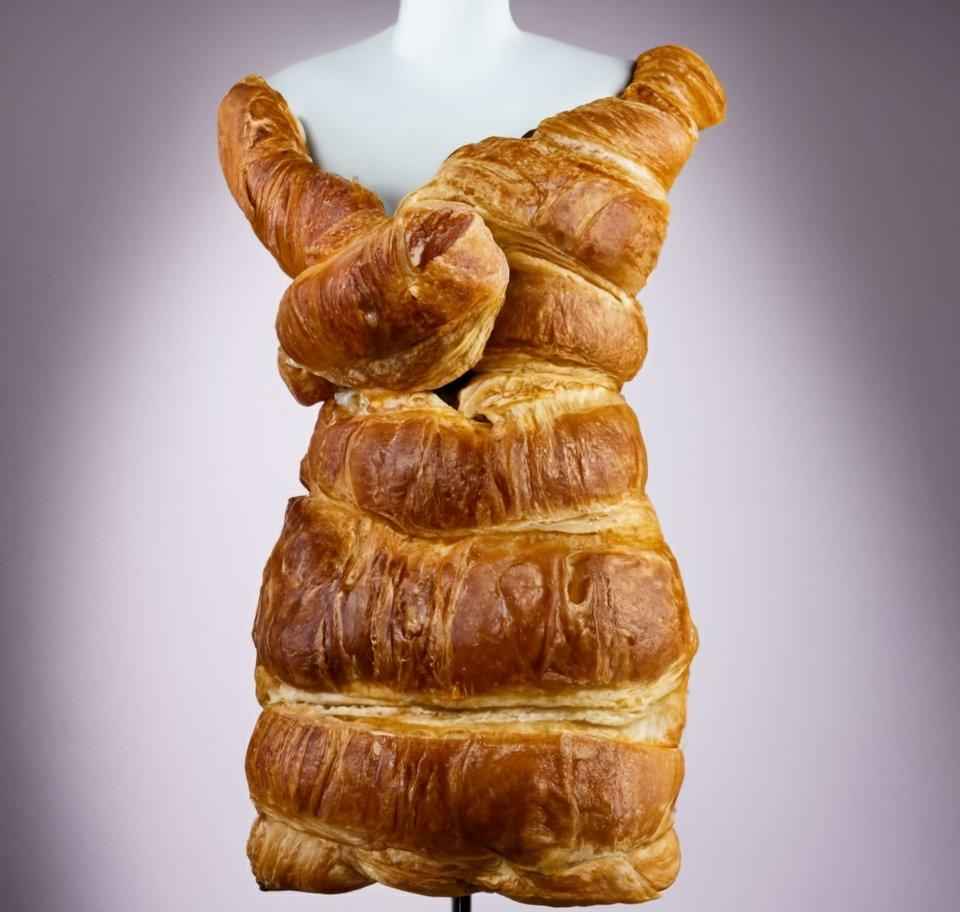}}&
\raisebox{-0.45\height}{\includegraphics[width=\teaserwid]{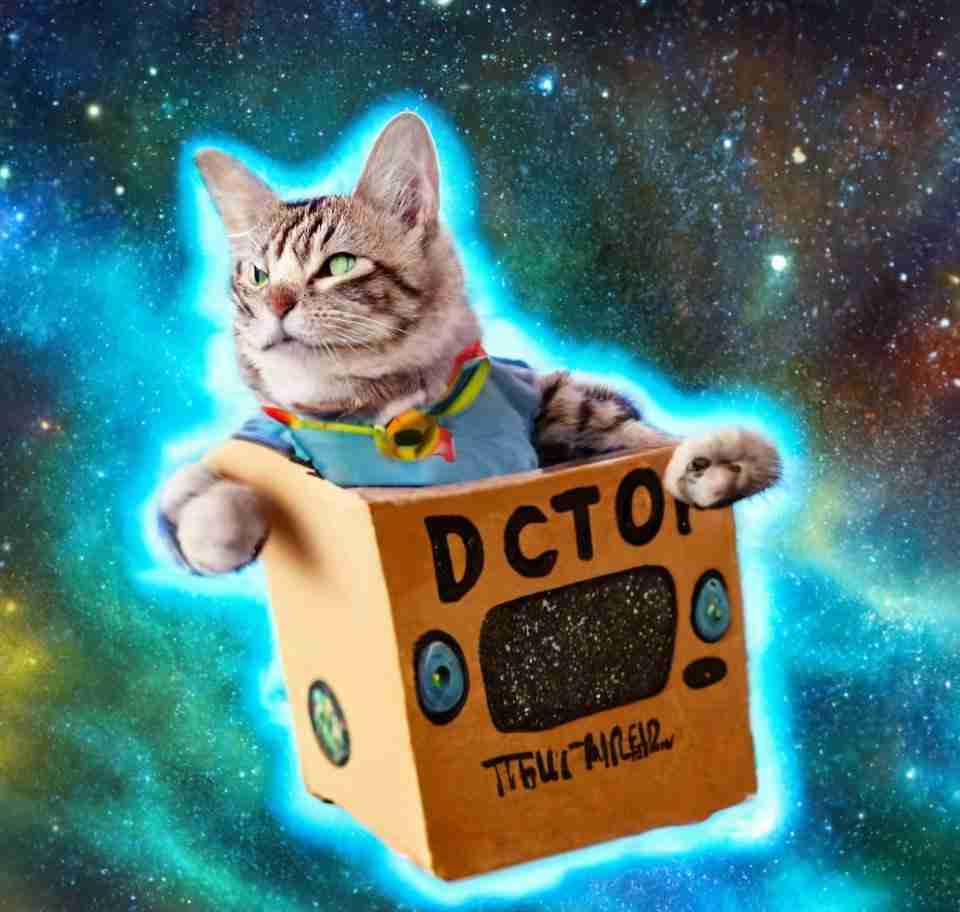}}&
\raisebox{-0.45\height}{\includegraphics[width=\teaserwid,height=\teaserwid]{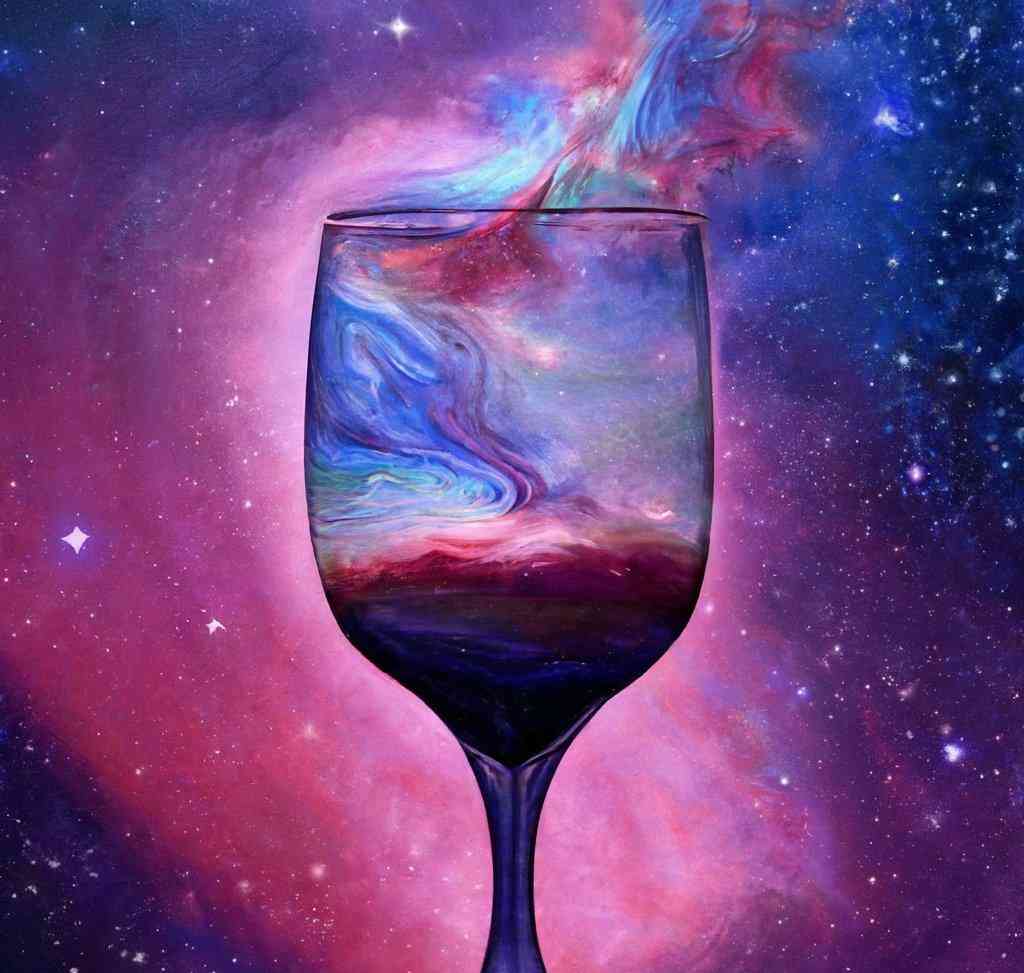}}&
\raisebox{-0.45\height}{\includegraphics[width=\teaserwid,height=\teaserwid]{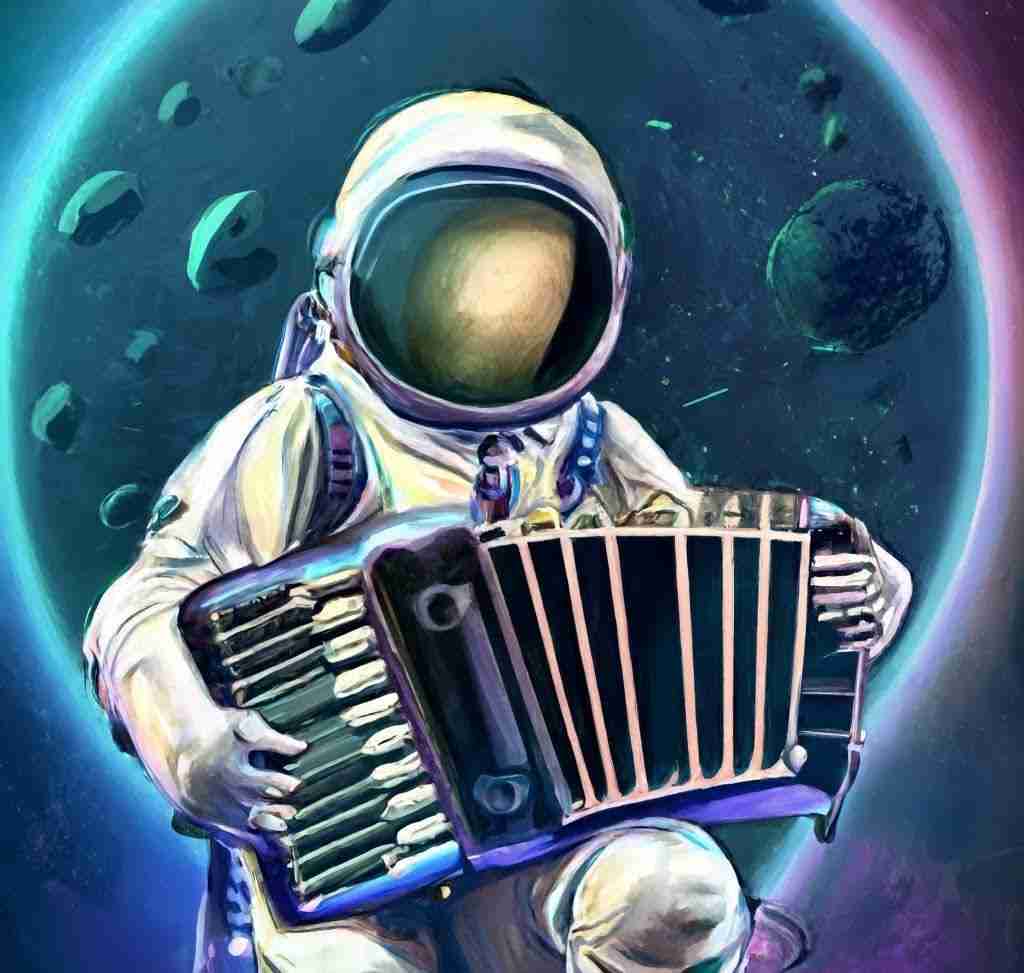}}&
\raisebox{-0.45\height}{\includegraphics[width=\teaserwid,height=\teaserwid]{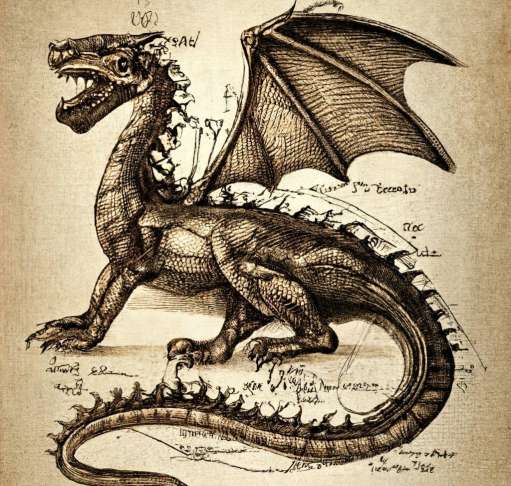}}&
\raisebox{-0.45\height}{\includegraphics[width=\teaserwid,height=\teaserwid]{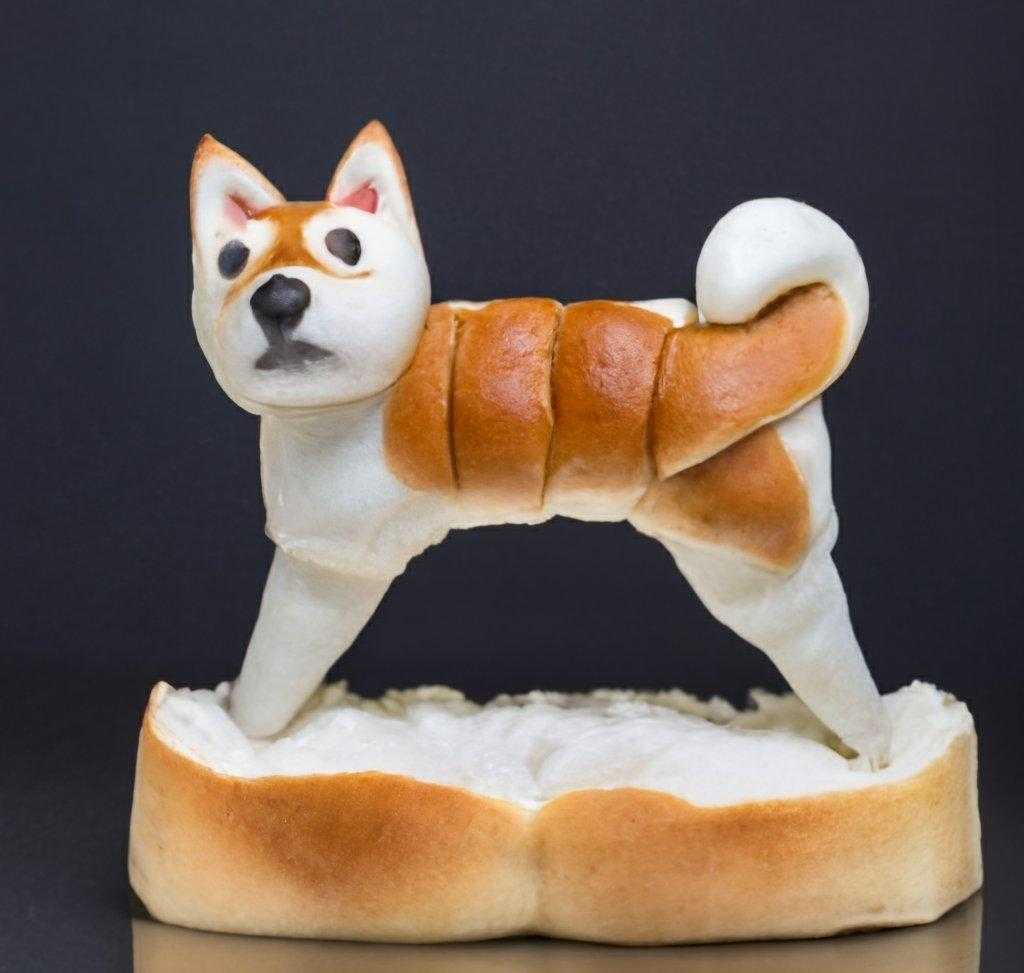}} &
\raisebox{-0.45\height}{\includegraphics[width=\teaserwid,height=\teaserwid]{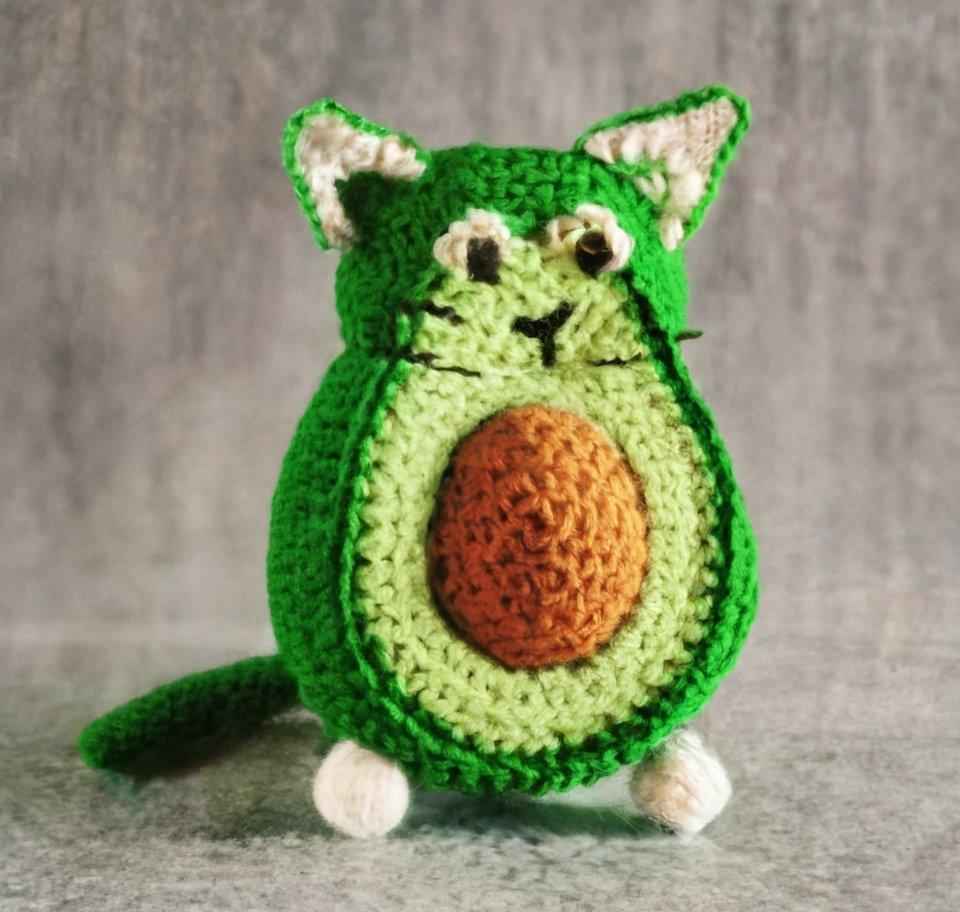}}
% \vspace{.05in}
\\
\rotatebox[origin=c]{90}{Midjourney}&
\raisebox{-0.45\height}{\includegraphics[width=\teaserwid]{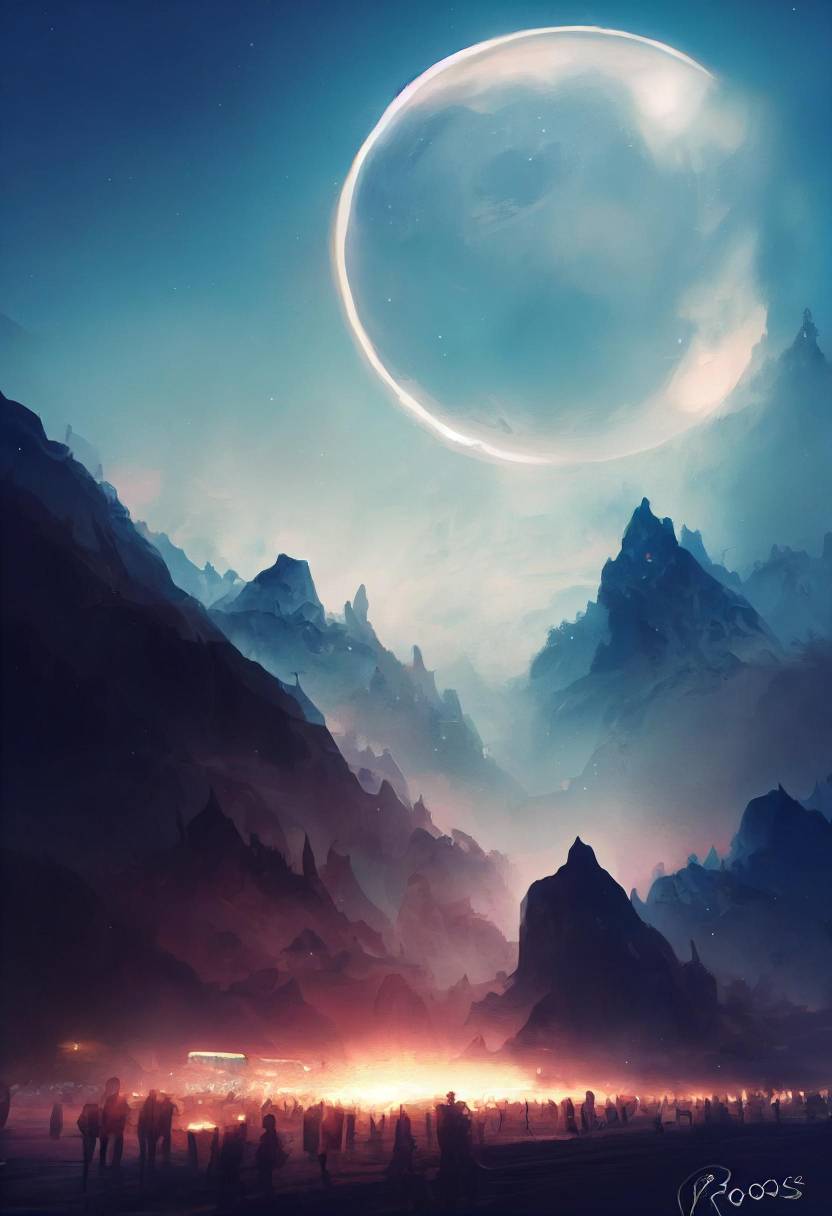}}&  
\raisebox{-0.45\height}{\includegraphics[width=\teaserwid]{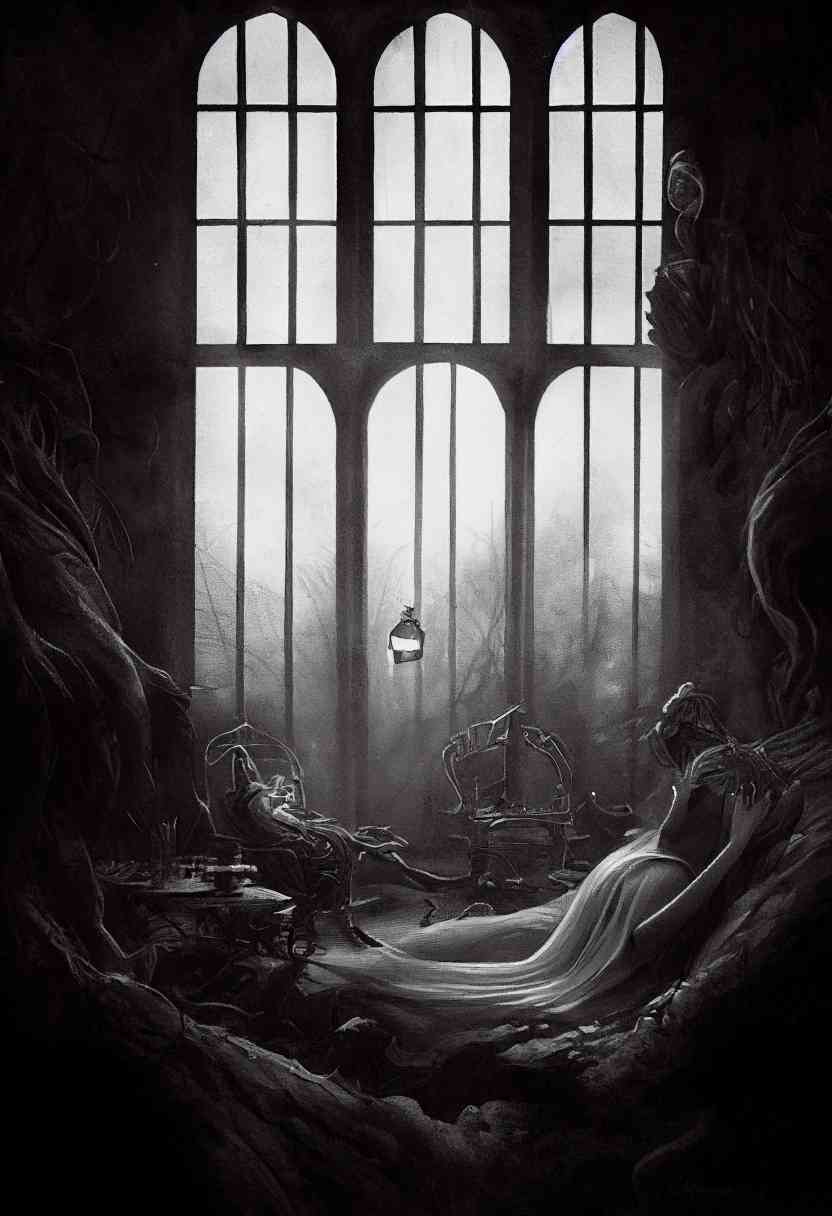}}&
\raisebox{-0.45\height}{\includegraphics[width=\teaserwid]{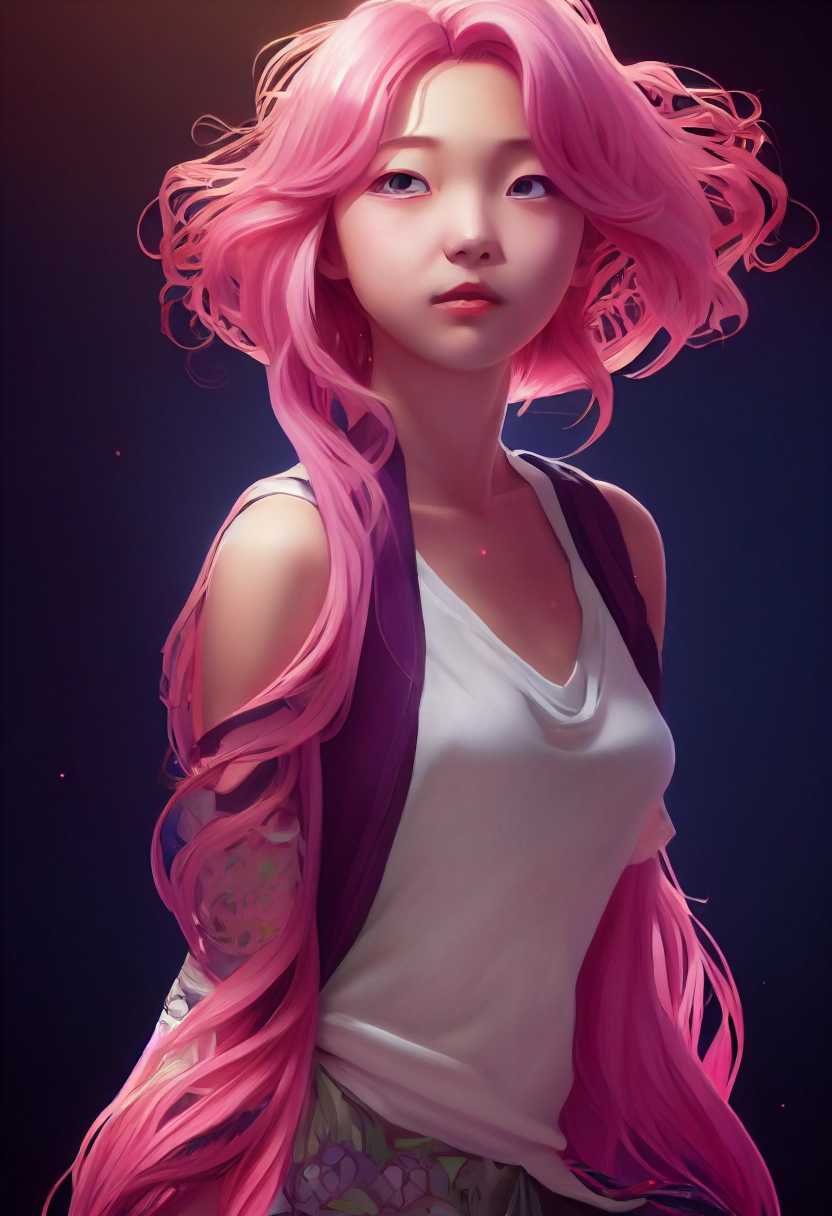}}&
\raisebox{-0.45\height}{\includegraphics[width=\teaserwid]{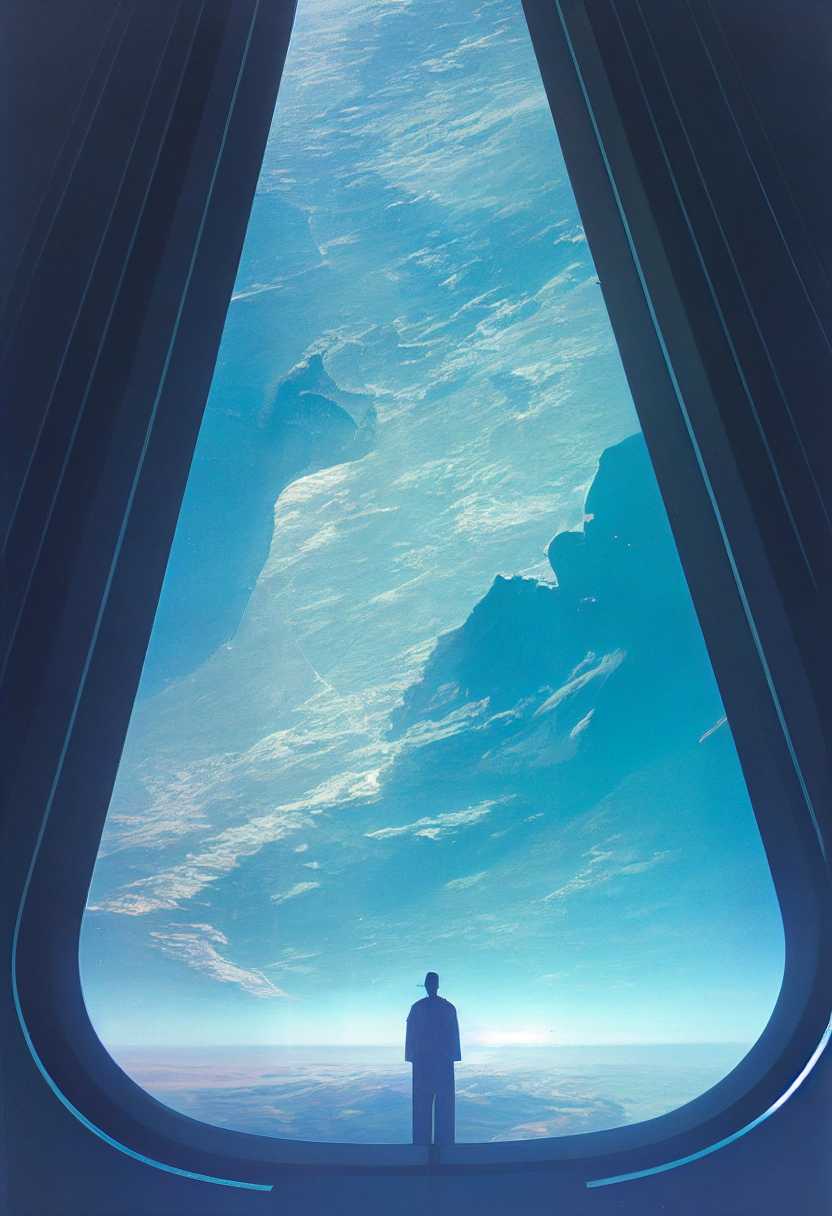}}&
\raisebox{-0.45\height}{\includegraphics[width=\teaserwid]{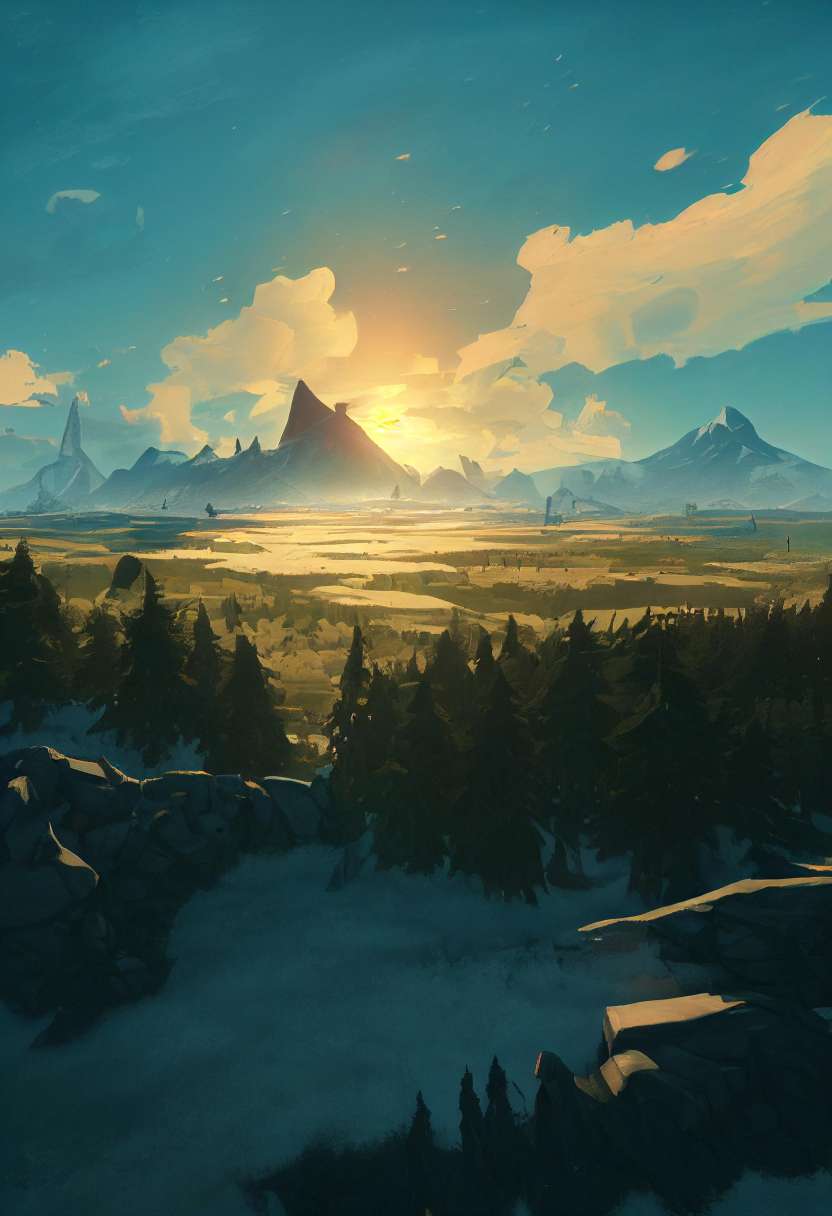}}&
\raisebox{-0.45\height}{\includegraphics[width=\teaserwid]{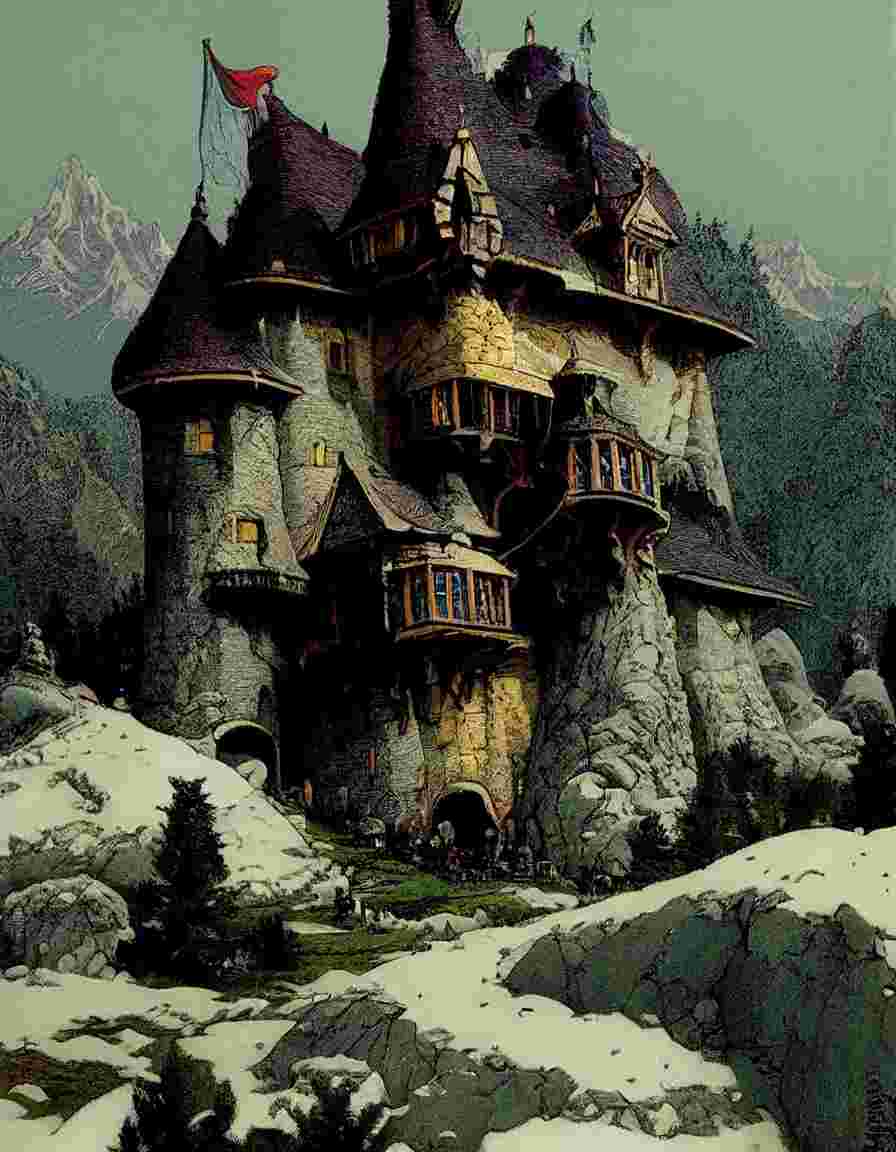}}&
\raisebox{-0.45\height}{\includegraphics[width=\teaserwid,height=\teaserwid]{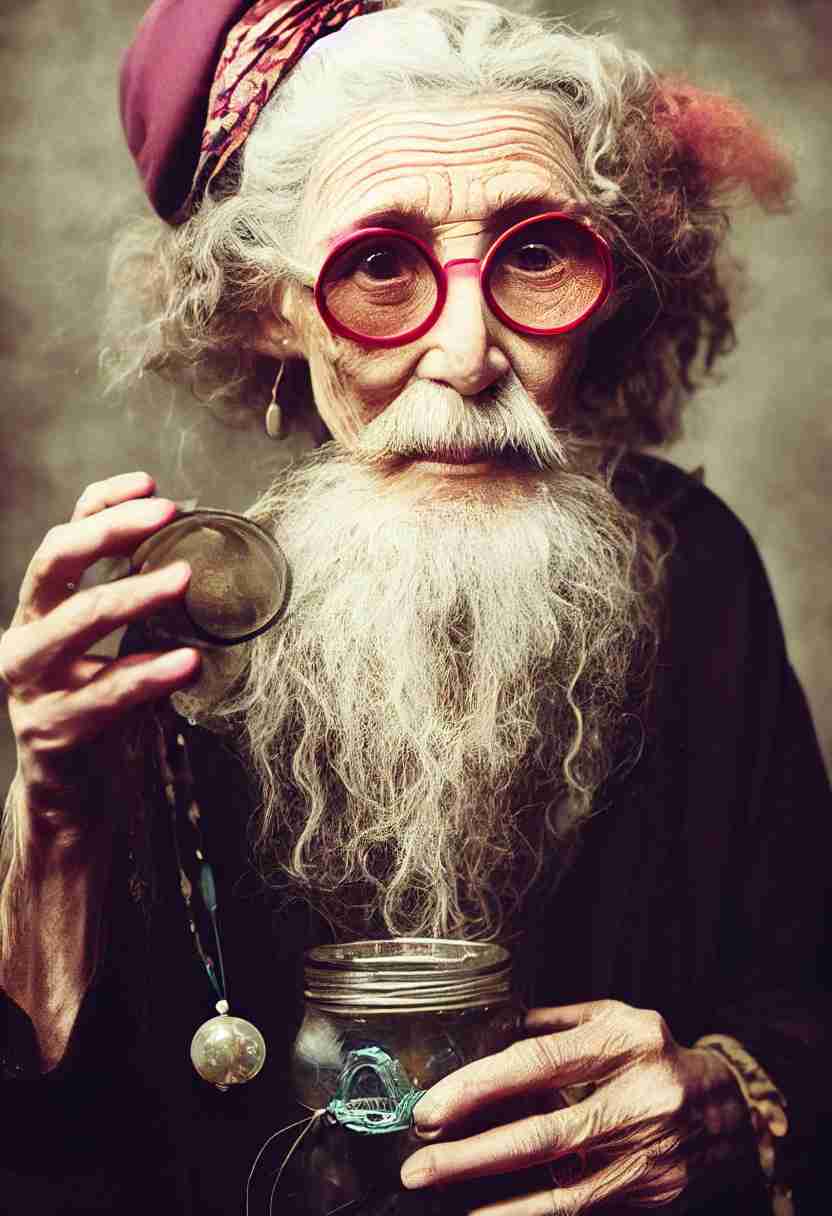}}&
\raisebox{-0.45\height}{\includegraphics[width=\teaserwid,height=\teaserwid]{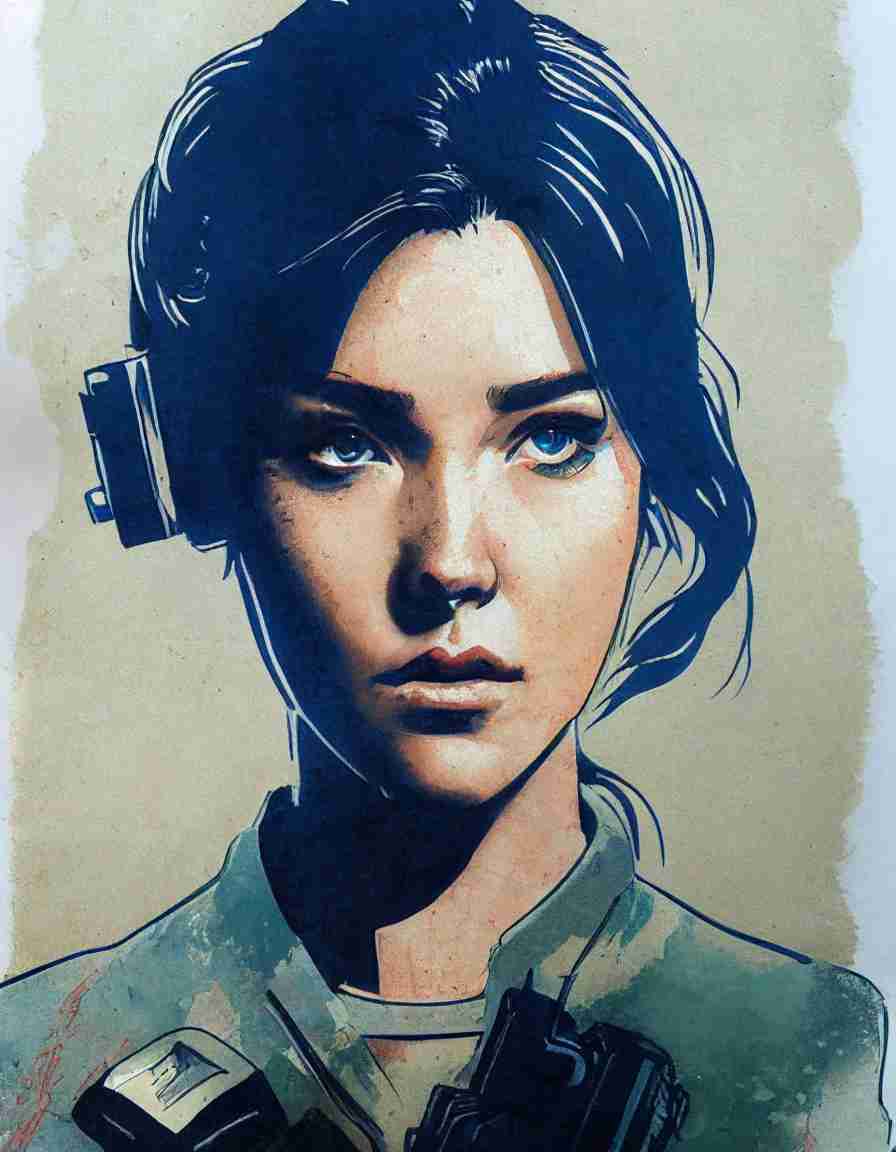}}&
\raisebox{-0.45\height}{\includegraphics[width=\teaserwid,height=\teaserwid]{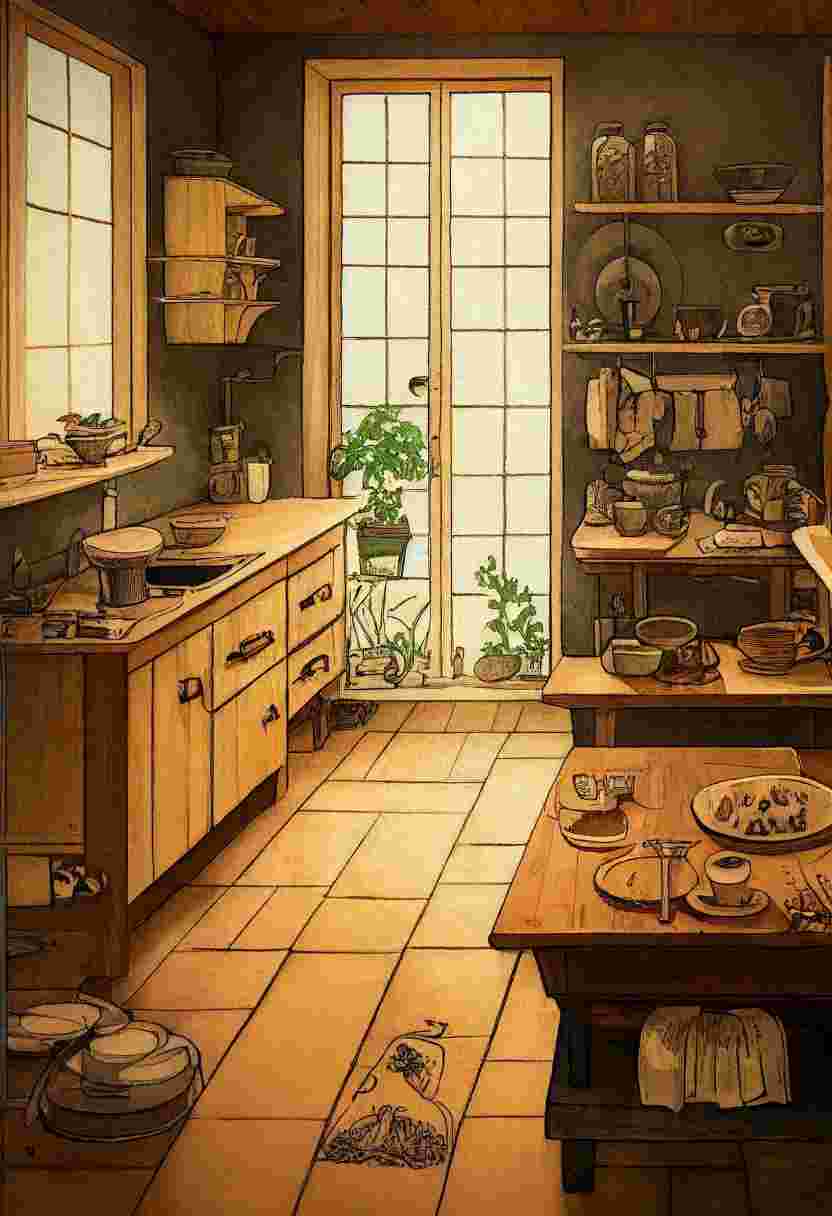}}&
\raisebox{-0.45\height}{\includegraphics[width=\teaserwid,height=\teaserwid]{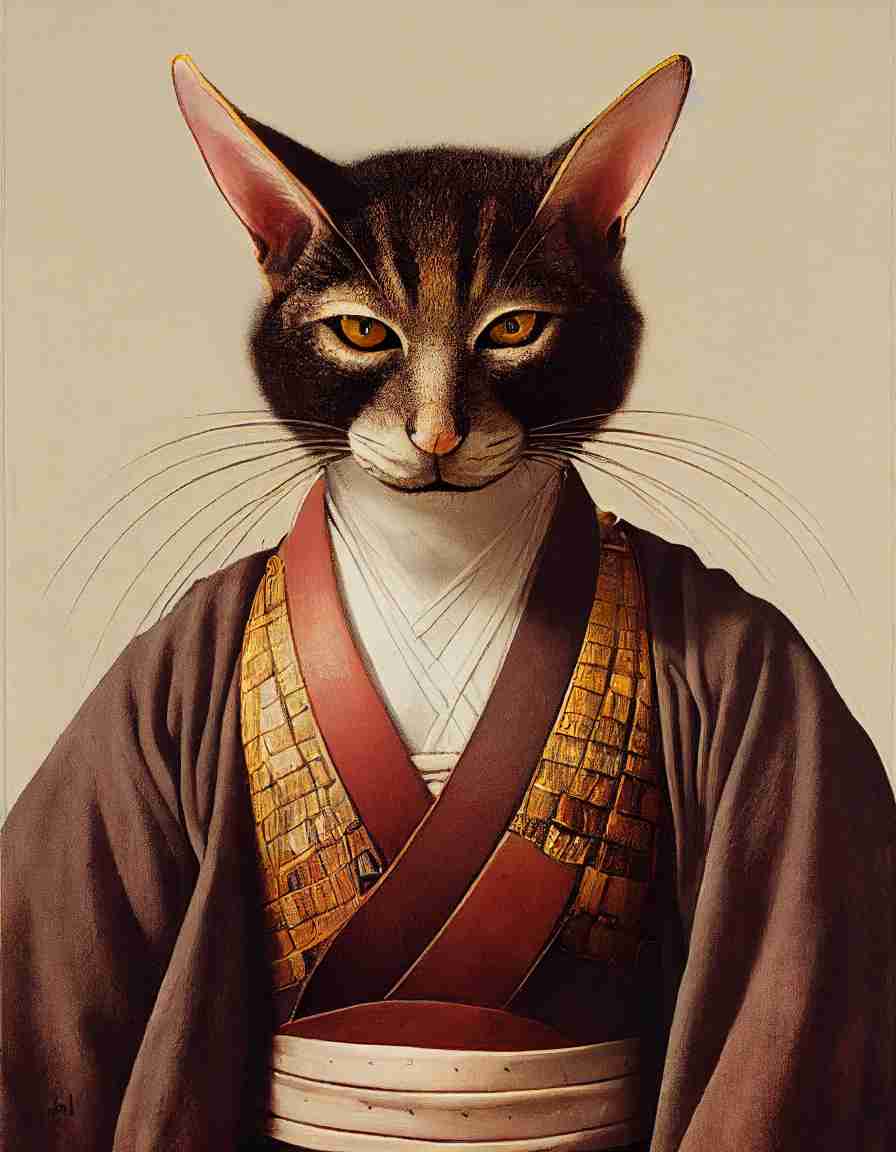}} &
\raisebox{-0.45\height}{\includegraphics[width=\teaserwid,height=\teaserwid]{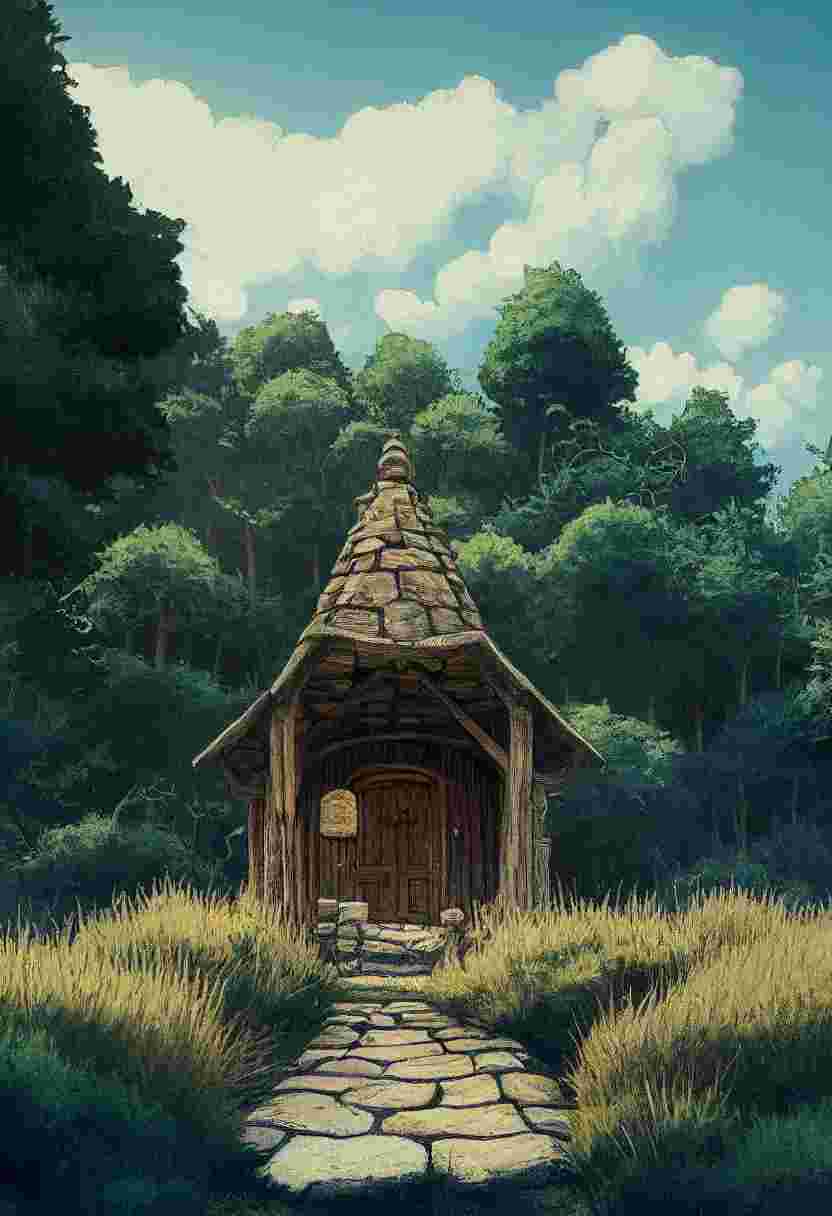}}
\vspace{.05in}
\\
\rotatebox[origin=c]{90}{Parti}&
\raisebox{-0.45\height}{\includegraphics[width=\teaserwid]{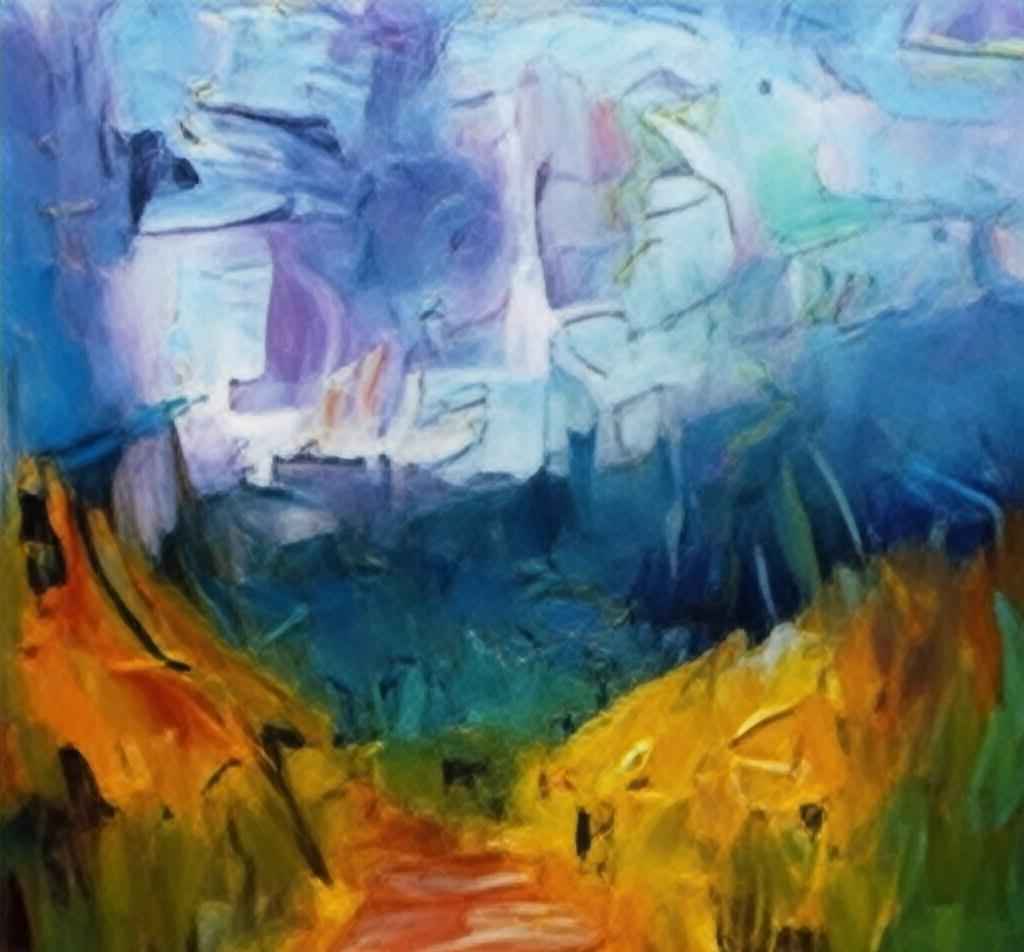}}&  
\raisebox{-0.45\height}{\includegraphics[width=\teaserwid]{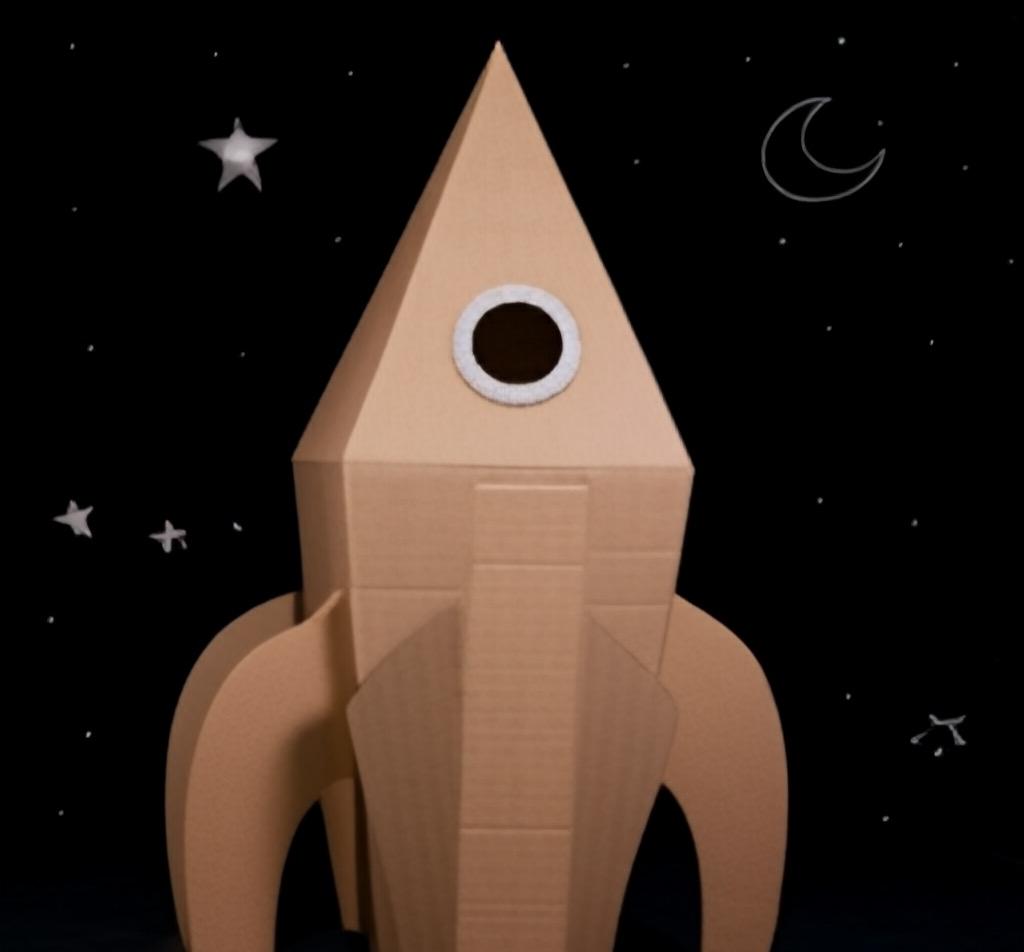}}&
\raisebox{-0.45\height}{\includegraphics[width=\teaserwid]{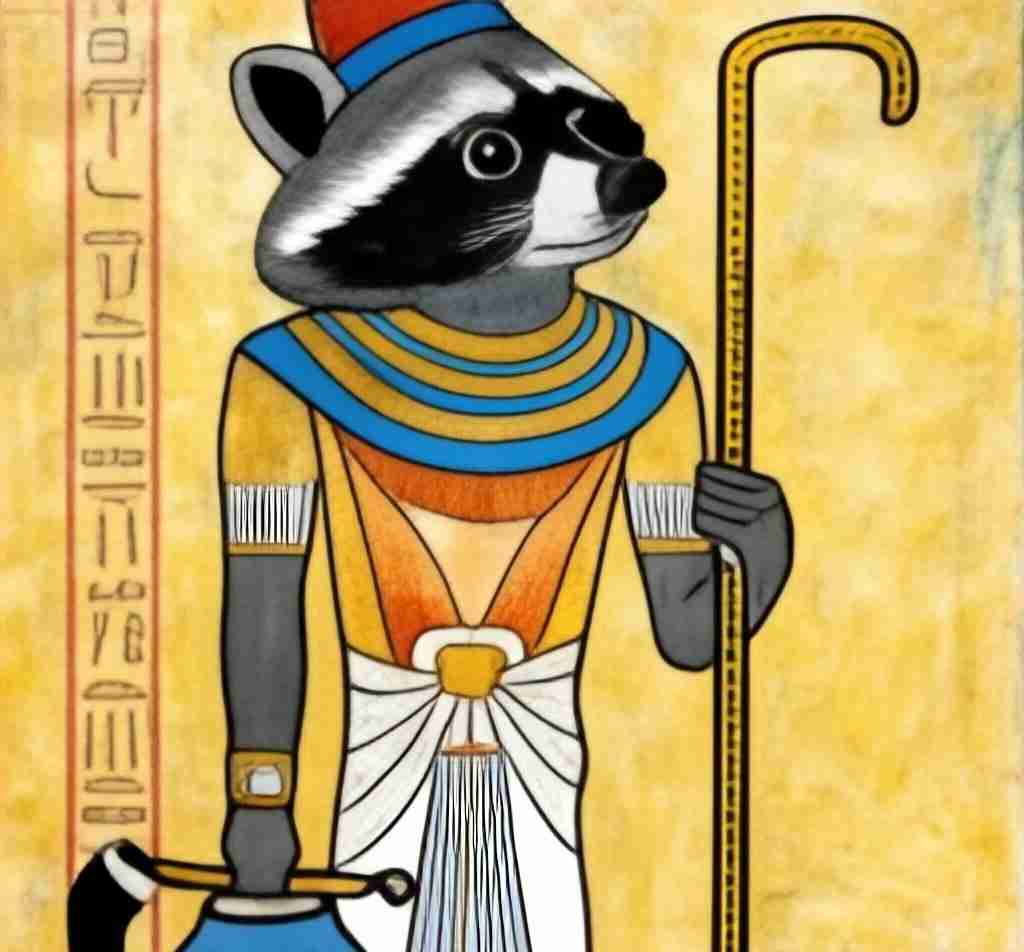}}&
\raisebox{-0.45\height}{\includegraphics[width=\teaserwid]{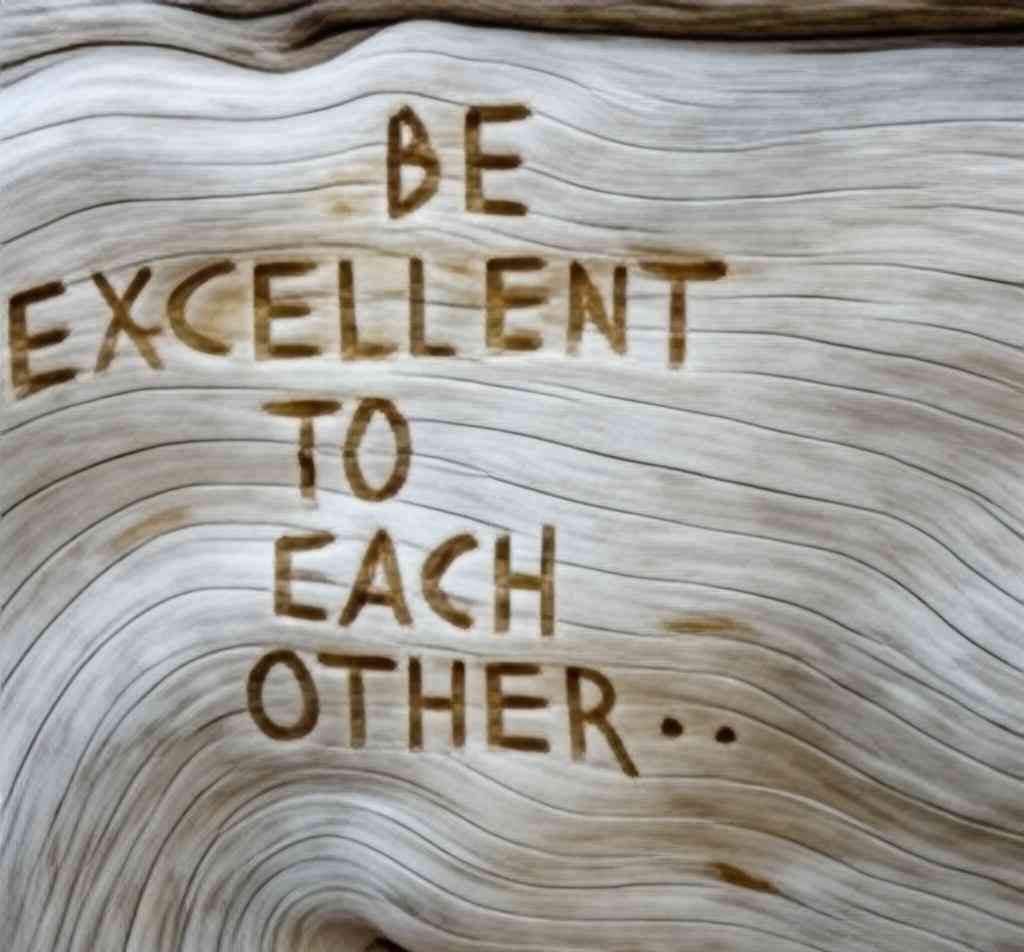}}&
\raisebox{-0.45\height}{\includegraphics[width=\teaserwid]{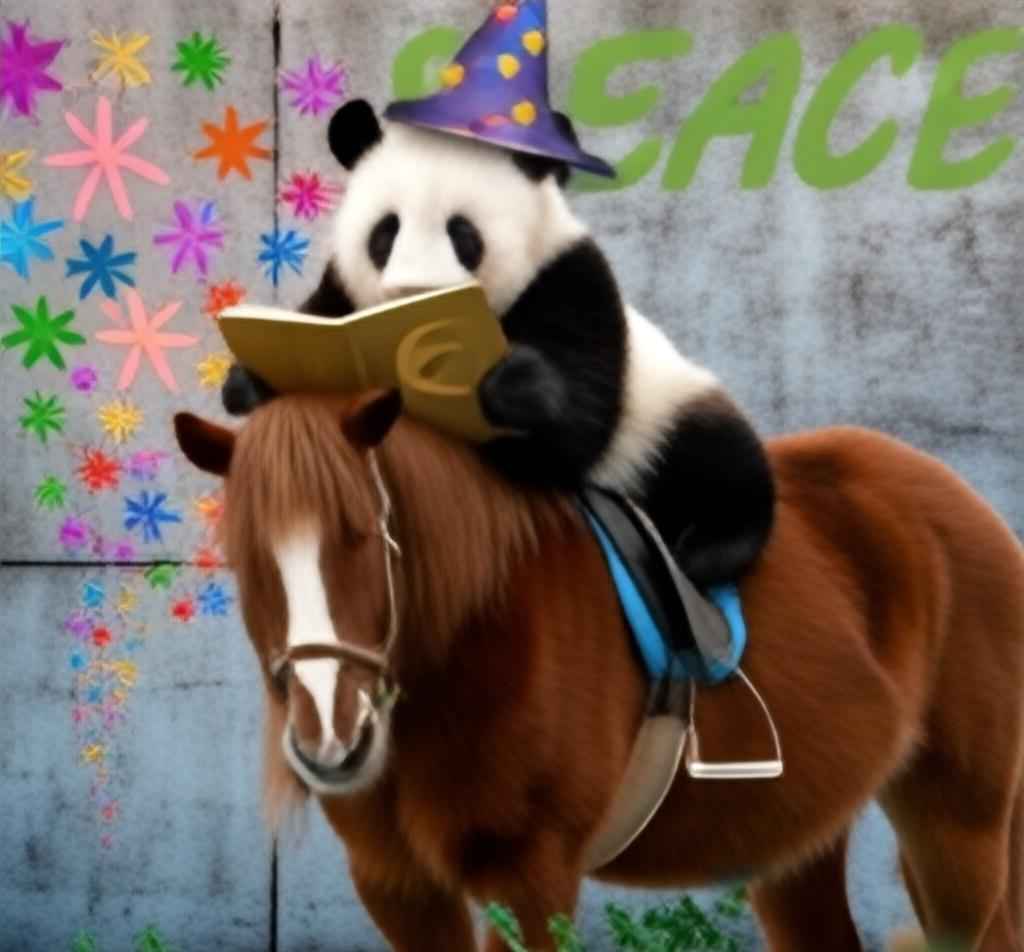}}&
\raisebox{-0.45\height}{\includegraphics[width=\teaserwid]{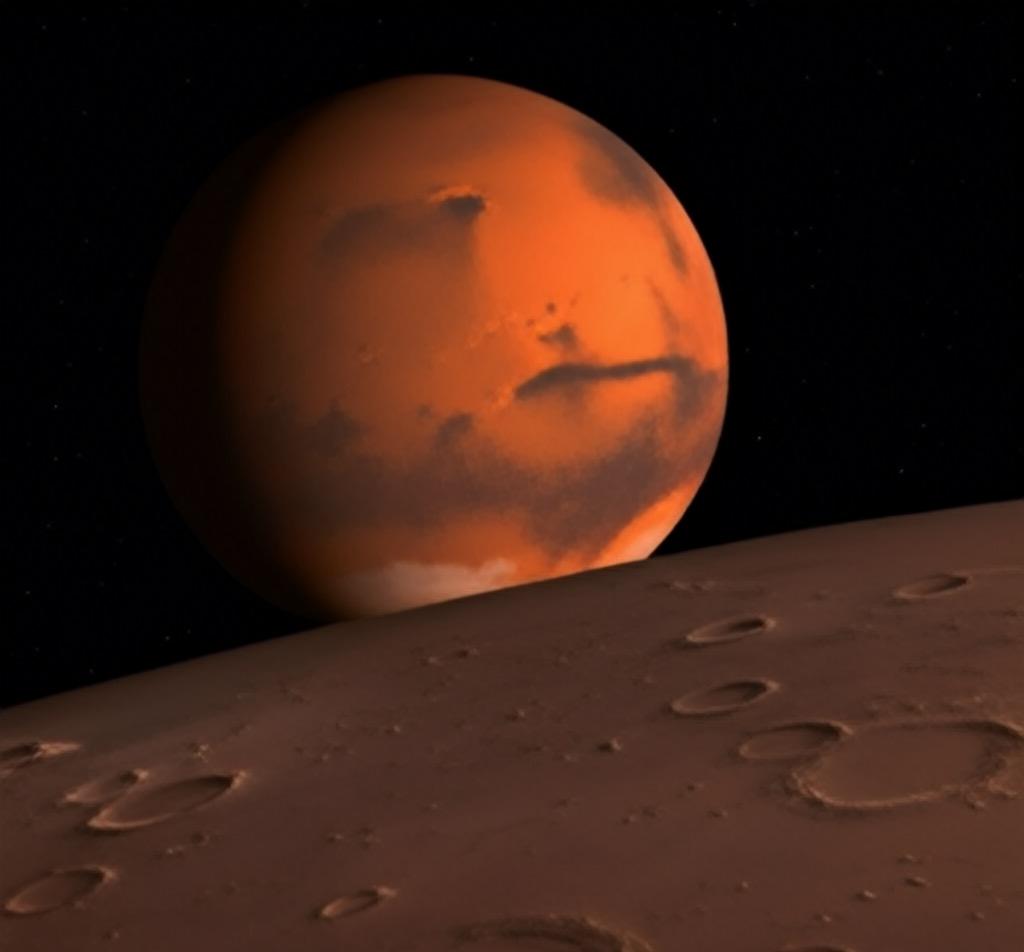}}&
\raisebox{-0.45\height}{\includegraphics[width=\teaserwid,height=\teaserwid]{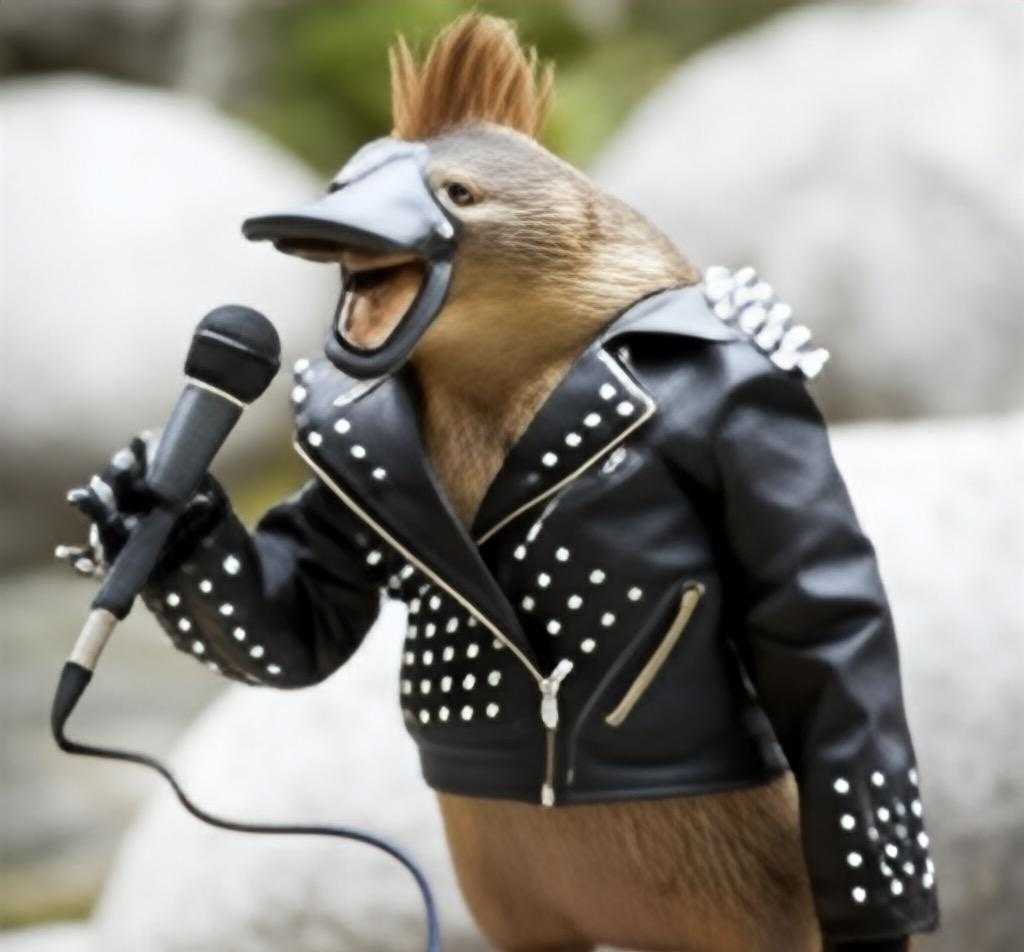}}&
\raisebox{-0.45\height}{\includegraphics[width=\teaserwid,height=\teaserwid]{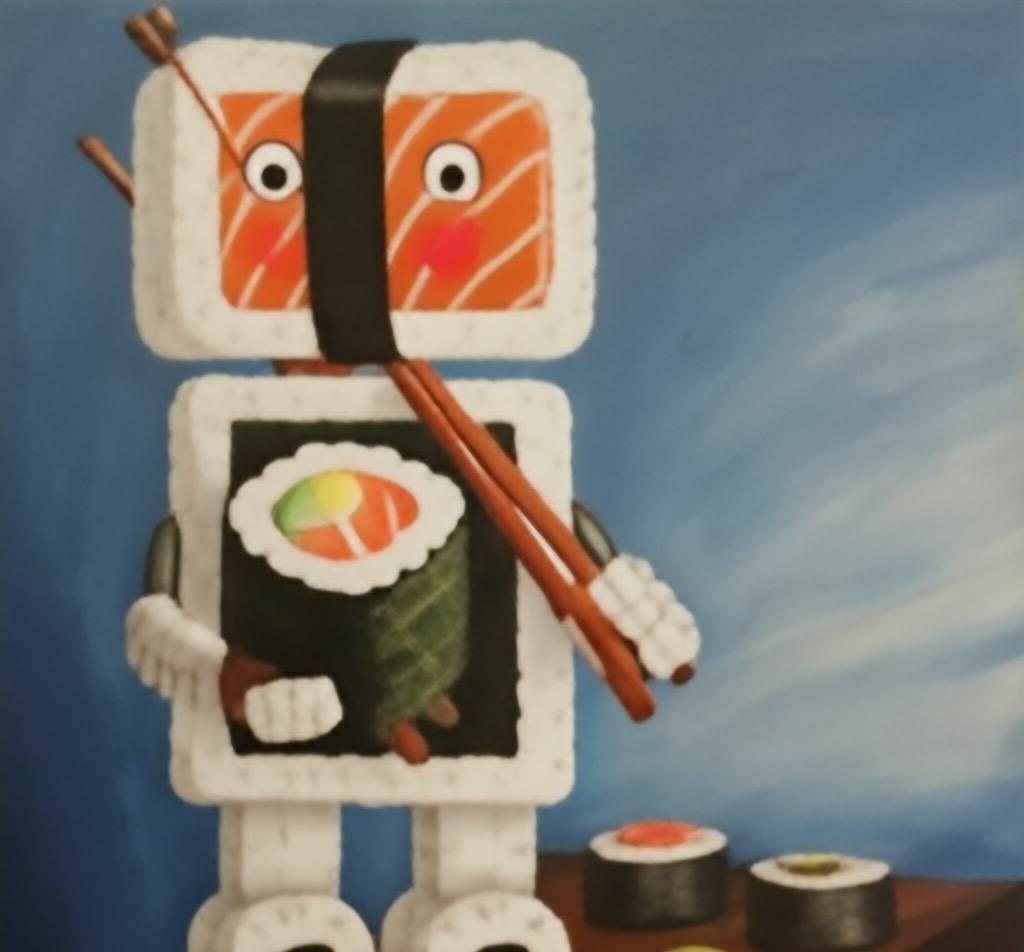}}&
\raisebox{-0.45\height}{\includegraphics[width=\teaserwid,height=\teaserwid]{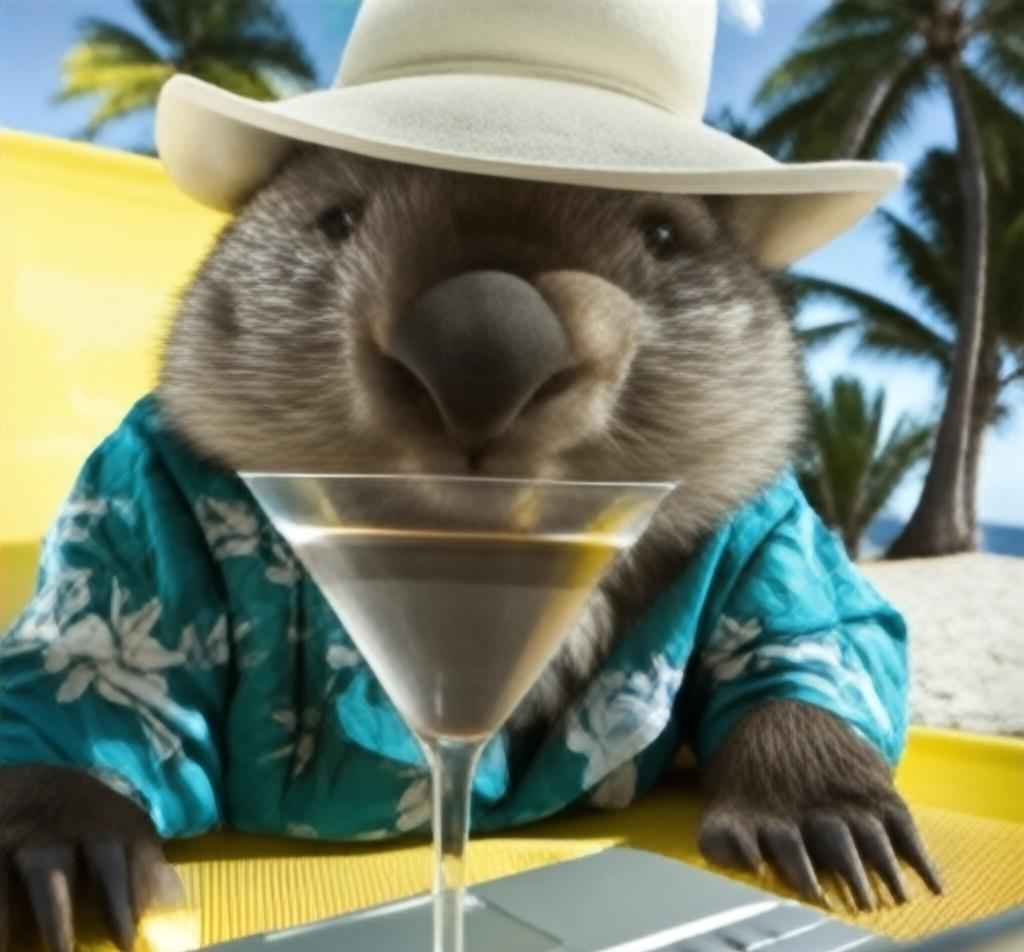}}&
\raisebox{-0.45\height}{\includegraphics[width=\teaserwid,height=\teaserwid]{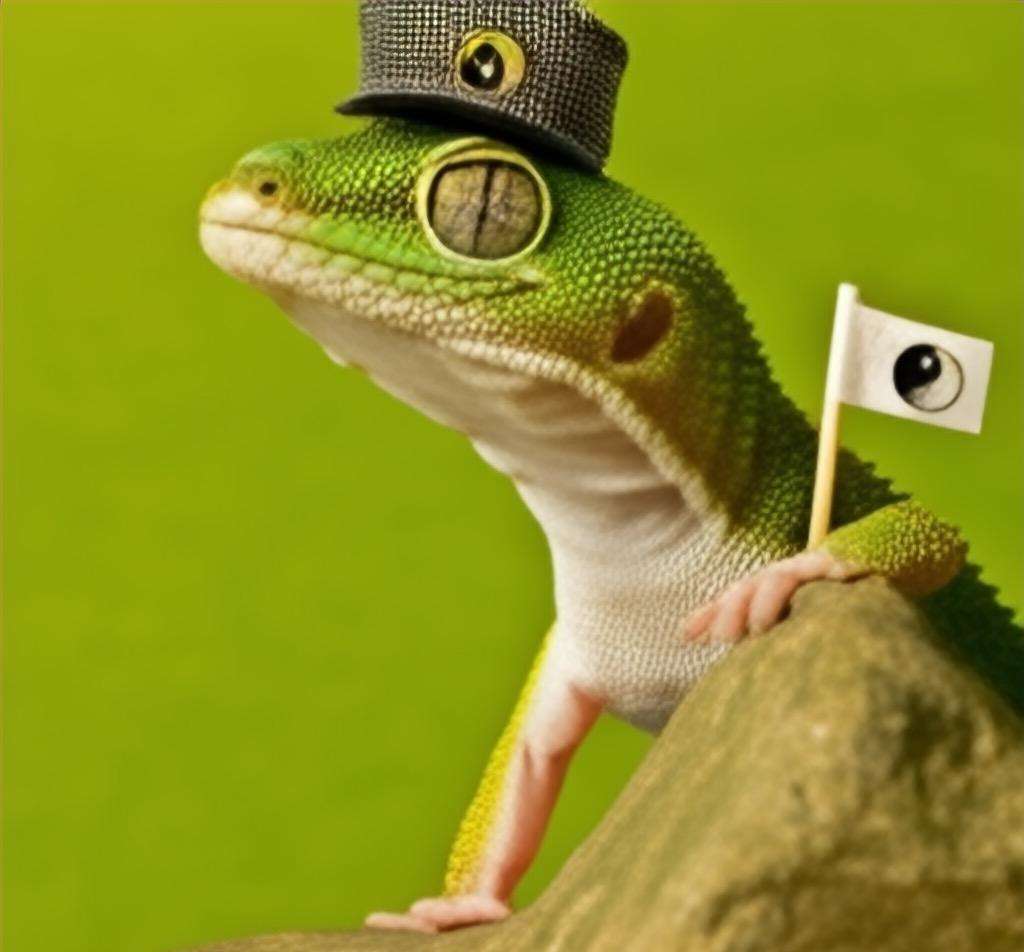}} &
\raisebox{-0.45\height}{\includegraphics[width=\teaserwid,height=\teaserwid]{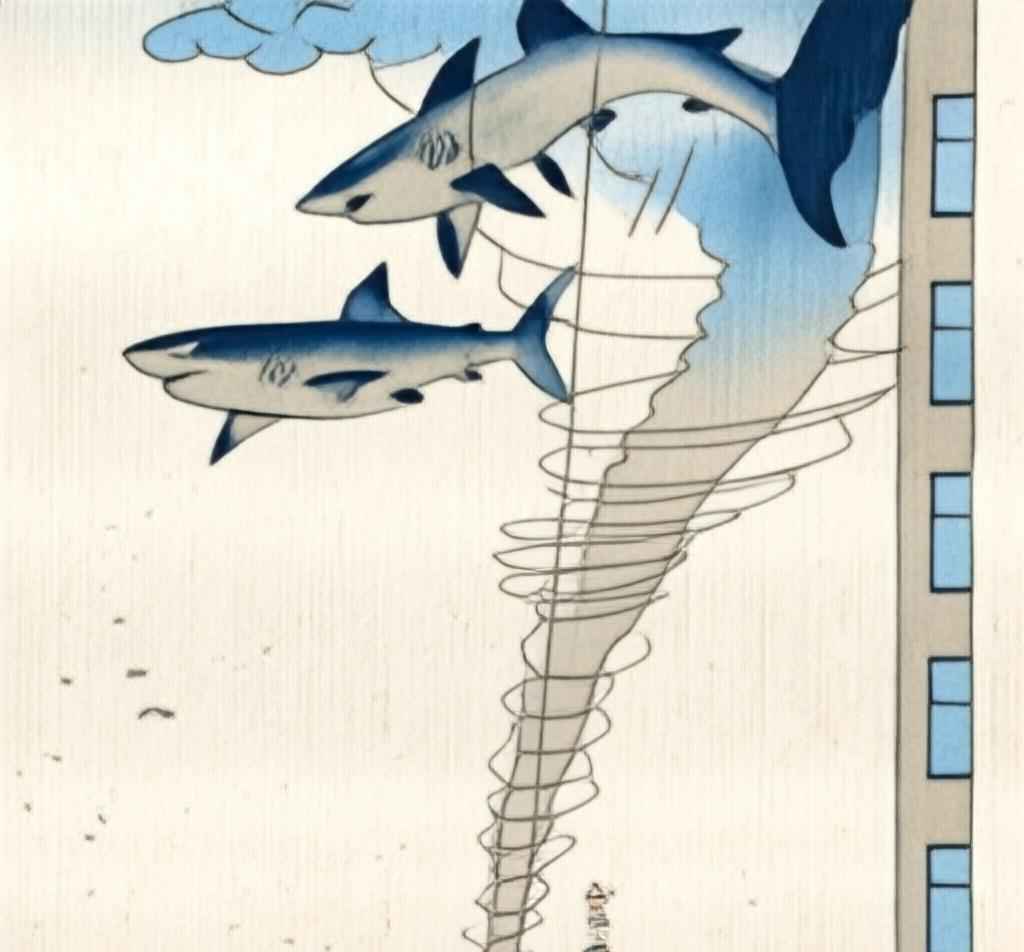}}
\vspace{.05in}
\\
\end{tabular}}
\vspace{-.1in}
\captionof{figure}{Examples of the established deepart detection database (DDDB). The examples of LAION-5B \cite{schuhmann2022laion} are conventional artworks (conarts), and the rest examples (i.e., StableDiff \cite{rombach2021highresolution},DALL-E 2 \cite{ramesh2022hierarchical},Imagen \cite{saharia2022photorealistic},Midjourney \cite{david2022mj}, and Parti \cite{yu2022scaling}) are deepfake artworks (deeparts) produced by generative models. Our data and code will be released.}
\label{fig:dddb}

\end{center}
% }]
{
  \renewcommand{\thefootnote}%
    {\fnsymbol{footnote}}
  \footnotetext[1]{Corresponding Author}
  \footnotetext[2]{Corresponding Author}
}

\begin{abstract}
Deepfake technologies have been blurring the boundaries between the real and unreal, likely resulting in malicious events. By leveraging newly emerged deepfake technologies, deepfake researchers have been making a great upending to create deepfake artworks (deeparts), which are further closing the gap between reality and fantasy. To address potentially appeared ethics questions, this paper establishes a deepart detection database (DDDB) that consists of a set of high-quality conventional art images (conarts) and five sets of deepart images generated by five state-of-the-art deepfake models. This database enables us to explore once-for-all deepart detection and continual deepart detection. For the two new problems, we suggest four benchmark evaluations and four families of solutions on the constructed DDDB.
The comprehensive study demonstrates the effectiveness of the proposed solutions on the established benchmark dataset, which is capable of paving a way to more interesting directions of deepart detection. The constructed benchmark dataset and the source code will be made publicly available.

\end{abstract}

\section{Introduction}
\label{sec:intro}

There has been a propensity to view deepfake technologies as destructive to the supposed boundaries between the real and unreal, leading to potentially detrimental effects. Despite this, deepfake researchers are continuing to make breakthroughs by wielding newly emerged deepfake technologies to create artworks, which are called \emph{deeparts} throughout this paper. The new deepart techniques include Stable Diffusion\cite{rombach2021highresolution}, DALL-E \cite{ramesh2021zero,ramesh2022hierarchical}, Imagen \cite{saharia2022photorealistic},  Midjourney \cite{david2022mj}, and Parti \cite{yu2022scaling}\footnote{There are also emerging some video-based deeparts like Make-A-Video \cite{singer2022make} and Imagen Video \cite{ho2022imagen} by applying deepfake technologies with video processing techniques.}, to name a few. As shown in Figure \ref{fig:dddb}, compared to conventional deepfakes, deeparts have been making the boundary between reality and fantasy much more blurry.

The ubiquitous emergence of deeparts results in a debate on whether deeparts are really artistic. Some believe they are not real arts because they are created by machines, while others argue that deeparts are indeed real arts since they generally request a human element to create them. Nevertheless, neither of these two sides deny the necessity of detecting deeparts and identifying their origins, which could be regarded as their copyrights like those of traditional artworks. As deeparts produce a tiny gap between reality and fantasy, distinguishing them from conventional arts becomes much more challenging than detecting traditional deepfakes from authentic images. Moreover, their high diversity makes identifying their origins very much non-trivial. Hence, we concentrate on exploring the problem of deepart detection with its corresponding copyright identification problem.

In this paper, we build a deepart detection database (DDDB). As demonstrated in Figure \ref{fig:dddb}, this database consists of a set of high-quality conventional art (\emph{conart}) images from LAION-5B \cite{schuhmann2022laion} and five sets of deepfake art (\emph{deepart}) images that are generated by five state-of-the-art deepart models, i.e., Stable Diffusion (StableDiff)
\cite{rombach2021highresolution}, DALL-E 2 \cite{ramesh2022hierarchical}, Imagen \cite{saharia2022photorealistic}, Midjourney \cite{david2022mj}, and Parti \cite{yu2022scaling}, respectively. It is known that all deeparts cannot be collected at once. Instead, new deeparts always appear time by time. Moreover, real-world settings often require privacy protection and satisfy storage constraints. To meet such real-world constraints based on DDDB, we suggest exploring two desirable deepart detection problems. One is \emph{once-for-all deepart detection (ODD)}. It learns from either one single deepfake dataset or any presently available datasets once for the generalization to all seen and unseen deepfakes \cite{wang2020cnn}. As ODD has no access to the newly appeared deepart data for training, it is intractable for ODD to identify the copyrights of the lately appeared deeparts. A more favorable setting is \emph{continual deepart detection (CDD)}. 
It enables both deepart detection and copyright identification by learning from a data stream of sequentially appeared deepart data.

Following the general setup of continual deepfake detection \cite{marra2019incremental,kim2021cored,li2022continual}, each learning session of the suggested CDD should use one model's generated deeparts as positive samples, and its used training data (conarts) as negative samples for learning. However, except for StableDiff, the majority of current state-of-the-art deepart models (e.g., DALL-E and Imagen) veil their model weights and their used training data (conarts) for the protection of intellectual properties. To fill the gap for these models, additionally acquiring other pure conarts is non-trivial and expensive, since the Internet is now full of deeparts. An alternative solution is to reuse the used conarts of StableDiff, 
but explicitly accessing the previously/lately learned data is not allowed for continual learning \cite{shokri2015privacy, wang2022dualprompt}. Therefore, we suggest keeping the natural setup where most of deepart models have no officially released conarts.
In this case, on DDDB, besides the session on StableDiff that learns two classes, the rest ones learn on one single class (deepart) if the rehearsal on early learned conarts is prohibited. In other words, it is essentially a mixed two-class and one-class continual learning problem. This is one of the main challenges for the suggested CDD, and it highly differs from traditional continual learning problems. 

For the two suggested deepart detection problems (ODD and CDD), we design four benchmark evaluations on DDDB: one is for the ODD problem simulating the once-for-all learning scenes, and the other three are for the CDD problem handling three continual learning scenarios from easy to hard. To address the four benchmarks, we suggest four families of solutions respectively. 
To address the ODD benchmark, we revisit the state-of-the-art networks like vision transformer (ViT) \cite{dosovitskiy2020image} with either fine-tuning or prompt-tuning strategies. 
% The trained ViT models show promising generalization capability on the other four deepart datasets. 
For the first CDD benchmark that allows for the rehearsal on early encountered deeparts and LAION-5B conarts, we propose a sharing scheme to enable the failed general rehearsal-based methods like Foster \cite{wang2022foster} to work again. For the second CDD benchmark that merely allows for the rehearsal on LAION conarts, we suggest a similar sharing method to fix the malfunctioned state-of-the-art exemplar-free methods like S-Prompts \cite{wang2022sprompt}. 
Particularly, for the third CDD benchmark that does not allow any rehearsal, we discover most of the exemplar-free methods collapse due to the complete missing of negative class (conarts) in most sessions of continual learning. To save them, we exploit a unified transformation framework to address this particular continual learning problem. \emph{Through extensive empirical studies, the CDD methods generally show clear superiority over the ODD ones, and thus this paper favors benchmarking deepart detection with continual learning.}    

In conclusion, our main contributions are four-fold:
\begin{itemize}
\item We establish the first deepart detection database (DDDB)\footnote{A concurrent work \cite{wang2022diffusiondb} collects a dataset of Stable Diffusion generated deepfake images with a discussion on its potential use for deepfake detection on Stable Diffusion.}. 
\item We probe two novel problems, i.e., once-for-all deepart detection (ODD) and continual deepart detection (CDD). 
Notably, CDD studies the mixed two-class and one-class continual learning for the first time up to our knowledge.
\item For these two new problems (ODD and CDD), we suggest four benchmark evaluations.

\item We propose four groups of solutions for the four benchmarks on DDDB. In particular, we propose a general transformation framework to rescue those failed/collapsed methods for the most challenging CDD benchmark.
\end{itemize}

\section{Related Work}
\label{sec:related}

\noindent\textbf{Deepfake Detection Datasets and Methods.} 
To prevent the malicious use of deepfakes, numerous datasets (e.g., \cite{korshunov2018deepfakes,li2018ictu,li2019celeb,dufour2019contributing,rossler2019faceforensics++,li2022continual,wang2020cnn}) have been suggested for deepfake detection. For example, \cite{rossler2019faceforensics++} manipulates real faces from YouTube using deepfake techniques. \cite{wang2020cnn} collects a database of deepfakes generated by GAN-based methods and other deepfake techniques, which are trained on real non-art images/videos.

Many deepfake detection methods \cite{wang2020cnn,yang2022deepfake} have been proposed on these datasets.
Particularly, \cite{wang2020cnn} suggests an once-for-all learning approach for the deepfake detection. It uses pre-trained ResNet-50 \cite{he2016deep} on ImageNet, and further trains it on one single ProGAN deepfake dataset. 
With careful pre- and post-processing and data augmentation, \cite{wang2020cnn} studies that the once-for-all learning based model can generalize well to unseen deepfakes. 
Inspired by \cite{wang2020cnn}, to address the challenging once-for-all deepart detection, we further study both fine-tuning and prompt-tuning strategies for a set of state-of-the-art networks like vision transformer (ViT) \cite{dosovitskiy2020image}.

In addition, \cite{marra2019incremental,kim2021cored,li2022continual} suggest the setting of continual deepfake detection.
Notably, \cite{marra2019incremental} studies this setting on a collection of GAN-based deepfake datasets, each of which has a set of real images. Due to the full availability of both reals (negative samples) and fakes (positive samples) for each learning session, \cite{marra2019incremental} adapts one of the traditional continual learning methods
through a multi-task learning scheme over both deepfake recognition and detection tasks. 
Different from these works,
our proposed CDD problem suffers from a severe missing of negative samples (conarts), which challenges the conventional continual learning a lot according to our study in this paper. Moreover, the much more blurry boundary between positive samples (deeparts) and negative samples (conarts) aggravates the challenge.

\noindent\textbf{General Continual Learning Methods.}
Continual learning (lifelong learning or incremental learning) aims at learning from a data stream with previously learnt data being not accessible.
The main challenge is catastrophic forgetting~\cite{mai2022online}. There are two main groups of methods: (1) rehearsal-based ones (e.g., \cite{rebuffi2017icarl,lopez2017gradient,wu2019large,zhou2021co,wang2022foster,marra2019incremental}), and (2) rehearsal-free ones (e.g., \cite{li2017learning,kirkpatrick2017overcoming,wang2021learning,wang2022sprompt,wang2022dualprompt,wang2022isolation}). 
The rehearsal-based methods generally revisit one set of old samples from the previous tasks for less forgetting on old knowledge. 
In contrast, rehearsal-free methods address forgetting commonly by distilling the responses of the old model to constrain the new model with currently available samples.
For our suggested CDD problems, all these methods need a careful adaptation. In particular, we discover that most of the existing rehearsal-free methods\footnote{One-class continual learning methods like \cite{asadi2021tackling, hu2021continual, hu2020hrn} are also not fully suitable for our CDD which learns two classes from the base session.} fail/collapse in the most challenging CDD setting where no buffer is allowed to save any historical data. To rescue these methods, we propose a unified scheme in this paper.

\section{Benchmark Database}
\label{sec:data}

To promote the research on deepart detection, we establish a database of deepart and conart images from five state-of-the-art deepfake generative models and one dataset of conventional artworks respectively.

\subsection{Database Construction}

We collect high-quality art images generated by five current state-of-the-art generative models: Stable Diffusion
\cite{rombach2021highresolution}, DALL-E2 \cite{ramesh2021zero,ramesh2022hierarchical}, Imagen \cite{saharia2022photorealistic}, Midjourney \cite{david2022mj}, and Parti \cite{yu2022scaling}.
All chosen models are for text-to-image synthesis, which allows people to use language (prompts) to create artworks freely.
The generated deepfake art images (\emph{deeparts}) are regarded as positive samples.
As only StableDiff unveils its training data (LAION-5B \cite{schuhmann2022laion}), we merely use LAION-5B to form the set of conventional art images (\emph{conarts}). It provides artworks created by humans, which are either digital artworks or traditional paintings. These conarts are used as negative samples. Examples of these art images are presented in Figure~\ref{fig:dddb}.

\begin{table}[]
\vspace{-0.3cm}
\resizebox{0.95\linewidth}{!}{
\begin{tabular}{c|cc|cc}
\hline
\multirow{2}{*}{Models} & \multicolumn{2}{c|}{Train} & \multicolumn{2}{c}{Test} \\ \cline{2-5} 
 & Conart  & \multicolumn{1}{c|}{Deepart} & Conart & Deepart \\ \hline
Stable Diffusion \cite{rombach2021highresolution} & 62,154 & 65,556 & 1,030 & 1,086 \\
DALL-E2 \cite{ramesh2022hierarchical} & - & 775 & 186 & 194 \\
Imagen \cite{saharia2022photorealistic} & - & 178 & 44 & 46 \\
Midjourney \cite{david2022mj} & - & 4,306 & 1,027 & 1,077 \\
Parti \cite{yu2022scaling} & - & 155 & 38 & 40 \\
 \hline
\end{tabular}
}
\caption{Statistics (image numbers) of the established deepart detection database (DDDB) that consists of five deepfake models generated deeparts as well as a set of conarts. Note that all the conarts are from LAION-5B \cite{schuhmann2022laion}. `-' indicates no conarts are released officially by the corresponding deepart model.}
\label{tab:dddb}
\vspace{-0.3cm}
\end{table}

The principles of our data collection are three-fold: public availability, high-quality artworks, and reliable sources. All deeparts are either created by ourselves using available generators or from social media (Instagram, Twitter, etc.). 
We carefully make sure all deeparts are created by corresponding generators and 
they have high artistic values.
All conarts are collected from LAION-5B, which contains 5 billion image-text-pairs crawled from web pages between 2014 and 2021. The reasons for selecting LAION-5B are two-fold. Firstly, the related works for deeparts have been emerging since 2021, so it is unlikely to collect high-quality deeparts on the Internet at that time. Even if there are few deeparts, manually identifying them is not challenging. Secondly, this dataset is used for Stable Diffusion, and thus we can use it as negative samples to construct a training set for DDDB.

Table~\ref{tab:dddb} lists the statistics of each training/test set in the DDDB. Except for Stable Diffusion, as of now all the rest deepart models do not release their used training data of conarts. All the training/test conarts are selected from LAION-5B with high artistic scores.
For deepart detection over Stable Diffusion, the training deeparts are either generated by us using the officially released Stable Diffusion model or crawled from Lexica\footnote{\scriptsize{\url{https://lexica.art/}}} that collects high-quality arworks generated by Stable Diffusion.    
In order to generate high-quality deepart images using Stable Diffusion, we need to get appropriate prompts. We first randomly select around 100,000 conarts from LAION-5B with high artistic scores. Then we use CLIP Interrogator\footnote{\scriptsize{\url{https://github.com/pharmapsychotic/clip-interrogator}}} to generate descriptions matching the given images. Finally, we use these descriptions as prompts with Stable Diffusion to generate deep artworks. Finally, deeparts for training set are selected from these generated images with high quality.
The test set of deeparts are all from Lexica. 
For deepart detection on the other four models, as their pre-trained models are all not publicly available, we form their training/test sets of deeparts by collecting their generated deeparts from their official websites as well as their social media sources.
As some deeparts have watermarks or logos from the Internet, we remove them by cropping these images to avoid trivial tricks for detection. 
For test sets, we select deeparts the same way and manually select high quality conarts from LAION-5B.

\subsection{Probed Problems}
\label{sec:prob_def}

To simulate the real-world settings, based on DDDB, we suggest probing the \emph{once-for-all deepart detection (ODD)} and \emph{continual deepart detection (CDD)} problems. ODD learns from one single deepart dataset once using one detection model, that can be applied for the generalization to all the deepart dataset. In contrast, CDD learns from a data stream of all the sequentially available deepart datasets, with a unified detection model being trained and applied to detect all the deeparts. Unlike the other four deepart models, the pre-trained model of Stable Diffusion has been officially released. Thus we are able to collect a big dataset of its generated deeparts. This enables us to use the StableDiff dataset as a base for both the ODD and CDD problems. For CDD, the remaining four datasets are learned sequentially in a shuffling order.

Formally, at the base learning session, the model accesses the incoming data $\mathbb{D} _0=\left \{ x_i^0,y_i^0 \right \} _{i=0}^{N_0}$, where $x_i^0$ is the $i$-th deepart/conart image from the base session that is indicated by $0$, $y_i^0$ is its label, and $N_0$ is the total number of samples at this session.
In this session, $y_i^0 \in \{-1,1\}$ where $-1$ indicates the negative class (conart) and $1$ represents the positive class (deepart).
Both ODD and CDD go through this base session. After this session, the ODD model stops learning while the CDD model moves on to learn the rest four deepart datasets  $\mathbb{D} _s=\left \{ x_i^s,y_i^s \right \} _{i=s}^{N_s}, s=1,...S$, sequentially.

\subsection{Benchmark Evaluations}

For the two probed problems, we design four benchmark scenarios for them as follows.
\begin{itemize}
\item \emph{Benchmark1 (ODD)} is designed for once-for-all deepart detection. It merely allows training on the dataset of Stable Diffusion once, followed by an evaluation on all the test sets in Table \ref{tab:dddb}.
\item \emph{Benchmark2 (CDD1)} aims for continual deepart detection allowing the rehearsal on early appeared deeparts and LAION-5B conarts.
\item \emph{Benchmark3 (CDD2)} is tailored for continual deepart detection that merely allows for the rehearsal on LAION-5B conarts.
\item \emph{Benchmark4 (CDD3)} targets for continual deepart detection that prohibit any rehearsal.
\end{itemize}
\vspace{0.1cm}

For all the ODD/CDD benchmarks, we follow \cite{wang2020cnn,li2022continual} to suggest evaluating models' performance on each dataset using both detection accuracy and average precision. In detail, both the scores are first computed for each of the five datasets separately, and they are then averaged with an equal weight, because every task is assumed to be equally important. Accordingly, each ODD/CDD model should report its average detection accuracy (AA) and mean average precision (mAP). Besides, for all the CDD benchmarks, similar to traditional continual learning evaluations, we also suggest computing the average forgetting (AF). 

In addition, Copyright Infringement Protection for artworks is also highly valuable \cite{cetinic2022understanding}.
We should not only detect whether the given picture is a deepart, but also identify what deepart model generates this picture.
Thus, for all the CDDs, we suggest further reporting copyright identification accuracy (CA), which calculates the accuracy that the picture is classified into the correct deepfake model, with all conarts being classified as one single class due to the non-trivial annotations on their real origins.

The calculation details of AA, mAP, AF, and CA are presented in Appendix.

\section{Suggested Methods}

In this section, we suggest four families of solutions for the designed four benchmarks, respectively. In particular, we rescue state-of-the-art rehearsal-free methods that collapse or fail in CDD3.

\subsection{Revisiting State-of-the-art Methods for ODD}

The key idea of the suggested ODD approaches is to leverage the generalization capability of big deep learning models that were trained on a big dataset. A natural approach is to revisit state-of-the-art image classification models like ResNet \cite{he2016deep} (recommended by \cite{wang2020cnn}) and ViT \cite{dosovitskiy2020image} (a larger model) that are trained on ImageNet.
Another principle is to adapt the pre-trained model efficiently to a new big dataset for the downstream task, meanwhile maintaining its generalization capability to other datasets for the same downstream task. We study both our collected StableDiff deepart dataset and the ProGAN deepfake dataset suggested by \cite{wang2020cnn}. Based on these two datasets, besides the use of fine-tuning (FT) suggested by \cite{wang2020cnn}, we further recommend the use of the prompt-tuning (PT) technique \cite{jia2022vpt,wang2022dualprompt, wang2022sprompt} on the pre-trained model for a more efficient adaption to our deepart detection task.
The PT technique transfers large pre-trained transformer to downstream tasks using only a tiny amount of trainable parameters as additional input tokens while keeping other parameters frozen. It has been proved ~\cite{wang2022sprompt,wang2022dualprompt,wang2021learning} that PT is efficient for adapting to downstream tasks.
In addition, we also revisit the data augmentation technique for the potential improvement of the generalization according to the study in \cite{wang2020cnn}. The technique applies either Gaussian blur or JPEG for the augmentation. It suggests the augmentation of Blur+JPEG ($a\%$), which means the image is possibly blurred and JPEG-ed with the probability of $a\%$.   

\subsection{Adapting Previous Methods for CDD1/CDD2}

Most traditional continual learning methods are designed for class-incremental learning and their classifiers are for multiple classes.
Hence, for CDD, we suggest keeping their duplicated multi-class classifiers for copyright identification, as done in \cite{li2022continual}.
Given a sample $x$, we apply $\hat y = \mathrm{argmax}( \varphi (x))$, where $\varphi (x)$ is the logits produced by the model. If $\hat y$ is one of the conarts/deeparts classes, we will predict $x$ to conarts/deeparts. This design also enable us to predict which deeparts it is.

As CDD1 allows for the rehearsal on both early appeared negative (LAION-5B conarts) and positive samples (deeparts), the general rehearsal-based continual learning methods can be applied to this benchmark scenario. 
The problem for CDD1 is that all the non-base sessions merely has positive class (deeparts).
However, most continual learning classifiers are trained with a cross-entropy loss that requires at least two classes' data for training ~\cite{hu2021continual,hu2020hrn}.
To address this problem, 
we suggest \emph{sharing} the stored negative samples for replay from the base session among all incremental sessions. In this case, a model can be continually learned with the co-occurrence of positive and negative samples.

For the CDD2 benchmark, the rehearsal is only performed on the negative samples from LAION-5B in the base learning session. With no rehearsal on the old positive samples, all the rehearsal-based methods fail to work in the context of CDD2. In this case, we suggest using general rehearsal-free methods. However, their commonly used cross-entropy loss collapses when moving to the non-base sessions where only positive samples are available and the right corresponding negative samples are missing. To handle this issue, like CDD1, we suggest adapting them to share the LAION-5B conarts as the negative samples for all the non-base sessions.

\begin{figure*}[!] 
\centering 
\includegraphics[width=0.98\textwidth]{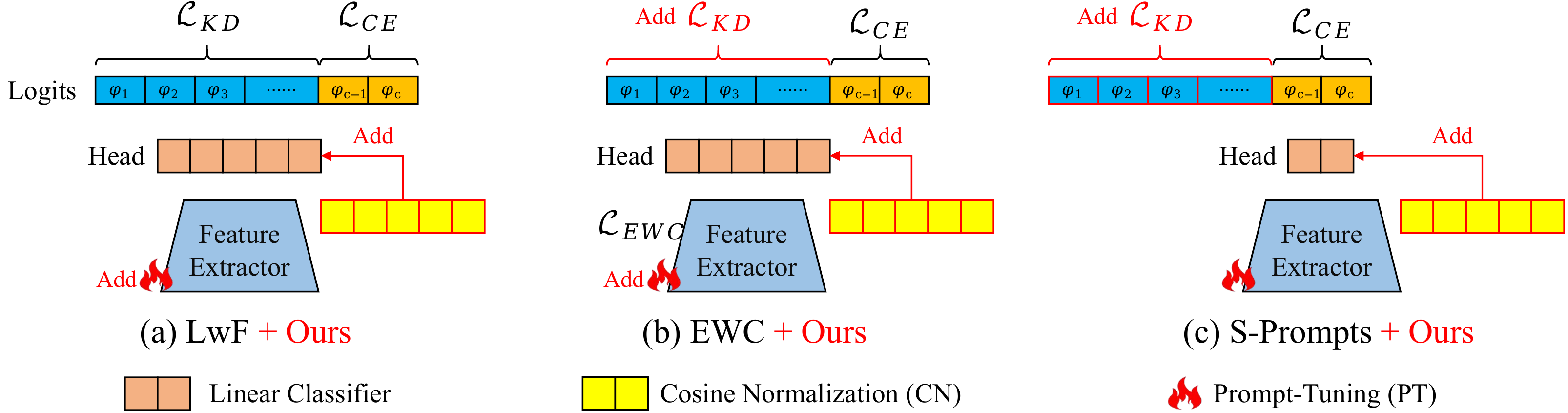} 
\vspace{-0.1cm}
\caption{Illustration of the proposed framework to rescue the rehearsal-free methods (a) LwF \cite{li2017learning}, (b) EWC \cite{kirkpatrick2017overcoming} and (c) S-Prompts \cite{wang2022sprompt} using the three suggested techniques: (1) Knowledge Distillation (KD), (2) Cosine Normalization (CN), and (3) Prompt-Tuning (PT).}
\label{fig:method}
\vspace{-0.4cm}
\end{figure*}

\subsection{Rescuing Collapsed/Failed Methods for CDD3}
\label{subsec:rescue}

The CDD3 benchmark scenario prohibits any rehearsals on samples from the earlier learning sessions. This case clearly disables all the rehearsal-based continual learning methods. The missing of negative samples for the non-based sessions makes most of the rehearsal-free methods fail and even collapse. This is mainly due to the common designs of the general rehearsal-free methods to avoid the use of a rehearsal buffer. Let's take two methods for an analysis: one is LwF \cite{li2017learning} which is one of the most popular methods, and the other is S-Prompts \cite{wang2022sprompt} that is the current state-of-the-art for rehearsal-free continual deepfake detection. For each new session, these two methods both incrementally adds a new classifier head that is optimized by its own cross-entropy loss, since it is not trivial to apply an unified cross-entropy loss to all the classifier heads for all the classes without being allowed for a rehearsal on classes from the old learning sessions. The common design leads to three main issues in the context of CDD3. To address these three issues, 
we suggest a general framework including three essential components to rescue the rehearsal-free methods for CDD3.

\emph{Issue1 (loss malfunction).} The most commonly used cross-entropy loss always collapses when meeting one single class in the non-base sessions of CDD3, as it is known that cross-entropy requests at least two classes for training \cite{hu2020hrn,hu2021continual}. One of the promising solutions is to use one-class learning losses like negative log-likelihood based one-class suggested by \cite{hu2021continual}. However, the one-class learning losses are not desired for the base session of CDD3, which has two classes. Hence, we recommend the other solution that enforces a \emph{knowledge distillation (KD)} based regularization used in \cite{li2017learning} to the cross-entropy loss.

\emph{Issue2 (head misalignment).} The common design results in highly different magnitudes of the separate classifier heads' outputs, since each of which is optimized by one specific loss function. A similar phenomena was studied in \cite{hou2019learning, zhou2021co}, which is rehearsal-based methods with all the classifier heads being unified together. It discovers that magnitudes of both the weights and the biases of the linear classifier for the new session's classes are significantly higher than those for the classes for the old sessions. This incurs a high bias in the predictions that favor the new classes. Intuitively, CDD3 highly aggravates this head imbalance issue, due to the independent optimization on these heads as well as the one-class effect in the non-base sessions. To address this issue, we suggest applying the \emph{Cosine Normalization (CN)} technique \cite{hou2019learning} for a normalization of the separate classifier heads, to save the general rehearsal-free methods. 

\emph{Issue3 (feature bias).} Compared to regular continual learning scenarios, the one-class based optimization has a clearly higher risk to guide the networks to overfit the new data in the non-based sessions when we use the conventional fine-tuning techniques. This might lead to a remarkable feature bias that favors the current session's data \cite{hu2021continual}.
It is non-trivial and inefficient to determine which layers should be fine-tuned and which ones should be frozen. 
Besides, it is also challenging and not efficient to enforce a weight constraint to regularize the weight update.
Therefore, we suggest the use of the \emph{prompt-tuning} technique \cite{jia2022vpt,wang2022sprompt}, as we empirically find it is capable of addressing the feature bias issue 
% (see Figure \ref{fig:feture_bias}) 
while it has shown a superior efficiency for transfer learning \cite{wang2022dualprompt,wang2022sprompt}.

According to this analysis, we propose a general model transformation framework to rescue the previous rehearsal-free methods and adapt them to CDD3. 
The key steps of the framework include: Firstly, using \emph{prompt-tuning} instead of \emph{fully fine-tuning} on the state-of-the-art feature extractor like pre-trained ViT. Secondly, using \emph{cosine normalization} to the linear classifier. Thirdly, performing \emph{knowledge distillation} on the model's output by appending a KD loss on the overall loss function. In Figure~\ref{fig:method}, we select two classical rehearsal-free methods (LwF~\cite{li2017learning}, EWC~\cite{kirkpatrick2017overcoming}) and one state-of-the-art method (S-Prompts~\cite{wang2022sprompt}) as examples to 
illustrate the proposed framework. Appendix presents the computation details of the three steps.

\section{Experiments}
\label{sec:method}

\begin{table*}[]
\vspace{-0.3cm}
\centering
\resizebox{1\linewidth}{!}{
\begin{tabular}{cllcccccccc}
\hline
\multicolumn{1}{l}{} & \multicolumn{1}{l}{} & & \multicolumn{5}{c}{Once-for-all Deepart Detection (ODD)} & \multicolumn{1}{c}{} \\ 
\cline{4-8}
\multicolumn{1}{l}{\multirow{-2}{*}{Training Dataset}} & \multicolumn{1}{l}{\multirow{-2}{*}{Methods}}   & \multirow{-2}{*}{Augments}  & StableDiff & DALL-E2  & Imagen & Midjourney & Parti & \multicolumn{1}{c}{\multirow{-2}{*}{AA}} & \multicolumn{1}{c}{\multirow{-2}{*}{mAP}}\\
\hline
&  & Blur+JPEG (0.1) & 48.20  & 49.47   & 48.86  & 49.14 & 50.00 & 49.14 & 48.85 \\
\multirow{-2}{*}{ProGAN} & \multirow{-2}{*}{ResNet-FT} & Blur+JPEG (0.5) & 48.20  & 49.47   & 48.86  & 48.15 & 48.68 & 48.67 & 43.66 \\ \hline
&  & Blur+JPEG (0) & 96.85 & 48.39 & 50.52 & 63.23  & 49.71 & 61.74 & 75.83 \\
&  & Blur+JPEG (0.1) & 96.73  & 48.83   & 50.17  & 64.81 & 49.32 & 61.97 & 75.32 \\
& \multirow{-3}{*}{ResNet-FT} & Blur+JPEG (0.5) & 94.97  & 48.21   & 50.41  & 65.32 & 49.54 & 61.69 & 74.51 \\
& & Blur+JPEG (0) & \textbf{98.96}  & 48.94   & 52.27  & 61.97  & 49.72 & 62.37 & 77.54 \\
& & Blur+JPEG (0.1) & 95.82  & 49.13   & 51.36  & 63.71 & \textbf{53.89} & \textbf{62.78} & \textbf{80.23} \\
& \multirow{-3}{*}{ViT-FT} & Blur+JPEG (0.5) & 89.37  & \textbf{51.85} & \textbf{53.64}  & \textbf{67.87} & 47.37 & 62.02 & \underline{79.03} \\
& & Blur+JPEG (0) & \underline{98.11} & 48.94 & 51.14 & 61.12 & 52.63 & 62.39 & 78.95 \\
& & Blur+JPEG (0.1)  & 95.87 & 46.01  & 49.93 & 59.39 & 50.32 & 60.30 & 72.98 \\
 \multirow{-9}{*}{Stable Diffusion} & \multirow{-3}{*}{ViT-PT} & Blur+JPEG (0.5) & 90.55 & \underline{50.76} & \underline{52.49} & \underline{64.92} & \underline{53.80} & \underline{62.50} & 77.73 \\
\hline 
\end{tabular}
}
\vspace{-0.3cm}
\caption{Results of the ODD benchmark evaluation. FT: fine-tuning, PT: prompt-tuning, AA: average detection accuracy, mAP: mean average precision, Blur+JPEG ($a\%$): image is blurred and JPEG-ed with the probability of $a\%$, \textbf{Bold}: \textbf{best}, \underline{Underline}: \underline{second best}}
\label{tab:bi_track}
\end{table*}

In this section, we study a number of state-of-the-art methods with our suggested adaptations for the four benchmark evaluations on our established DDDB.

\noindent\textbf{Competing Methods.}
For ODD, we  apply the state-of-the-art deepfake CNN detector (CNNDet)~\cite{wang2020cnn} as a baseline method.
CNNDet uses ResNet-50 as the backbone that is pre-trained on ImageNet. It fine-tunes ResNet-50 on ProGAN for ODD.
We also use the suggested data augmentation methods Blur+JPEG to train the CNNDet on Stable Diffusion.
In addition, 
we replace ResNet-50 with ViT-B/16 (pre-trained on ImageNet) as an enhanced deepart detector with our suggested fine-tuning and prompt-tuning strategies. For CDD, we evaluate  state-of-the-art continual learning methods, \emph{all of which are adapted by our suggested techniques}.
The rehearsal-based methods are iCaRL~\cite{rebuffi2017icarl}, BiC~\cite{wu2019large}, GEM~\cite{lopez2017gradient}, Coil~\cite{zhou2021co} and Foster~\cite{wang2022foster}.
% Besides, Foster also needs to save an additional network.
The rehearsal-free methods are LwF~\cite{li2017learning}, EWC~\cite{kirkpatrick2017overcoming} and S-Prompts~\cite{wang2022sprompt}.
More implementations details and studies are presented in Appendix.

\noindent
\textbf{ODD Results.} 
Table ~\ref{tab:bi_track} reports the benchmarking results of ODD.
We investigate the generalization ability of the detector trained on StableDiff.
The highlights and the conclusions are listed as follows:
(1) Deeparts is obviously different from deepfakes. The original deefake detector trained on ProGAN dataset completely fail in the deepart detection task, whose binary classification accuracy is below $50\%$. The learned feature and model from ProGAN is completely unsuitable for deepart detection.
(2) The suggested data augmentation benefit deepart detection in terms of mAP.
The Gaussian blur and JPEG augmentation can enhance the generalization capability of CNNDet on the deepfake detection task \cite{wang2020cnn}.
Similarly, for deepart detection, using such data augmentation brings marginal improvement in terms of AA, while rising the mAP score by about 3\% over the no augmentation case.
(3) The performances of StableDiff and Midjourney are consistent, which indicts that some parts learnt from Stable Diffusion can be also used to detect Midjourney deeparts.
But Midjourney is a commercial software, and it is challenging to find out any connections between Midjourney and Stable Diffusion.
(4) The performance of ViT is generally better than ResNet-50, showing the superiority of the used transformer. (5) The prompt-tuning (PT) case is competitive with the fine-tuning (FT) case in terms of AA, while its mAP score is clearly worse than FT's.  

\begin{table*}[]
\centering
\resizebox{0.98\linewidth}{!}{
\begin{tabular}{cclccccccccc}
\hline
& & \multicolumn{1}{c}{} & \multicolumn{5}{c}{Continual Deepart Detection (CDD)} & & \multicolumn{1}{c}{} \\ \cline{4-8}
\multirow{-2}{*}{Benchmarks}         & \multirow{-2}{*}{Buffer Size} & \multicolumn{1}{l}{\multirow{-2}{*}{Methods}} & \multicolumn{1}{c}{StableDiff}    & \multicolumn{1}{c}{DALL-E2} & \multicolumn{1}{c}{Imagen} & \multicolumn{1}{c}{Midjourney} & \multicolumn{1}{c}{Parti} & \multirow{-2}{*}{AA} & \multicolumn{1}{c}{\multirow{-2}{*}{AF}} & \multicolumn{1}{c}{\multirow{-2}{*}{mAP}} & \multicolumn{1}{c}{\multirow{-2}{*}{CA}}\\ \hline
& & iCaRL & 60.30 & 57.94 & 77.27 & 62.12 & 57.89 & 62.38 & -8.84 & 68.26 & 48.10 \\
& & iCaRL* & 70.32 & 73.02 & 78.41 & 70.29 & 68.42 & 72.09 & -6.80 & 89.46 & 56.05 \\
& & BiC & 59.55 & 60.58 & 79.55 & 63.50 & 57.89 & 64.21 & -5.92 & 78.11 & 45.25 \\
& & BiC* & 72.54 & 78.04 & 79.55 & 72.96 & 70.32 & 74.68 & -5.48 & \underline{90.37} & 53.09 \\
& & GEM & 56.00 & 57.94 & 72.73 & 64.50 & 61.84 & 62.60 & -12.39 & 69.54 & 31.14 \\
& & GEM* & 66.35 & 70.11 & 84.09 & 82.75 & 67.11 & 74.08 & -8.99 & 85.30 & \underline{57.43} \\
& & Coil & 60.49 & 60.32 & 73.86 & 61.83 & 60.53 & 63.41 & -10.02 & 73.24 & 46.05 \\
& & Coil* & 70.57 & 78.57 & 86.36 & 75.95 & 64.47 & \underline{75.18} & \underline{-5.81} & 90.08 & 56.97 \\
& & Foster & 61.39 & 63.23 & 65.91 & 63.12 & 65.79 & 63.89 & -12.53 & 82.73 & 39.65 \\
& \multirow{-10}{*}{1000} & Foster* & 72.40 & 76.46 & 81.82 & 73.48 & 73.68 & \textbf{75.57} & \textbf{-5.73} & \textbf{91.15} & \textbf{60.27} \\ 
\cline{2-12} 
& & iCaRL* & 66.30 & 69.05 & 76.14 & 68.77 & 65.79 & 69.21 & -7.78 & 88.83 & \textbf{55.02} \\
& & BiC* & 69.75 & 68.78  & 73.86  & 70.72 & 71.05 & 70.83 & -7.47 & 88.85 & 43.85\\
& & GEM* & 58.79 & 60.05 & 72.73 & 62.74 & 60.53 & 62.97 & -16.85 & 52.43 & 24.76 \\
& & Coil* & 72.73 & 75.40 & 75.23 & 65.43 & 65.79 & \underline{70.92} & \textbf{-5.80} & \textbf{89.69} & 46.18 \\
\multirow{-15}{*}{CDD1} & \multirow{-5}{*}{500} & Foster* & 68.05 & 68.52 & 81.82 & 68.16 & 69.74 & \textbf{71.26} & \underline{-7.18} & \underline{89.28} & \underline{48.05} \\ \hline
& & LwF & 48.63 & 49.74 & 53.41 & 48.86 & 60.53 & 52.23 & -27.67 & 60.30 & \underline{48.80} \\
& & LwF* & 52.81 & 55.82 & 64.77 & 51.89 & 67.11 & 58.48 & -18.08 & \textbf{87.14} & \textbf{62.94} \\
& & EWC & 54.65 & 57.92 & 69.12 & 51.98 & 61.23 & 58.98 & -23.96 & 67.11 & 45.83 \\
& & EWC* & 60.78 & 62.17 & 77.27 & 64.35 & 59.21 & \underline{64.76} & -16.00 & \underline{78.18} & 38.09\\
& \multirow{-5}{*}{1000} & S-Prompts* & 57.61 & 67.72 & 76.14 & 70.91 & 67.11 & \textbf{67.90} & \textbf{-2.77} & 75.63 & 48.76 \\
% &  & S-liPrompts & 59.36 & 69.58  & 76.14  & 66.54 & 72.37 & 68.79 & -8.51 & 74.92 & 43.60 \\
\cline{2-12} 
& & LwF* & 50.52 & 55.03 & 54.55 & 59.65 & 61.84 & 56.32 & -25.19 & \underline{72.32} & \textbf{52.54} \\
& & EWC* & 57.14 & 65.87 & 69.32 & 67.06 & 53.95 & \underline{62.67} & \underline{-19.74} & 67.14 & 36.27 \\
\multirow{-8}{*}{CDD2} & \multirow{-3}{*}{500} & S-Prompts* & 55.67 & 64.29 & 72.73 & 61.55 & 67.11 & \textbf{64.27} & \textbf{-5.80} & \textbf{73.67} & \underline{37.53}\\
\hline
&  & LwF* & 51.32 & 51.06 & 51.14 & 51.19 & 51.32 & 51.21 & -11.70 & 51.24 & 28.19 \\
&  & EWC* & 51.01 & 50.06 & 51.14 & 51.19 & 49.32 & 50.54 & -11.65 & 63.77 & 23.54 \\
&  & S-Prompts* & 49.72 & 48.94 & 48.86 & 48.81 & 48.68 & 49.01 & -11.04 & 60.32 & 39.20 \\
\cline{3-12} 
&  & LwF*+Ours & 84.74 & 50.79 & 60.23 & 73.91 & 56.58 & \underline{65.25} & -0.66 & \textbf{87.70} & \underline{56.99}\\
&  & EWC*+Ours & 87.57 & 49.21 & 70.45 & 63.83 & 61.84 & \textbf{66.58} & \textbf{0.43} & 83.32 & 51.39 \\ 
\multirow{-6}{*}{CDD3} & \multirow{-6}{*}{0} & S-Prompts*+Ours & 95.65 & 48.41 & 57.95 & 69.49 & 52.63 & 64.83 & \underline{-0.21} & \underline{87.18} & \textbf{68.52} \\
\hline
&  & ResNet-FT & 86.05 & 76.19 & 71.36 & 83.89 & 75.26 & 78.55 & NA & 81.09 & 58.32 \\
&  & ViT-FT & 96.08 & 88.36 & 89.77 & 95.44 & 86.84 & 91.30 & NA & 99.20 & 94.35 \\ 
\multirow{-3}{*}{JDD} & \multirow{-3}{*}{NA} & ViT-PT & 93.48 & 81.22 & 77.27 & 93.39 & 86.84 & 86.44 & NA & 98.92 & 90.15 \\ \hline
\end{tabular}
}
\vspace{-0.3cm}
\caption{Results of the three CDD benchmark evaluations. JDD: Joint-training Deepart Detection that is often regarded as an upper bound of continual learning settings \cite{wang2022sprompt}. *: using pre-trained ViT-B/16, w/o *: using ResNet, FT: fine-tuning, PT: prompt-tuning, AA: average detection accuracy, mAP: mean average precision, CA: copyright identification accuracy, \textbf{Bold}: \textbf{best}, \underline{Underline}: \underline{second best}}
\label{tab:mainresult}
\vspace{-0.2cm}
\end{table*}

\begin{figure*}[t] 
\centering 
\includegraphics[width=0.95\textwidth]{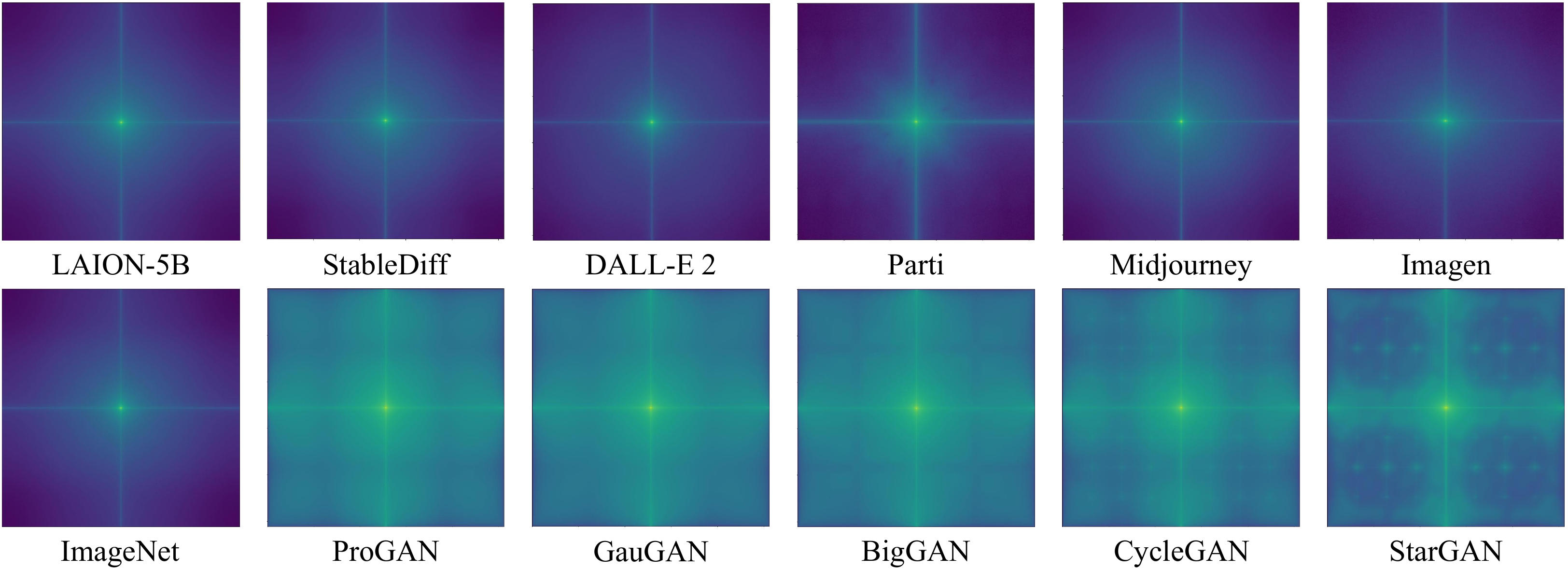} 
\vspace{-0.2cm}
\caption{Frequency analysis on each deepfake/deepart model generated images. The periodic patterns (dots or lines) of our used deepart models (StableDiff, DALL-E2, Parti, Midjourney, Imagen) generated deepfake art images are highly close to those of conarts from LAION-5B and the real images from ImageNet. In contrast, images generated by early deepfake models (ProGAN~\cite{karras2018progressive},GauGAN~\cite{park2019SPADE}, BigGAN~\cite{brock2018large}, CycleGAN~\cite{zhu2017unpaired}, StarGAN~\cite{choi2018stargan}) have different periodic patterns.}
\label{fig:vis1}
\vspace{-0.4cm}
\end{figure*}

\begin{figure}[h] 
\centering 
\includegraphics[width=0.75\textwidth]{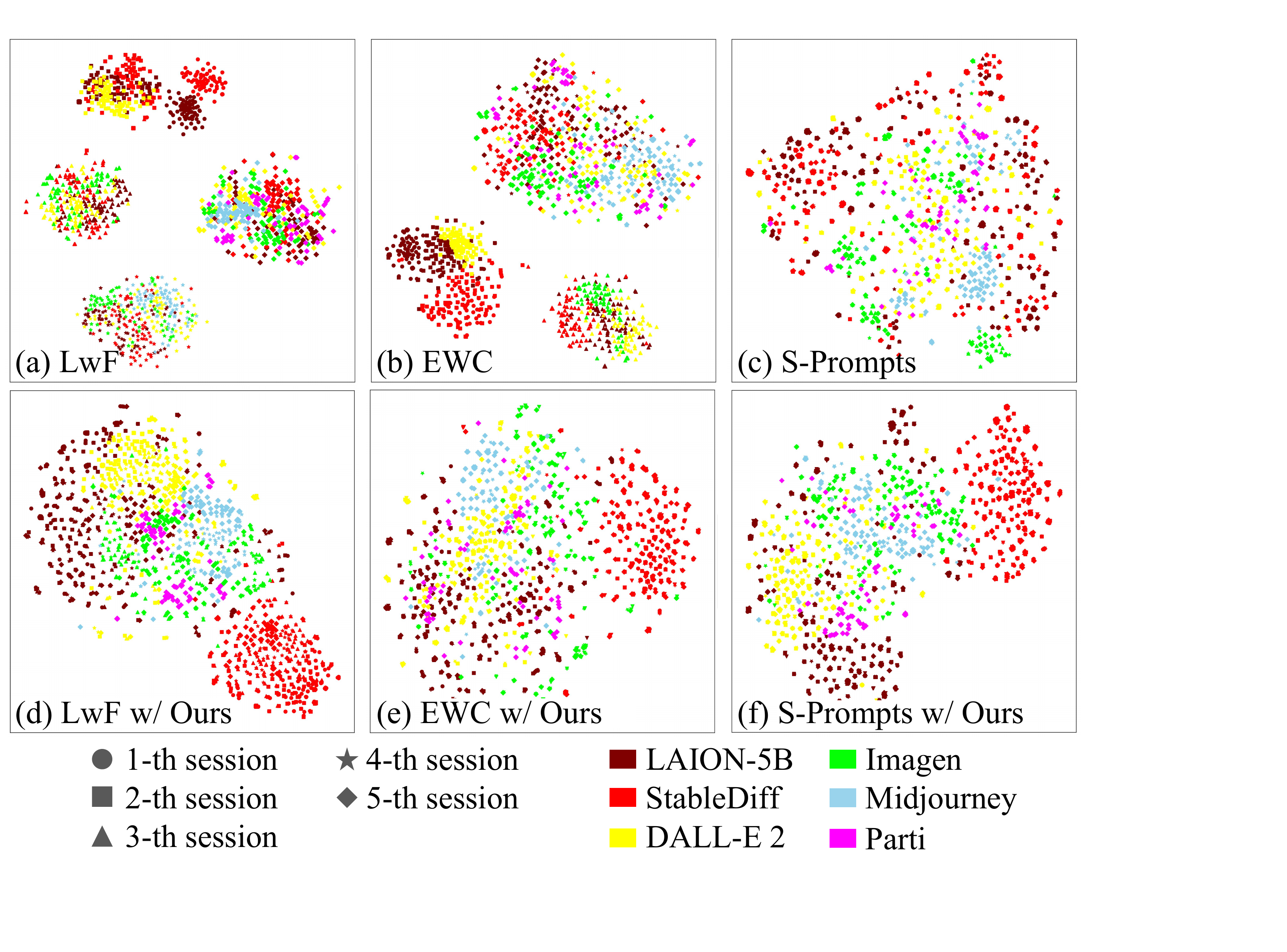}
% \vspace{-0.7cm}
\caption{t-SNE visualization of learned feature spaces of LwF, EWC and S-Prompts for CDD3. (a), (b), (c): without using our proposed transformation framework, (d), (e), (f): using our framework.}
\label{fig:feture_bias}
\end{figure}

\noindent
\textbf{CDD Results.}
Table~\ref{tab:mainresult} reports the evaluation results of the competing methods for the three CDD benchmarks.
By comparing the rehearsal-based methods for CDD1, we find that Foster outperforms the other methods on both two buffer-size settings.
It obtains a slightly $0.5\%$ relative improvement over Coil in terms of final average accuracy, while it has to save an additional network during incremental learning.
Among the rehearsal-free methods for CDD2, S-Prompts achieves the best performance when the rehearsal buffer size is $1000$.
When the buffer size is decreased to $500$, its accuracy drops around $6.12\%$.
The performances of both rehearsal-based and rehearsal-free methods drop significantly when the memory size becomes smaller for CDD1 and CDD2.
For CDD3, all the original rehearsal-free methods collapse, as their results are far lower than those of the ODD methods and even just above random guesses.
\emph{After applying our framework, all the methods achieve remarkable improvements (relative improvement of $27.41\%$  for LwF in terms of AA, relative gain of $19.53\%$ for EWC, and relative improvement of $32.30\%$ for S-Prompts).}
As discussed in Sec.\ref{subsec:rescue}, one of the main reasons is these methods suffer from less feature bias after applying our framework. This can be verified by Figure~\ref{fig:feture_bias}, where each deepart model's feature subspace has less shift after going through the 5 learning sessions when using ours.
Besides, it is fairly surprising that some of the adapted CDD3 methods even surpasses most of the CDD2 methods (besides S-Prompts).
This is very likely that using the base session's conart data only makes the binary classifier trainable, but these methods still suffer from severe forgetting.
It can also be seen from the AF metric that our adapted CDD3 methods have very low forgetting, even outperforming those CDD1 methods with $1000$ samples.
The less forgetting may attribute to our suggested PT that reduces feature bias.

\begin{table}[] 
\vspace{-0.3cm}
\centering
\resizebox{0.35\linewidth}{!}{
\begin{tabular}{ccccc}   
\hline
KD & CN & PT & AA & AF \\ \hline
&  &  & 50.01 & -14.53 \\
\checkmark  &   &  & 51.21 & -11.70 \\
& \checkmark  &  & 51.20 & -12.20 \\
& & \checkmark & 51.74 & -11.01 \\
\checkmark  & \checkmark &  & 59.43 & -5.17 \\
  & \checkmark & \checkmark & 58.20 & -5.53 \\
\checkmark  &  & \checkmark & 57.83 & -9.12 \\
\checkmark  & \checkmark & \checkmark & 65.25 & -0.66 \\
\hline
\end{tabular}
}
\vspace{-0.3cm}
\caption{Ablation Study. KD: Knowledge Distillation, CN: Cosine Normalization, PT: Prompt-tuning, AA: Average Accuracy, AF: Average Forgetting degree.}
\label{table:ablation}
\vspace{-0.5cm}
\end{table}

\noindent
\textbf{Ablation Study.} Table \ref{table:ablation} shows the ablation on the three main components of our suggested framework based on LwF, with the most preliminary case being the one that merely performs fine-tuning without any constraints.
The ablation justifies the necessity of using knowledge distillation, Cosine normalization, and prompt-tuning for CDD3. It can be seen that uniting all the three components can achieve the most promising result.

\noindent
\textbf{Deeparts vs. Deepfakes.}
We follow \cite{wang2020cnn} to visualize the average frequency spectra using Discrete Fourier Transform on each deepfake/deepart model' generated images in Figure \ref{fig:vis1}. The results show that the conart (LAION-5B) images, the real non-art (ImageNet) images, and our collected deepart images look very alike (all have very few periodic patterns), while the deepfake images (from GAN models used in \cite{li2022continual}) have many visible patterns (dots or lines). This reflects that the deeparts are very close to the real ones and much more photorealistic that the deepfakes, showing the great challenge of our suggested deepart detection.
\section{Conclusions and Discussions}
\label{sec:conclusion}

Deepfake artworks have recently attracted much attention, and the Internet has been flooded with many artificial intelligence works.
How to detect and identify deepart is being put on the agenda.
We have responded promptly to this surge and have launched the study on deepart detection, with four contributions. We establish the first deepart detection database (DDDB). We then define two novel deepart detection problems (ODD and CDD) as well as four benchmarks. We further suggest four families of solutions for these four benchmarks respectively, and propose strategies to enhance previous deepfake detection and continual learning methods. Finally we perform extensive evaluation, which demonstrates the effectiveness of our proposed solutions and more importantly provide the first wave's results to facilitate the study of deepart detection.

Deepart detection is an open problem, and we expect our work could pave a good way to more interesting directions. As for future work, some promising avenues for deepart detection can be further explored. Examples include opening an access to easily-acquired/low-quality conventional artworks, collecting more high-quality conarts and deeparts from public resources like \cite{wang2022diffusiondb}, and leveraging the associated prompts of deeparts.
 
\section*{Acknowledgments}
This work is funded by the Singapore Ministry of Education (MOE) Academic Research Fund (AcRF) Tier 1 grant (MSS21C002). This work is also supported by the National Natural Science Foundation of China (62076195) and the Fundamental Research Funds for the Central Universities (AUGA5710011522).

\bibliography{07_references}
\bibliographystyle{iclr2022_conference}

\appendix

\appendix
\label{sec:appendix}

\section{Computation of Suggested Benchmark Evaluation Metrics}

There are four main evaluation metrics of DDDB benchmarks, including average detection accuracy (AA), average forgetting degree (AF), mean average precision (mAP) and copyright identification accuracy (CA).

Following \cite{li2022continual, lopez2017gradient}, we use the accuracy matrix $B \in \mathbb{R}^{n\times n}$, where each entry $B_{i,j}$ indicates the test accuracy of the $i$-th session after training the $j$-th session and $n$ is the total number of involved tasks.
The ODD benchmark only has base session training and its accuracy matrix has only one row, while the CDD benchmarks have multiple rows in their accuracy matrices.
For both ODD and CDDs, AA can be computed by
\begin{equation}
    AA=\frac{1}{n}\sum_{i=1}^{n}B_{i,n},
\label{eq:aa}
\end{equation}
and AF can be calculated for CDDs by
\begin{equation}
    AF = \frac{1}{n-1} \sum_{i=1}^{n-1} BWT_{i},
\label{eq:af}
\end{equation}
where $BWT_{i}= \frac{1}{n-i-1}\sum_{j=i+1}^{n} (B_{i,j}-B_{i,i}) $.

The mAP score is calculated by the mean of areas under the precision-recall (PR) curves.
The calculation of mAP for the ODD benchmark is the same as~\cite{wang2020cnn}.
For the CDD benchmarks, we follow ~\cite{li2022continual} to adopt the multi-class classifier for binary classification, and we use the same way to calculate the mAP score.

The calculation of the copyright identification accuracy (CA) for CDDs is
\begin{equation}
    CA = \frac{N^{TP}_{deeparts}+N^{TN}_{conarts}}{N_{total}},
\label{eq:ca}
\end{equation}
where $N^{TP}_{deeparts}$ is the true positives that deeparts are classified as their corresponding generators, and $N^{TN}_{conarts}$ is the true negatives that conarts are classified correctly. $N_{total}$ is the total test samples. All the conarts are classified as single one class in the CA calculation.

\section{Detailed Algorithm of Proposed Transformation Framework on Rehearsal-free Methods for CDD3}

\begin{algorithm} [t]
\caption{Transformation process of proposed framework}
\label{alg_train}
\renewcommand{\algorithmicrequire}{\textbf{Input:}}
\renewcommand{\algorithmicensure}{\textbf{Output:}}
\begin{algorithmic}[1]
\STATE Initialization: Target Method, Cosine Normalization (CN), Knowledge Distillation (KD), Prompt-Tuning (PT).
% \STATE Prompting Setup within the Feature Extractor: 
\IF{Target Method uses Pre-trained ViT}
\STATE Add Prompts to the pre-tranined ViT.
\STATE Freeze all parameters of ViT.
\STATE Switch to PT.
\ELSE
\STATE Exchange backbone to ViT.
\STATE Add Prompts to the pre-tranined ViT.
\STATE Freeze all parameters of ViT.
\STATE Switch to PT.
\ENDIF
\IF{Target Method not use CN}
\STATE Add CN to the Classifier Head.
\ENDIF
\IF{Target Method not use KD}
\STATE Add KD to the training loss.
\ENDIF
\end{algorithmic}
\end{algorithm}

The overall algorithm of the proposed transformation framework is presented in Alg.\ref{alg_train}. It is an algorithm for the transformation on given rehearsal-free methods for CDD3. The used three main techniques are detailed below.

\textbf{Knowledge Distillation (KD).}
Inspired by~\cite{li2017learning}, we adopt knowledge distillation to mimic the response of previous session's models.
Formally, knowledge distillation can be commonly computed by
\begin{equation}
    L_{kd} = -\sum_{i=1}^{C} \frac{\exp (v_i / \tau )}{\sum_{C}^{j=1} \exp (v_j / \tau)}  \log ( \frac{\exp (z_i / \tau )}{\sum_{C}^{j=1} \exp (z_j / \tau)}  ),
\label{eq:kd_loss}
\end{equation}
where $v_i$ is the logits of image $x$ produced by previous model, $z_i$ is the logits produced by current model, $\tau$ is the temperature parameter and $C$ is the total class number.

\textbf{Cosine Normalization (CN).}
Cosine Normalization is introduced by previous works~\cite{zhou2021co} in incremental learning to tackle class imbalance problem.
Cosine classifier uses the normalized features and class weights to do multiplication:

\begin{equation}
    \varphi (x) = \left ( \frac{ W }{\left \| W \right \| _2 }  \right ) ^{T} \left ( \frac{ f(x) }{\left \| f(x) \right \| _2 } \right ) ,
\label{eq:cosine_classifier}
\end{equation}
where $f(x)$ is the image feature produced by the backbone network $f$, $W$ is the class weights, and $\varphi (x)$ is the predict scores before softmax.

Due to the output after cosine normalization being restricted to the range of $\left [ -1,1 \right ] $, the score magnitudes produced by separate classifiers can be calibrated to the same level. 
Cosine classifier can tackle the highly imbalanced data problem that we can only access to positive data during incremental learning.

\textbf{Prompt-Tuning (PT).}
Prompt-tuning uses a small set of learnable parameters (named prompts) as part of the input to transfer the pre-trained model to downstream tasks.
During transfer learning, all transformer parameters are frozen except only the prompts can be updated.

Previous continual learning methods~\cite{wang2021learning, wang2022sprompt} only use prompts on the first input layer, restricting the model transfer ability.
Instead, like \cite{jia2022vpt}, we suggest adopting prompts at every Transformer layer, and the input tokens $x^i$ for the i-th layer are $x^i = [x^i_{img}, p^i ,x^i_{cls}]$,
where $x^i_{img}$ and $x^i_{cls}$ are the image tokens and the pre-trained class tokens of ViT at $i$-th layer respectively. $p^i$ is the prompts at $i$-th layer. The total set of prompts is denoted by $\mathcal{P} = \left \{ p^i \in \mathbb{R}^{l \times d} | 0<i< \Psi  \right \}$,
where $d$ is the embedding dimension of transformer, $l$ is the length of the corresponding prompts, and $\Psi$ represents the prompt depth. 
A classifier head (usually a linear full connected layer) mapping the final layer's class token is used to calculate the predication as $\varphi (x) = \mathrm {Head}(x_{cls}^{final}).$

As studied in \cite{wang2021learning, wang2022sprompt}, prompt-tuning is an efficient weight constraint for continual learning to tackle feature bias and catastrophic forgetting.
When we update all model parameters on single-class data, the model is able to easily change the feature space to overfit the current session's data.
Nevertheless, as the model only updates a few parameters on a new task, it can be forced to learn a classifier on a relatively good feature space, as shown in Fig. 4 of the main paper.

\section{Training Details of Suggested/Transformed Methods}
All of our experiments were implemented by PyTorch with four NVIDIA RTX 3090 GPUs. 
To ensure a fair comparison, each method starts with the same ImageNet pre-trained ViT-B/16, or we use their origin implementation (usually ResNet-18 for incremental learning on Imagenet).
All methods keep their specific hyper-parameters as their original implementations.
For methods using pre-trained ViT as the backbone, we only tune the learning rate and the number of epochs.
We use $10$ epochs with $0.001$ learning rate for all pre-trained ViT-based methods throughout our experiments.
The proposed framework is insensitive to the setting of hyper-parameters and we do not tune much.
Specifically, we use $30$ epochs with an initial learning rate $0.1$ for all the adapted CDD3 methods (e.g., EWC, LwF and S-Prompts), since prompt-tuning needs more iterations to fit the training sets.
We adopt SGD optimizer with a momentum of 0.9 and the cosine annealing scheduler~\cite{loshchilov2016sgdr}.
The prompt length $l$ we use in the main paper is $10$, and the depth $\Psi$ we use is $11$.
The knowledge distillation loss is the same as that in LwF.
We use the batch size of $128$ for all methods.
The data augmentation strategy we use for all methods is simple.
We resize all images to $224 \times 224$ and use random horizontal flip and random crop.

\section{Extra Memory Overhead of Proposed Transformation Framework}

The main additional memory overhead of the proposed framework lies in the extra use of prompts for prompt-tuning.
The length of prompts $l$ is $10$ and the dimension of a single prompt $d$ is $768$.
The depth $\Psi$ we add prompts to ViT is $11$ and thus the total needed parameter for prompt-tuning is $84,480$.
The prompts are initialized at the beginning, and we add fixed parameters to the original pre-trained ViT-B/16 model throughout incremental learning, leading to a paltry total parameter increase of $0.09\%$.

\end{document}